\documentclass{qam-l}

\usepackage{amssymb}

\usepackage{graphicx}
\usepackage{array}

\theoremstyle{definition}

\theoremstyle{remark}

\numberwithin{equation}{section}

\usepackage{times}
\usepackage{helvet}
\usepackage{courier}
\usepackage{graphicx} 
\usepackage{subfig}
\usepackage{algorithm}
\usepackage{algorithmic}
\usepackage{amsmath}
\usepackage{url}
\usepackage{amsmath}
\usepackage{amssymb}
\usepackage{mathrsfs}
\usepackage{array}
\usepackage{bm}
\usepackage{multirow}
\usepackage{hyperref}
\usepackage{tikz}
\usetikzlibrary{matrix}
\usepackage{comment}

\def\E{{\rm E}}
\def\N{{\rm N}}
\def\KL{{\rm KL}}
\def\P{P_{\rm data}}
\def\L{\mathcal{L}}

\def\hh{\hat{h}}
\def\th{\tilde{h}}
\def\hX{\hat{X}}
\def\tX{\tilde{X}}
\def\tn{\tilde{n}}

\def\tn{\tilde{n}}

\def\T{{\top}}

\begin{document}

\title[Three Families of Models]{A Tale of Three Probabilistic Families: Discriminative, Descriptive and Generative Models}

\author{Ying Nian Wu}
\address{Department of Statistics, University of California, Los Angeles}

\author{Ruiqi Gao}
\address{Department of Statistics, University of California, Los Angeles}

\author{Tian Han}
\address{Department of Statistics, University of California, Los Angeles}

\author{Song-Chun Zhu}
\address{Department of Statistics, University of California, Los Angeles}

\date{}

\begin{abstract}

The pattern theory of Grenander is a mathematical framework where patterns are represented by probability models on random variables of algebraic structures. In this paper, we review three families  of probability models, namely,  the discriminative models, the descriptive models, and the generative models. A discriminative model is in the form of a classifier. It specifies the conditional probability of the class label  given the input signal. A descriptive model specifies the probability distribution of the signal, based on an energy function defined on the signal. A generative model assumes that the signal is generated by some latent variables via a transformation. We shall review  these models within a common framework and explore their connections. We shall also review the recent developments that take advantage of the high approximation capacities of deep neural networks.   \\

\end{abstract}

\maketitle

\section{Introduction}

Initially developed by Grenander in the 1970s,  the pattern theory   \cite{grenander1970unified, grenander2007pattern}  is a unified mathematical framework for representing, learning and recognizing patterns that arise in science and engineering. The objects in pattern theory are usually of high complexity or dimensionality, defined in terms of  the constituent elements and the bonds between them. The patterns of these objects are characterized by both the algebraic structures governed by local and global rules, as well as the probability distributions of the associated random variables. Such a framework encompasses most of the probability models in various disciplines. In the 1990s, Mumford  \cite{mumford2010pattern}  advocated the pattern theoretical framework for computer vision, so that learning and inference can be based on probability models. 

Despite its generality, developing probability models in the pattern theoretical framework remains a challenging task.  In this article, we shall review three families of models,  which we call the discriminative models, the descriptive models, and the generative models, following the terminology of \cite{zhu2003statistical}. A discriminative model is in the form of a classifier. It specifies the conditional probability of the output class label  given the input signal. Such a model can be learned in the supervised setting where a training dataset of input signals and the corresponding output labels is provided. A descriptive model specifies the probability distribution of the signal, based on an energy function defined on the signal through some descriptive feature statistics extracted from the signal. Such models originated from statistical physics, where they are commonly called the Gibbs distributions \cite{gibbs2014elementary}. The descriptive models belong to the broader class of energy-based models \cite{Lecun2006} that include non-probabilistic models as well as models with latent variables. A generative model assumes that the signal is generated by some latent variables via a deterministic transformation. A prototype example  is factor analysis \cite{rubin1982algorithms}, where the signal is generated by some latent factors via a linear transformation. Both the descriptive models and generative models can be learned in the unsupervised setting where the training dataset only consists of input signals without the corresponding output labels. 

 In this paper,  we shall review  these three families of models within a common framework and explore their connections. We shall start from the flat linear forms of these models. Then we shall present the hierarchical non-linear models,  where the non-linear mappings in these models are parametrized by neural networks \cite{lecun1998gradient, krizhevsky2012imagenet} that have proved exceedingly effective in approximating non-linear relationships.     
  
Currently the most successful family of models are the discriminative models. A discriminative model is in the form of the  conditional  distribution of the class label given the input signal. The normalizing constant of such a probability model is a  summation over the finite number of class labels or categories.  It is readily available, so that the model can be easily learned from big datasets. The learning of the descriptive models and the generative models can be much more challenging. A descriptive model is defined as a probability distribution of the signal, which is usually of a high dimensionality. The normalizing constant of such a model is an integral over the high dimensional signal and is analytically intractable. A generative model involves latent variables that follow some prior distribution, so that the marginal distribution of the observed signal is obtained by integrating out the latent variables, and this integral is also analytically intractable. Due to the intractabilities of the integrals in the descriptive and generative models, the learning of such models usually requires Markov chain Monte Carlo (MCMC) sampling \cite{geman1984stochastic, liu2008monte}. Specifically, the learning of the descriptive models require MCMC sampling of the synthesized signals,  while  the learning of the generative models require MCMC sampling of  the latent variables. Nonetheless, we shall show that such learning methods work reasonably well \cite{XieLuICML, gao2017learning, HanLu2016},  where the gradient-based Langevin dynamics \cite{neal2011mcmc} can be employed conveniently for MCMC sampling, which is an inner loop within the gradient-based learning of the model parameters. 

Because of the high capacity of the neural networks in approximating highly non-linear mappings, the boundary between representation and computation is  blurred in neural networks. A deep neural network can be used to represent how the signal is generated or how the features are defined. It can also be used to approximate the solution of a computational problem such as optimization or sampling. For example, the iterative sampling  of the latent variables of a generative model can be  approximated by an inference model that provides the  posterior samples directly, as is the case with the wake-sleep algorithm \cite{hinton1995wake} and the variational auto-encoder (VAE) \cite{KingmaCoRR13, RezendeICML2014, MnihGregor2014}. As another example, the iterative sampling of a descriptive model can be approximated by a generative model that can generate the signal directly \cite{CoopNets2016, coopnets2018}.  In general, the solutions to the on-line computational problems can be encoded  by high capacity neural networks, so that iterative computations only occur in the off-line  learning of the model parameters. 

The three families of models do not exist in isolation. There are intimate connections between them. In \cite{guo2003modeling, guo2007primal}, the authors proposed to integrate the descriptive and generative models into a hierarchical model. In \cite{tu2002image, tu2006parsing}, the authors proposed data-driven MCMC where the MCMC is to fit the generative models, but the proposal distributions for MCMC transitions are provided by discriminative models. The discriminative model and the descriptive model can be translated into each other via the Bayes rule. Tu \cite{tu2007learning} exploited this relationship to learn the descriptive model via discriminative training, thus unifying the two models. Similarly, the discriminative model can be  paired with the generative model in the generative adversarial networks (GAN) \cite{goodfellow2014generative}, and  the adversarial learning has become an alternative framework to likelihood-based learning. The descriptive model and the generative model can also be paired up so that  they can jumpstart each other's MCMC sampling \cite{CoopNets2016, coopnets2018}. Moreover, the family of descriptive models and the family of generative models overlap in terms of undirected latent energy-based models \cite{Lecun2006}.

\section{Non-hierarchical linear forms of the three families} 

We shall first review the non-hierarchical linear forms of the three families of models within a common framework. 

\subsection{Discriminative models}

This subsection reviews the linear form of the discriminative models. 

The  table below displays the dataset for training the discriminative models: 
\begin{table}[h]
\centering
\begin{tabular}{|c|c|c|c|}
\hline
  & input & features & output \\
\hline\hline
 1   & $X_1^\T$ & $h_1^\T$         & $Y_1$  \\
 2   & $X_2^\T$ & $h_2^\T$         &  $Y_2$\\
...  &      &           &\\
$n$ &$X_n^\T$  &  $h_n^\T$        &  $Y_n$\\
\hline
\end{tabular}
\end{table}

There are $n$ training examples. For the $i$-th example, let $X_i = (x_{ij}, j = 1, ..., p)^\T$ be the $p$-dimensional input signal (the $(n, p)$ notation is commonly used in statistics to denote the number of observations and the number of predictors respectively). Let $Y_i$ be the outcome label. In the case of classification, $Y_i$ is categorical or binary. $h_i = (h_{ik}, k = 1, ..., d)^\T$ is the $d$-dimensional vector of features or hidden variables. 

The discriminative models can be represented by the diagram below:
\begin{eqnarray}
\begin{array}[c]{cc}
{\rm output}: & Y_i\\
& \uparrow\\
{\rm features}: & h_i\\
& \uparrow\\
{\rm input}: & X_i
\end{array}  
\label{eq:d1}
\end{eqnarray}
where the vector of features $h_i$ is computed from $X_i$ via $h_i = h(X_i)$. In a non-hierarchical or flat model, the feature vector $h_i$ is designed, not learned, i.e., $h()$ is a pre-specified non-linear transformation. 

For the case of binary classification where $Y_i \in \{+1, -1\}$, $(Y_i, X_i)$ follow a logistic regression 
\begin{eqnarray}
    \log \frac{\Pr(Y_i = +|X_i)}{\Pr(Y_i = -|X_i)} = h_i^\T \theta + b, \label{eq:logistic}
\end{eqnarray}
where $\theta$ is the $d$-dimensional vector of weight or coefficient parameters, and $b$ is the bias or intercept parameter. The classification can also be based on the perceptron model 
\begin{eqnarray}
\hat{Y}_i = {\rm sign}(h_i^\T \theta + b),
\end{eqnarray}
 where ${\rm sign}(r) = +1$ if $r\geq 0$, and ${\rm sign}(r) = -1$ otherwise. Both the logistic regression and the perceptron can be  generalized to the multi-category case. The bias term $b$ can be absorbed into the weight parameters $\theta$ if we fix $h_{i1} = 1$. 

Let $f(X) = h(X)^\T \theta$. $f(X)$ captures the relationship between $X$ and $Y$.  Because $h(X)$ is non-linear, $f(X)$ is also non-linear. We say  the model is  in the linear form because  it is linear in $\theta$, or $f(X)$ is a linear combination of the features in $h(X)$. The following are the choices of $h()$ in various discriminative models. 

 {\em Kernel machine} \cite{cortes1995support}: $h_i = h(X_i)$ is implicit,  and the dimension of $h_i$ can potentially be infinite. The implementation of this method is based on the kernel trick $\langle h(X), h(X')\rangle = K(X, X')$, where $K$ is a kernel that is explicitly used by the classifier such as the support vector machine  \cite{cortes1995support}. $f(X) = h(X)^\T \theta$ belongs to the reproducing kernel Hilbert space where the norm of $f$ can be defined as the Euclidean norm of $\theta$, and the norm is used to regularize the model. A Bayesian treatment leads to the Gaussian process, where $\theta$ is assumed to follow ${\rm N}(0, \sigma^2 I_d)$, and $I_d$ is the identity matrix of dimension $d$. $f(X)$ is a Gaussian process with ${\rm Cov}(f(X), f(X')) = \sigma^2 K(X, X')$. 

 {\em Boosting machine} \cite{freund1997decision}: For $h_i = (h_{ik}, k = 1, ..., d)^\T$, each $h_{ik} \in \{+, -\}$ is a weak classifier or a binary feature extracted from $X$, and $f(X) = h(X)^\T \theta$ is a committee of weak classifiers. 

 {\em CART}   \cite{breiman1984classification}: In the classification and regression trees, there are $d$ rectangle regions $\{R_k, k = 1, ..., d\}$ resulted from recursive binary partition of the space of $X$, and each $h_{ik} = 1(X_i \in R_k)$ is the binary indicator such that $h_{ik} = 1$ if $X_i \in R_k$ and $h_{ik} = 0$ otherwise. $f(X) = h(X)^\T \theta$ is a piecewise constant function.

 {\em MARS}  \cite{friedman1991multivariate}: In the multivariate adaptive regression splines, the components of $h(X)$ are hinge functions such as  $\max(0, x_j -  t)$ (where $x_j$ is the $j$-th component of $X$, $j = 1, ..., p$, and $t$ is a threshold) and their products. It can be considered a continuous version of CART. 
 
{\em Encoder and decoder}: In the diagram in (\ref{eq:d1}), the transformation $X_i \rightarrow h_i$ is called an encoder, and the transformation $h_i \rightarrow Y_i$ is called a decoder. In the non-hierarchical model, the encoder is designed, and only the decoder is learned. 

The outcome $Y_i$ can also be continuous or a high-dimensional vector. The learning then becomes a regression problem. Both classification and regression are about supervised learning because for each input $X_i$, an output $Y_i$ is provided as supervision. The reinforcement learning is similar to supervised learning except that the guidance is in the form of a reward function.

\subsection{Descriptive models} 

This subsection describes the linear form of the descriptive models and the maximum likelihood learning algorithm. 

The descriptive models \cite{zhu2003statistical} can be learned in the unsupervised setting, where $Y_i$ are not observed, as illustrated by the table below:
 \begin{table}[h]
\centering
\begin{tabular}{|c|c|c|c|}
\hline
  & input & features & output \\
\hline\hline
 1   & $X_1^\T$ & $h_1^\T$         & ?  \\
 2   & $X_2^\T$ & $h_2^\T$         &  ?\\
...  &      &           &\\
$n$ &$X_n^\T$  &  $h_n^\T$        &  ?\\
\hline
\end{tabular}
\end{table}

The linear form of the descriptive model is an exponential family model. It specifies a probability distribution on the signal $X_i$ via an energy function that is a linear combination of the features,  
\begin{align} 
  p_\theta(X) &= \frac{1}{Z(\theta)} \exp \left[ h(X)^\T \theta \right] p_0(X), \label{eq:D0}
\end{align}
where $h(X)$ is the $d$-dimensional feature vector extracted from $X$, and $\theta$ is the $d$-dimensional vector of weight parameters. $p_0(X)$ is a known reference distribution such as the white noise model $X \sim {\rm N}(0, \sigma^2 I_p)$, or the uniform distribution within a bounded range. 
\begin{eqnarray}
Z(\theta) = \int \exp[h(X)^\T \theta] p_0(X) dX = \E_{p_0}\{\exp[h(X)^\T \theta]\}
\end{eqnarray}
 is the normalizing constant ($\E_{p}$ denotes the expectation with respect to $p$). It is analytically intractable.

The descriptive model (\ref{eq:D0}) has the following information theoretical property \cite{della1997inducing, zhu1997minimax, amari2007methods}. Let $\P$ be the distribution that generates the training data $\{X_i\}$. Let $\Theta = \{p_\theta, \forall \theta\}$ be the family of distributions defined by the descriptive model.  Let $\Omega = \{p: \E_{p}[h(X)] = \hat{h}\}$, where $\hat{h} = \E_{\P}[h(X)]$.  $\hat{h}$ can be estimated from the observed data by the sample average $\sum_{i=1}^{n} h(X_i)/n$. $\Omega$ is the family of distributions that reproduce the observed $\hat{h}$. 
 \begin{figure}[h]
	\begin{center}
		\includegraphics[width=.3\linewidth]{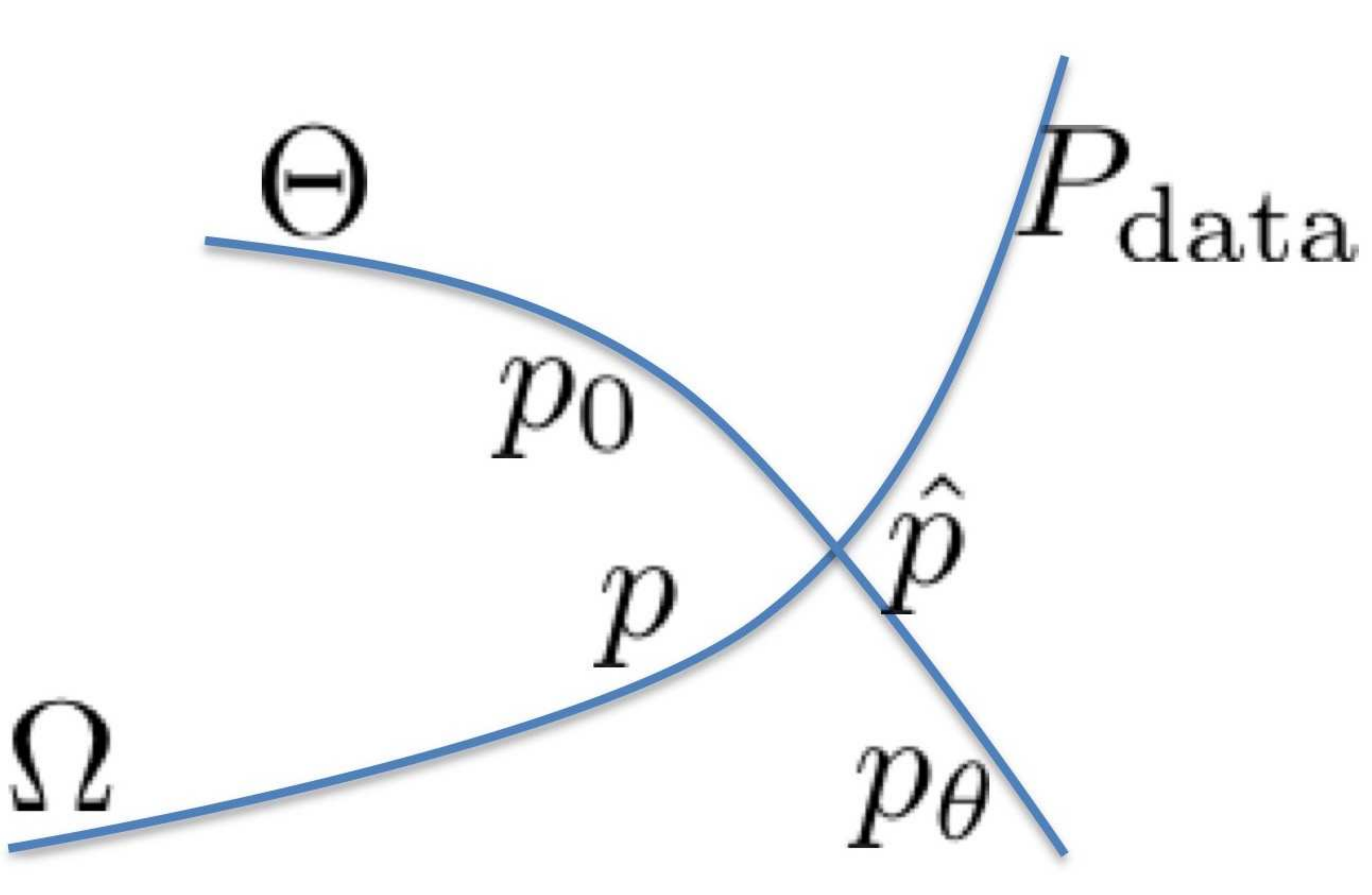}			
		\caption{The two curves illustrate $\Theta$ and $\Omega$ respectively, where each point is a probability distribution.}
		\label{fig:info}
	\end{center}	
\end{figure}
 Let $\hat{p} = p_{\hat{\theta}} \in \Theta \cup \Omega$ be the intersection between $\Theta$ and $\Omega$. Then for any $p_\theta \in \Theta$ and any $p \in \Omega$, we have $\KL(p\|p_\theta) = \KL(p\|\hat{p}) + \KL(\hat{p}\|p_\theta)$, which can be interpreted as a Pythagorean property that defines orthogonality. $\KL(p\|q) = \E_{p}[\log (p(X)/q(X))]$ denotes the Kullback-Leibler divergence from $p$ to $q$. Thus $\Theta$ and $\Omega$ are orthogonal to each other, $\Theta \perp \Omega$, as illustrated by Figure \ref{fig:info}. 

This leads to the following dual properties of $\hat{p}$, which can be considered the learned model: 

 (1) Maximum likelihood.  $\hat{p} = \arg\min_\Theta \KL(\P\|p_\theta)$. That is, $\hat{p}$ is the projection of $\P$ on $\Theta$. $\KL(\P\|p_\theta) = \E_{\P}[\log \P(X)] - \E_{\P}[\log p_\theta(X)]$. The second term $\E_{\P}[\log p_\theta(X)]$ is the population version of the log-likelihood. Thus minimizing $\KL(\P\|p_\theta)$ is equivalent to maximizing the likelihood. 
 
 (2) Maximum entropy: $\hat{p} = \arg\min_\Omega \KL(p\|p_0)$. That is, $\hat{p}$ is the minimal modification of $p_0$ to reproduce the observed feature statistics $\hat{h}$. $\KL(p\|p_0) = \E_p[\log p(X)] - \E_p[\log p_0(X)]$. If $p_0$ is the uniform distribution, then the second term is a constant, and the first term is the negative entropy. In that case, minimizing $\KL(p\|p_0)$ is equivalent to maximizing the entropy over $\Omega$. 

Given the training data $\{X_i\}$, let $\L(\theta) = \sum_{i=1}^{n} \log p_\theta(X_i)/n$ be the log-likelihood. The gradient of $\L(\theta)$ is 
\begin{eqnarray}
     \L'(\theta) = \frac{1}{n} \sum_{i=1}^{n} h(X_i) - \E_\theta[h(X)], 
 \end{eqnarray}
 because $\partial \log Z(\theta)/\partial \theta = \E_\theta[h(X)]$, where $\E_\theta$ denotes the expectation with respect to $p_\theta$. This leads to a stochastic gradient ascent algorithm for maximizing $\L(\theta)$, 
 \begin{eqnarray} 
     \theta_{t+1} = \theta_t + \eta_t  \left[\frac{1}{n} \sum_{i=1}^{n} h(X_i) - \frac{1}{\tn} \sum_{i=1}^{\tn} h(\tX_i)\right], 
 \end{eqnarray}
 where $\{\tX_i, i = 1, ..., \tn\}$ are random samples from $p_{\theta_t}$, and $\eta_t$ is the learning rate. The learning algorithm has an ``analysis by synthesis'' interpretation. The $\{\tX_i\}$ are the synthesized data generated by the current model. The learning algorithm updates the parameters in order to make the synthesized data similar to the observed data in terms of the feature statistics. At the maximum likelihood estimate ${\hat{\theta}}$, the model matches the data: $\E_{\hat{\theta}}[h(X)] = \E_{\P}[h(X)]$. 
    
 One important class of descriptive models are the Markov random field models   \cite{besag1974spatial, geman1986markov}, such as the Ising model in statistical physics. Such models play an important role in the pattern theory.

 \begin{figure}[h]
\begin{center}
\includegraphics[width=0.25\textwidth]{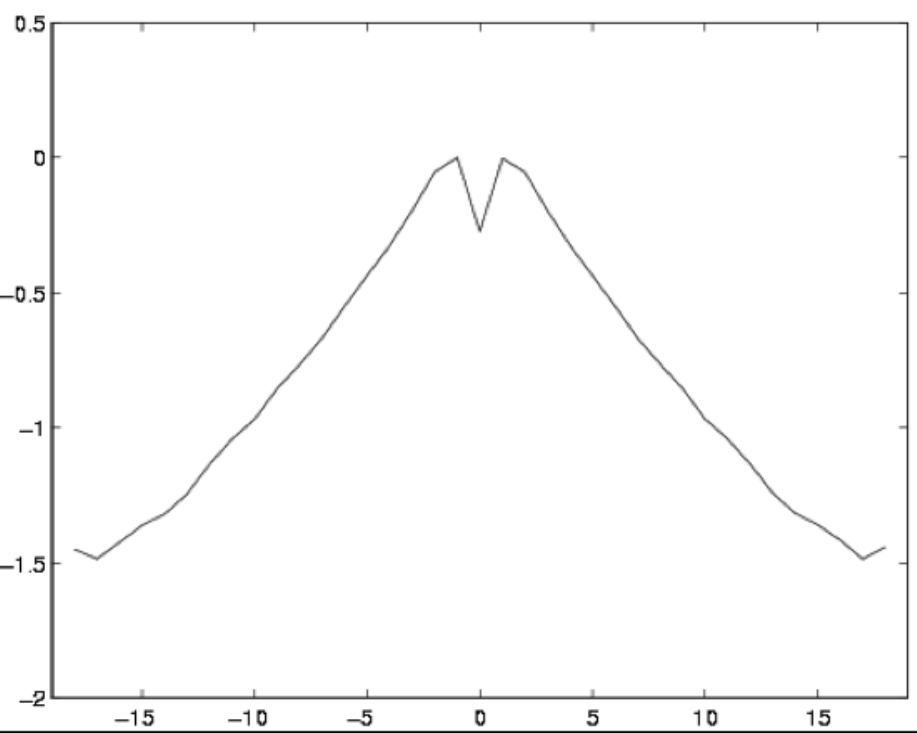}
\includegraphics[width=0.25\textwidth]{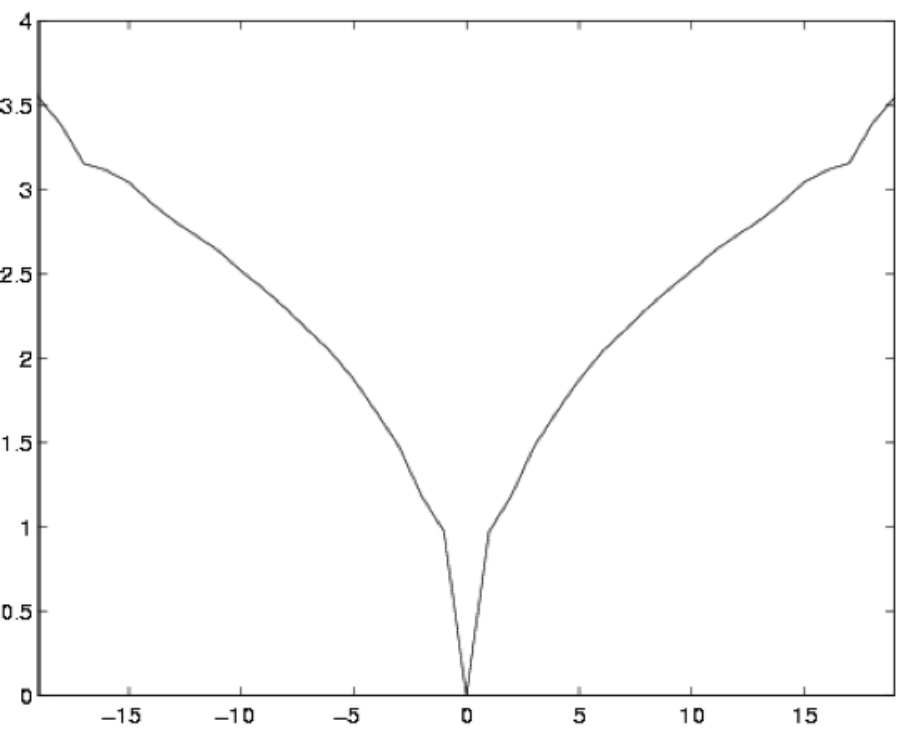}
\end{center}
		\caption{Two types of potential functions learned by \cite{zhu1997GRADE} from natural images. The function on the left encourages big filter responses and creates patterns via reaction, while the function on the right prefers small filter responses and smoothes out the synthesized image via diffusion.    }
		\label{fig:prior}
\end{figure}

One example of the descriptive model (\ref{eq:D0}) is the FRAME (Filters, Random field, And Maximum Entropy) model   \cite{zhu1997minimax, wu2000equivalence}, where $h(X)$ consists of histograms of responses from a bank of filters. In a simplified non-convolutional version, $h(X)^\T \theta = f(W X) = \sum_{k = 1}^{d} f_k(W_k X)$, where $W$ is a $d \times p$ matrix, and $W_k$ is the $k$-th row of $W$. $WX$ consists of the $d$ filter responses with each row of $W$ being a linear filter. $(f_k, k = 1, ..., d)$ are $d$ one-dimensional potential functions applied respectively to the $d$ elements of $WX$.  In the FRAME model, the rows of $W$ are a bank of Gabor wavelets or filters \cite{daugman1985uncertainty}. Given the filters, \cite{zhu1997GRADE} learned the potential functions $(- f_k, k = 1, ..., d)$ from natural images. There are two types of potential functions as shown in Figure \ref{fig:prior} taken from  \cite{zhu1997GRADE}. The function on the left encourages big filter responses while the function on the right prefers small filter responses. \cite{zhu1997GRADE} used the Langevin dynamics to sample from the learned model. The gradient descent component of the dynamics is interpreted as the Gibbs Reaction And Diffusion Equations (GRADE), where the function on the left of Figure \ref{fig:prior} is for reaction to create patterns, while  the function on the right is for diffusion to smooth out the synthesized image.

\begin{figure}[h]
\begin{center}
\includegraphics[width=0.6\textwidth]{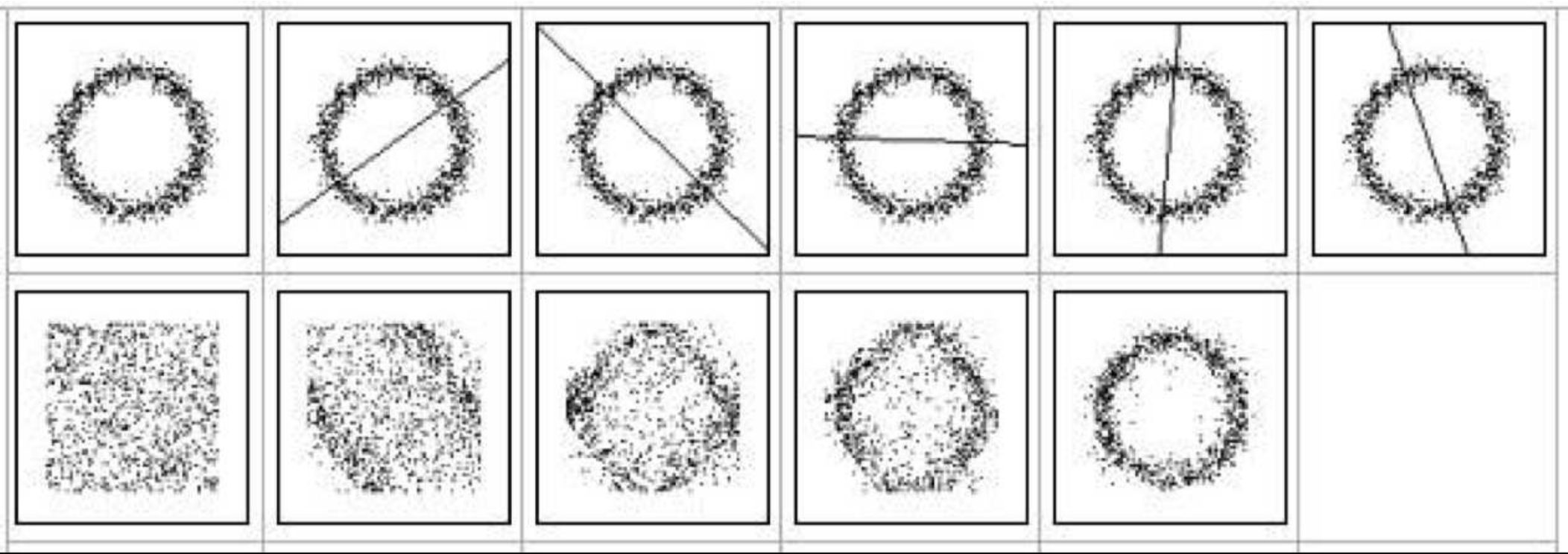}
\end{center}
		\caption{Learning a two dimensional FRAME model by sequentially adding rows to $W$ \cite{liu2001learning}. Each row of $W$ corresponds to a projection of the data. Each step finds the projection that reveals the maximum difference between the observed data and the synthesized data generated by the current model.  }
		\label{fig:2dframe}
\end{figure}

In \cite{liu2001learning}, the authors illustrated the idea of learning $W = (W_k, k = 1, ..., d)$ by a two-dimensional example. Each step of the learning algorithm adds a row $W_k$ to the current $W$. Each row corresponds to a projection of $X$. Each step finds a direction of the projection that reveals the maximum difference between the data points sampled from the current model and the observed data points. The learning algorithm then updates the model to match the marginal distributions of the model and the data in that direction.   After a few steps, the distribution of the learned model is almost the same as the distribution of the observed data. See Figure \ref{fig:2dframe} for an illustration. By assuming a parametric differentiable form for $f_k()$, $W$ can be learned by gradient descent. Such models are called product of experts \cite{Hinton2002a, teh2003energy} or field of experts \cite{roth2005fields}. 

\begin{figure}[h]
\begin{center}
\includegraphics[width=0.25\textwidth]{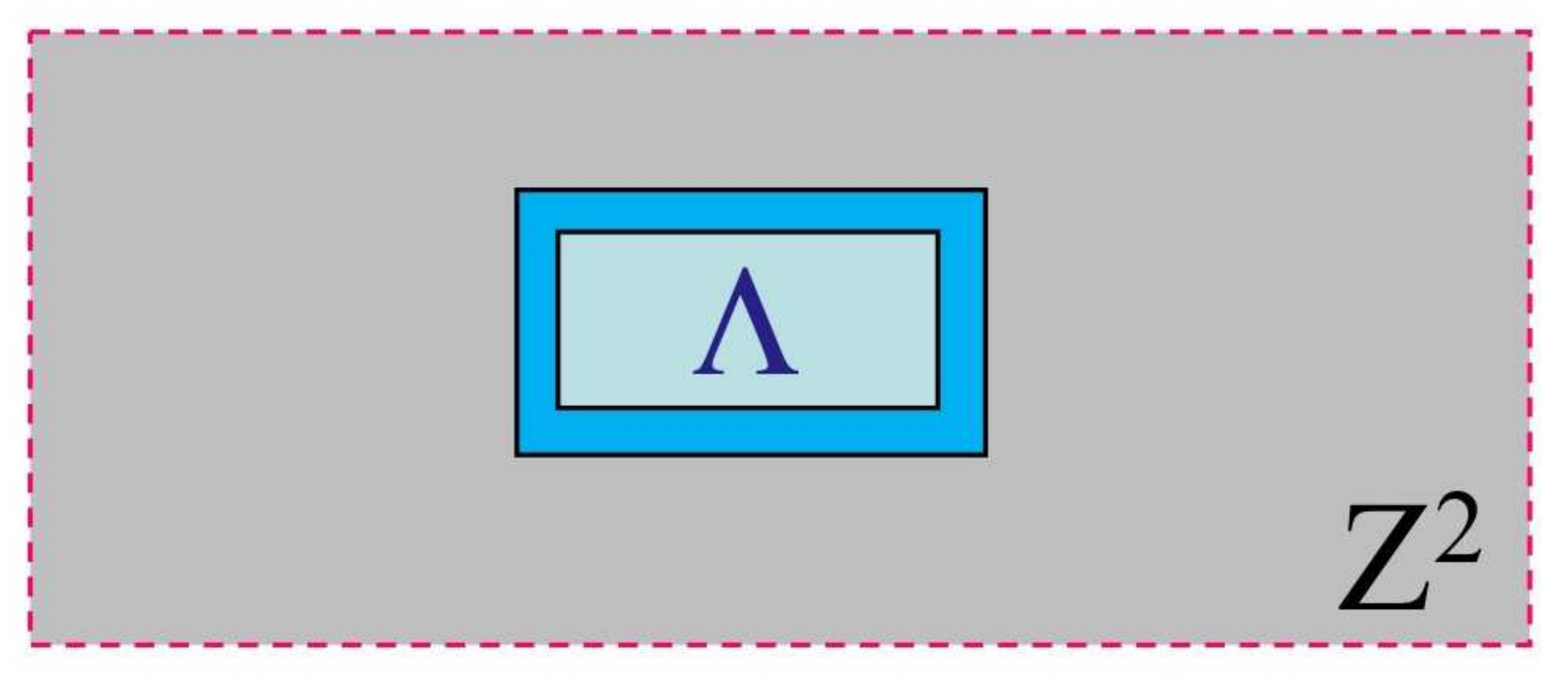}
\end{center}
		\caption{Under the uniform distribution of images defined on a large lattice (that goes to ${\bf Z}^2$) where the images share the same marginal histograms of filter responses, the conditional distribution of the local image patch given its boundary (in blue color)  follows the FRAME model.  }
		\label{fig:equivalence}
\end{figure}
The FRAME model is convolutional, where the rows of $W$ can be partitioned into different groups, and the rows in the same group are spatially translated versions of each other, like wavelets. They are called filters or kernels. The model can be justified by a uniform distribution over  the images defined on a large lattice that goes to ${\bf Z}^2$, where all the images share the same marginal histograms of filter responses. Under such a uniform distribution, the distribution of the local image patch defined on a local lattice $\Lambda$ conditional on its boundary (illustrated by the blue color, including all the pixels outside $\Lambda$ that can be covered by the same filters as the pixels within $\Lambda$) follows the FRAME model  \cite{wu2000equivalence}. See Figure \ref{fig:equivalence} for an illustration.

\subsection{Generative models} 

This subsection reviews various versions of the linear generative models. These models share the same linear form, but they differ in terms of the prior assumptions of the latent factors or coefficients. 

Like the descriptive models, the generative models can be learned in the unsupervised setting, where $Y_i$ are not observed, as illustrated below: 
 \begin{table}[h]
\centering
\begin{tabular}{|c|c|c|c|}
\hline
  & input & hidden  & output \\
\hline\hline
 1   & $X_1^\T$ & $h_1^\T$         & ?  \\
 2   & $X_2^\T$ & $h_2^\T$         &  ?\\
...  &      &           &\\
$n$ &$X_n^\T$  &  $h_n^\T$        &  ?\\
\hline
\end{tabular}
\end{table}

In a generative model, the vector $h_i$ is not a vector of features extracted from the signal $X_i$. $h_i$ is a vector of hidden variables that is used to generate $X_i$, as illustrated by the following diagram: 
\begin{eqnarray}
\begin{array}[c]{cc}
{\rm hidden}: & h_i\\
& \downarrow\\
{\rm input}: & X_i
\end{array}  
\label{eq:d2}
\end{eqnarray}
The components of the $d$-dimensional $h_i$ are variably called  factors, sources, components or causes. 

{\em Auto-encoder}: $h_i$ is also called a code in the auto-encoder illustrated by the following diagram: 
\begin{eqnarray}
\begin{array}[c]{cc}
{\rm code}: & h_i\\
& \uparrow\downarrow\\
{\rm input}: & X_i
\end{array}  
\end{eqnarray}
The direction from $h_i$ to $X_i$ is called the decoder, and the direction from $X_i$ to $h_i$ is called the encoder. The decoder corresponds to the  generative model in (\ref{eq:d2}), while the encoder can be considered the inference model. 

{\em Distributed representation and disentanglement}: $h_i = (h_{ik}, k = 1, ..., d)$ is called a distributed representation of $X_i$. Usually the components of $h_i$, $(h_{ik}, k  = 1, ..., d)$, are assumed to be independent, and $(h_{ik})$ are said to disentangle the variations in $X_i$. 

{\em Embedding}: $h_i$ can also be considered the coordinates of $X_i$, if we embed $X_i$ into a low-dimensional space, as illustrated by the following diagram: 
 \begin{eqnarray}
\begin{array}[c]{c}
\leftarrow  h_i  \rightarrow\\
\mid \\
\leftarrow X_i  \rightarrow
\end{array}  
\end{eqnarray}
In the training data, we find a $h_i$ for each $X_i$,  so that $\{h_i, i = 1, ..., n\}$ preserve the relative relations between $\{X_i, i = 1, ..., n\}$. The prototype example of embedding is multi-dimensional scaling, where we want to preserve the Euclidean distances between the examples. A more recent example of embedding is local linear embedding \cite{roweis2000nonlinear}. In the embedding framework, there are no explicit encoder and decoder.

{\em  Linear generative model}: The linear form of the generative model is as follows: 
\begin{eqnarray}
X_i = W h_i + \epsilon_i,  
\end{eqnarray}
for $i = 1, ..., n$, where $W$ is a $p \times d$ dimensional matrix ($p$ is the dimensionality of $X_i$ and $d$ is the dimensionality of $h_i$), and $\epsilon_i$ is a $p$-dimensional residual vector. 
The following are the interpretations of $W$: 

(1) Loading matrix:  Let $W = (w_{jk})_{p \times d}$. $x_{ij} \approx \sum_{k=1}^{d} w_{jk} h_{ik}$, i.e., each component of $X_i$, $x_{ij}$, is a linear combination of the latent factors. $w_{jk}$ is the loading weight of factor $k$ on variable $j$. 

(2) Basis vectors: Let $W = (W_k, k = 1, ..., d)$, where $W_k$ is the $k$-th column of $W$. $X_i \approx \sum_{k=1}^{d} h_{ik} W_k$, i.e., $X_i$ is a linear superposition of the basis vectors $(W_k)$, where $h_{ik}$ are the coefficients. 

(3) Matrix factorization: $(X_1, ..., X_n) \approx W (h_1, ..., h_n)$, where the $p \times n$ matrix $(X_1, ..., X_n)$ is factorized into the $p \times d$ matrix $W$ and the $d \times n$ matrix $(h_1, ..., h_n)$. 

The following are some of the commonly assumed prior distributions or constraints on $h_i$.

{\em Factor analysis} \cite{rubin1982algorithms}:  $h_i \sim {\rm N}(0, I_d)$, $X_i = W h_i + \epsilon_i$, $\epsilon_i \sim {\rm N}(0, \sigma^2 I_p)$, and $\epsilon_i$ is independent of $h_i$. The dimensionality of $h_i$, which is $d$, is smaller than the dimensionality of $X_i$, which is $p$. The factor analysis is very similar to the principal component analysis (PCA), which is a popular tool for dimension reduction. The difference is that in factor analysis, the column vectors of $W$ do not need to be orthogonal to each other. 

The factor analysis model originated from psychology, where $X_i$ consists of  the test scores of student $i$ on $p$  subjects. $h_i$ consists of the verbal intelligence and the analytical intelligence of student $i$ ($d = 2$). Another example is the decathlon competition, where $X_i$ consists of the scores of athlete $i$ on $p = 10$ sports, and $h_i$ consists of athlete $i$'s speed, strength and endurance ($d = 3$).

{\em Independent component analysis} \cite{hyvarinen2004independent}: In ICA, for $h_i = (h_{ik}, k = 1, ..., d)$, $h_{ik} \sim P_k$ independently, and $P_k$ are assumed to be heavy-tailed  distributions. For analytical tractability, ICA assumes that $d = p$, and $\epsilon_i = 0$. Hence $X_i = W h_i$, where $W$ is a squared matrix assumed to be invertible, so that  $h_i = A X_i$, where $A = W^{-1}$. Let $P(h_i) = \prod_{k=1}^{d} P_k(h_{ik})$. The marginal distribution of $X_i$ has a closed form $X_i \sim P(A X) |{\rm det}(A)|$. The ICA model is both a generative model and a descriptive model. 

{\em Sparse coding}   \cite{olshausen1997sparse}: In the sparse coding model, the dimensionality of $h_i$, which is $d$, is bigger than the dimensionality of $X_i$, which is $p$. However, $h_i = (h_{ik}, k = 1, ..., d)$ is a sparse vector, meaning that only a small number of $h_{ik}$ are non-zero, although for different example $i$, the non-zero elements in $h_i$ can be different. Thus unlike PCA,  sparse coding provides adaptive dimension reduction. $W = (W_k, k = 1, ..., d)$ is called a redundant dictionary because $d > p$, and each $W_k$ is a basis vector or a ``word'' in the dictionary.  Each $X_i \approx W h_i  = \sum_{k=1}^{d} h_{ik} W_k $ is explained by a small number of $W_k$ selected from the dictionary, depending on which $h_{ik}$ are non-zero. The inference of the sparse vector $h_i$ can be accomplished by Lasso or basis pursuit  \cite{tibshirani1996regression, chen1998atomic} that minimizes $\sum_{i=1}^{n} \left[\|X_i - W h_i\|^2 + \lambda \|h_i\|_{\ell_1}\right]$, which imposes the sparsity inducing  $\ell_1$ regularization   on $h_i$ with a regularization parameter $\lambda$. 

\begin{figure}[h]
\begin{center}
\includegraphics[width=0.4\textwidth]{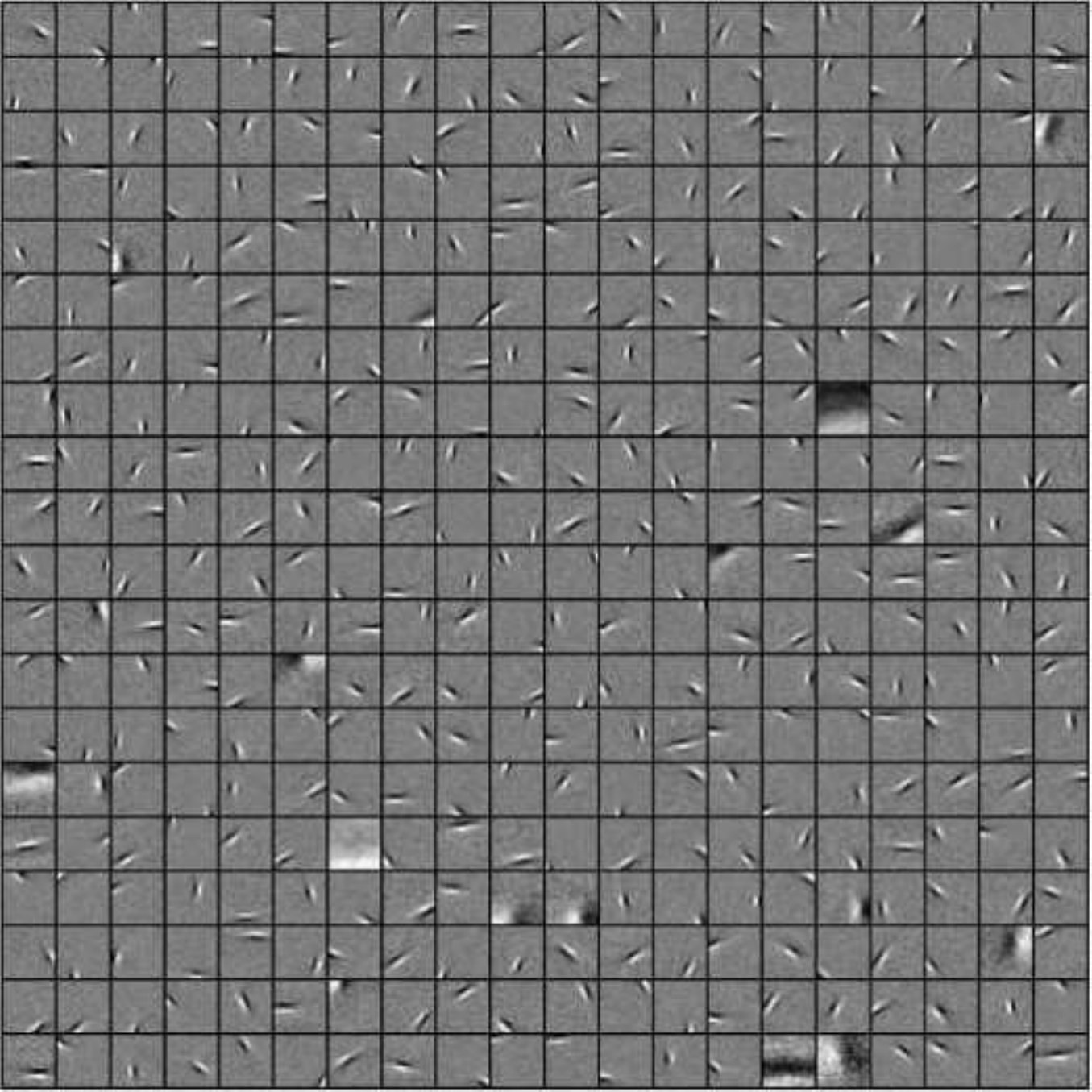}
\end{center}
		\caption{Sparse coding \cite{olshausen1997sparse}: learned basis vectors from natural image patches. Each image patch in the picture is a column vector of $W$. }
		\label{fig:sparse}
\end{figure}

A Bayesian probabilistic formulation is to assume a spike-slab prior: $h_{ik} \sim \rho \delta_0 + (1-\rho) {\rm N}(0, \tau^2)$ with a small $1-\rho$, which is the probability that $h_{ik}$ is non-zero. 

Figure \ref{fig:sparse} displays a sparse code learned from a training set of natural image patches of size $12 \times 12$    \cite{olshausen1997sparse}. Each column of $W$, $W_k$, is a basis vector that can be made into an image patch as shown in the figure. 

{\em Non-negative matrix factorization}   \cite{lee2001algorithms}: In NMF,  $h_i$ is constrained to have non-negative components, i.e., $h_{ik} \geq 0$ for all $k$. It is also called positive factor analysis   \cite{paatero1994positive}. The rationale for NMF is that the parts of a pattern should be additive and the parts should contribute positively.

{\em Matrix factorization for recommender system}  \cite{koren2009matrix}: In recommender system, $X_i = (x_{ij}, j = 1, ..., p)$ are the ratings of user $i$ on the $p$ items. For instance, in the Netflix example, there are $n$ users and $p$ movies, and $x_{ij}$ is user $i$'s rating of movie $j$. Let ${w}_j$ be the $j$-th row of matrix $W$, then  $x_{ij} = \langle {w}_j, h_i\rangle + \epsilon_{ij} $, where $h_i$ characterizes the desires of user $i$ in $d$ aspects, and ${w}_j$ characterizes the desirabilities of item $j$ in the corresponding aspects. The rating matrix $(X_i, i = 1, ..., n)$ thus admits  a rank $d$ factorization. The rating matrix is in general incomplete. However, we can still estimate $(h_i)$ and $({w}_j)$ from the observed ratings and use them to complete the rating matrix for the purpose of recommendation.

{\em Probabilistic formulation}: In the above models, there is a prior model $h_i \sim p(h)$ or a prior constraint such as $h_i$ is sparse or non-negative.  There is a linear generative model $X_i = W h_i + \epsilon_i$, with $\epsilon_i \sim {\rm N}(0, \sigma^2 I_p)$,  for $i = 1, ..., n$. This defines the conditional distribution $p(X|h; W)$. The joint distribution is $p(h) p(X|h;  W) = p(h, X| W)$. The marginal distribution is obtained by integrating out $h$: 
\begin{eqnarray}
p(X| W) = \int p(h) p(X|h; W)dh =  \int p(h, X| W) dh.
\end{eqnarray}
This integral is analytically intractable.  According to the Bayes rule, $h$ can be inferred from $X$ based on the posterior distribution, 
$p(h|X;  W) = p(h, X| W)/p(X| W)$, which is proportional to  $p(h, X| W)$ as a function of $h$. We call $p(h|X; W)$ the inference model.  

In the auto-encoder terminology, $p(h)$ and $p(X|h; W)$ define the decoder,  $p(h|X; W)$ defines the encoder. In factor analysis and independent component analysis, $h$ can be inferred in closed form. For other models, however, $h$ needs to be inferred by an iterative algorithm. 

{\em Restricted Boltzmann machine}  \cite{Hinton06}:  In RBM, unlike the above models, there is no explicit prior $p(h)$. The model is defined by the joint distribution
\begin{align}
(h_i, X_i) \sim p(h, X|  W) &= \frac{1}{Z( W)} \exp\left[  \sum_{j, k} w_{j k} x_j h_k \right]  \\
&= \frac{1}{Z( W)} \exp\left[  X^\T W h \right].
\end{align}
The above model assumes that both $h_i$ and $X_i$ are binary. Under the above model, both the generative distribution $p(X|h; W)$ and the inference distribution $p(h|X; W)$ are independent logistic regressions. We may modify the model slightly to make $X$ continuous, so that in the modified model, the generative distribution $p(X|h; W)$ is normal linear regression: $X = Wh + \epsilon$, with  $\epsilon \sim {\rm N}(0, \sigma^2 I_p)$. The  inference model, $p(h|X; W)$,  is logistic regression,  $h \sim {\rm logistic}(W^\T X)$, i.e., $\Pr(h_k = 1|X; W) = {\rm sigmoid}(\sum_{j=1}^{p} w_{j k} x_j)$, where ${\rm sigmoid}(r) = 1/(1+e^{-r})$. 

If we sum out $h$, the marginal distribution $p(X|W) = \sum_h p(h, X|W)$ can be obtained in closed form, and $p(X|W)$ is a descriptive model. 

{\em   RBM-like auto-encoder}  \cite{vincent2008extracting, Bengio-et-al-2015-Book}: The RBM leads to the following auto-encoder:  Encoder: $ h_k = {\rm sigmoid}(\sum_{j=1}^{p} w_{jk} x_j)$, i.e., $h = {\rm sigmoid}(W^\T X)$; Decoder: $X = Wh$. 

Like the descriptive model, the generative model can also be learned by maximum likelihood. However, unlike the ``analysis by synthesis'' scheme for learning the descriptive model, the learning algorithm for generative model follows an ``analysis by inference'' scheme. Within each iteration of the learning algorithm, there is an inner loop for inferring $h_i$ for each $X_i$. The most rigorous inference method is to sample $h_i$ from the posterior distribution or the inference distribution $p(h_i|X_i; W)$. After inferring $h_i$ for each $X_i$, we can then update the model parameters by analyzing the ``imputed'' dataset $\{(h_i; X_i)\}$,  by fitting the generative distribution $p(X|h; W)$. The EM algorithm \cite{dempster1977maximum} is an example of this learning scheme, where the inference step is to compute expectation with respect to $p(h_i|X_i; W)$. From a Monte Carlo perspective, it means we make multiple imputations \cite{rubin2004multiple} or make multiple guesses of $h_i$ to account for the uncertainties in $p(h_i|X_i; W)$. Then we analyze the multiply imputed dataset to update the model parameters. 

\section{Interactions between different families}

\subsection{Discriminative learning of descriptive model} 

This subsection shows that the descriptive model can be learned discriminatively. 

The descriptive model (\ref{eq:D0}) can be connected to the discriminative model (\ref{eq:logistic})  if we treat $p_0(X)$ as the distribution of the negative examples, and $p_\theta(X)$ as the distribution of the positive examples. Suppose we generate the data as follows: $Y_i \sim {\rm Bernoulli}(\rho)$, i.e., $\Pr(Y_i = 1) = \rho$, which is the prior probability of positive examples. $[X_i \mid Y_i = 1] \sim p_\theta(X)$, and $[X_i \mid Y_i = 0] \sim p_0(X)$. According to the Bayes rule
\begin{eqnarray}
    \log \frac{\Pr(Y_i = 1\mid X_i)}{\Pr(Y_i = 0 \mid X_i)} = h(X_i)^\T \theta - \log Z(\theta)  +\log [\rho/(1-\rho)],  \label{eq:Tu}
\end{eqnarray} 
which corresponds to (\ref{eq:logistic}) with $b = - \log Z(\theta) + \log[\rho/(1-\rho)]$. 

\begin{figure}[h]
\begin{center}
\includegraphics[width=0.7\textwidth]{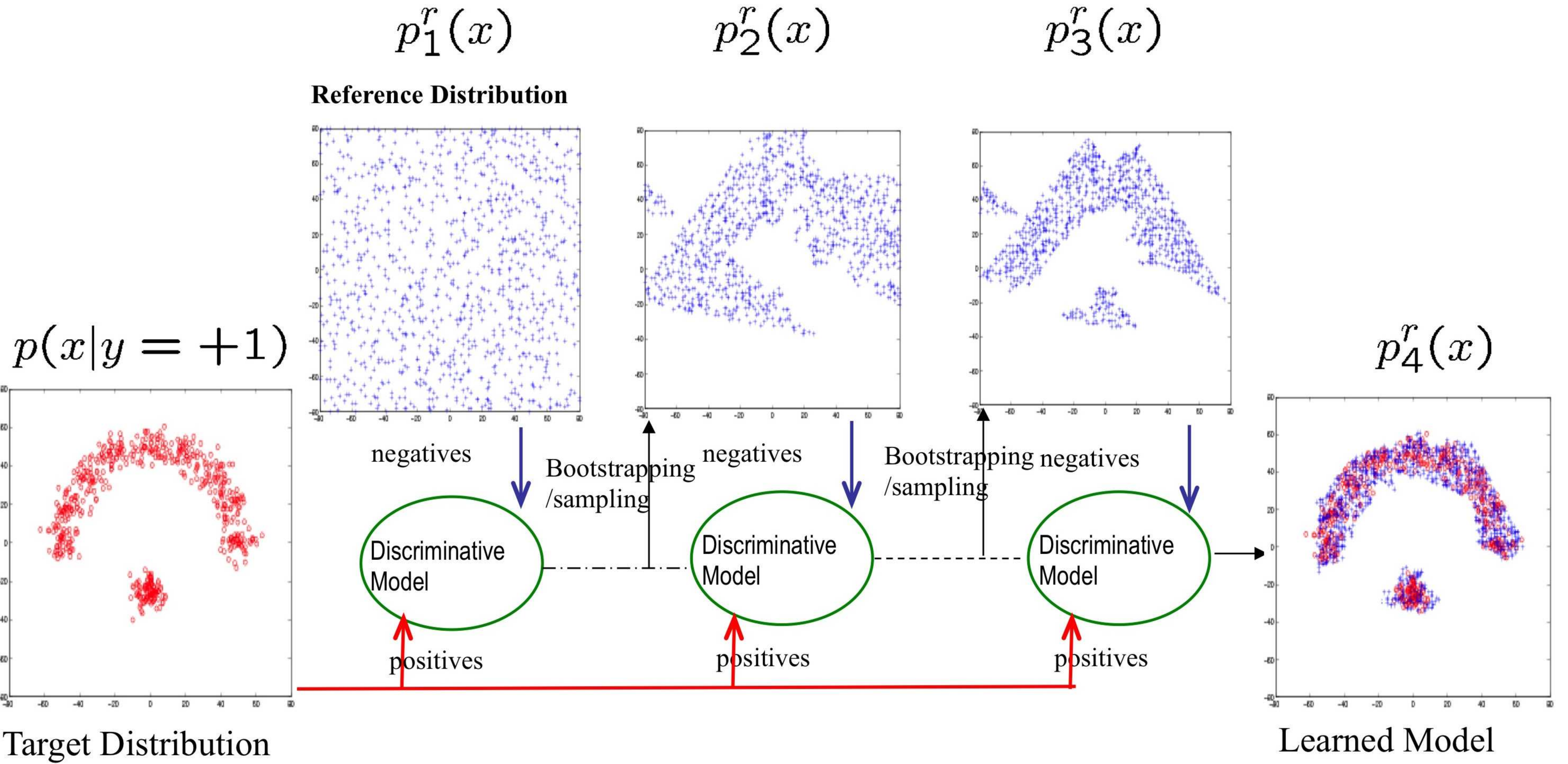}
\end{center}
		\caption{Discriminative learning of the descriptive model \cite{tu2007learning}. By fitting a logistic regression to discriminate between the observed examples and the synthesized examples generated by the current model, we can modify the current model according to the fitted logistic regression, so that the modified model gets closer to the distribution of the observed data. }
		\label{fig:Tu0}
\end{figure}

Tu \cite{tu2007learning} made use of this fact to estimate $p_\theta$ discriminatively. The learning algorithm starts from $p_0$. At step $t$, we let the current $p_t$ serve as the negative distribution, and generate synthesized examples from $p_t$. Then we fit a logistic regression by treating the examples generated by $p_t$ as the negative examples, and the observed examples as the positive examples. Let $\theta$ be the estimated parameter of this logistic regression. We then let $p_{t+1}(X) = \exp(h(X)^\T \theta) p_t(X)/Z$.  See \cite{tu2007learning} for an analysis of the convergence of the learning algorithm. 

Figure \ref{fig:Tu0} taken from  \cite{tu2007learning}  illustrates the learning process by starting from the uniform $p_0$. By iteratively fitting the logistic regression and modifying the distribution, the learned distribution converges to the true distribution. 

\subsection{Integration of descriptive and generative models} 

Natural images contain both stochastic textures and geometric objects (as well as their parts). The stochastic textures can be described by some feature statistics pooled over the spatial domain, while the geometric objects can be represented by image primitives or textons.
\begin{figure}[h]
\begin{center}
\includegraphics[width=0.2\textwidth]{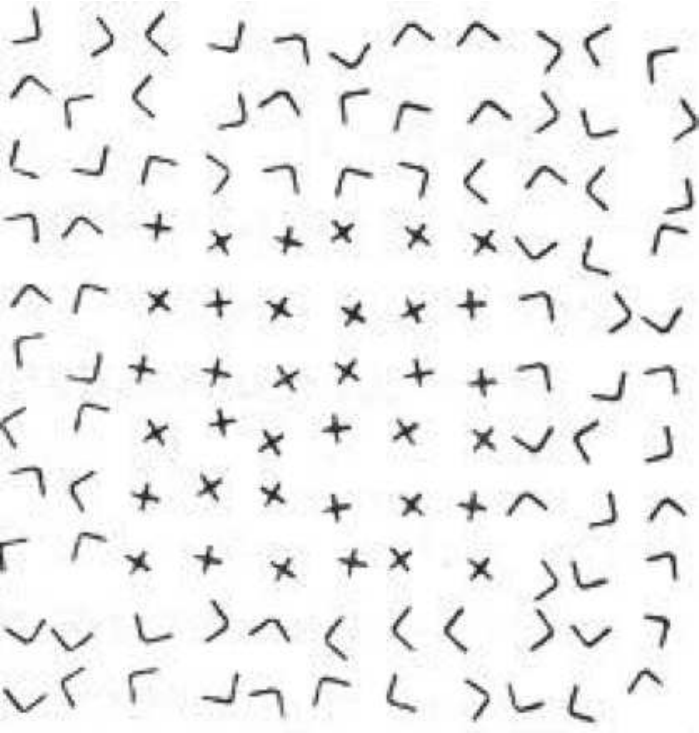}
\includegraphics[width=0.2\textwidth]{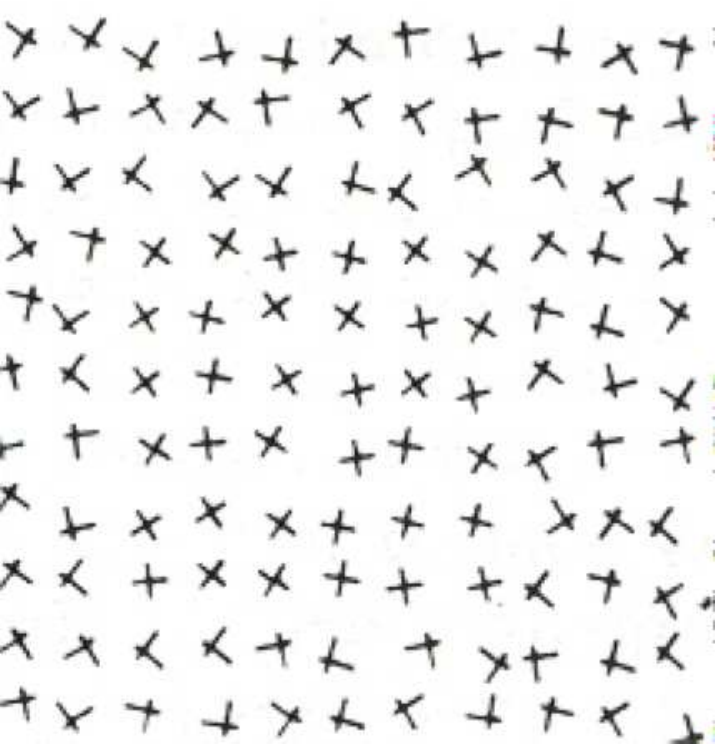}
\end{center}
		\caption{Pre-attentive vision is sensitive to local patterns called textons.}
		\label{fig:textons}
\end{figure}
The psychophysicist  Julesz  \cite{julesz1981textons} studied both texture statistics and textons. He conjectured that pre-attentive human vision is sensitive to local patterns called textons.  Figure \ref{fig:textons} illustrates the basic idea. 
\begin{figure}[h]
\begin{center}
\includegraphics[width=0.4\textwidth]{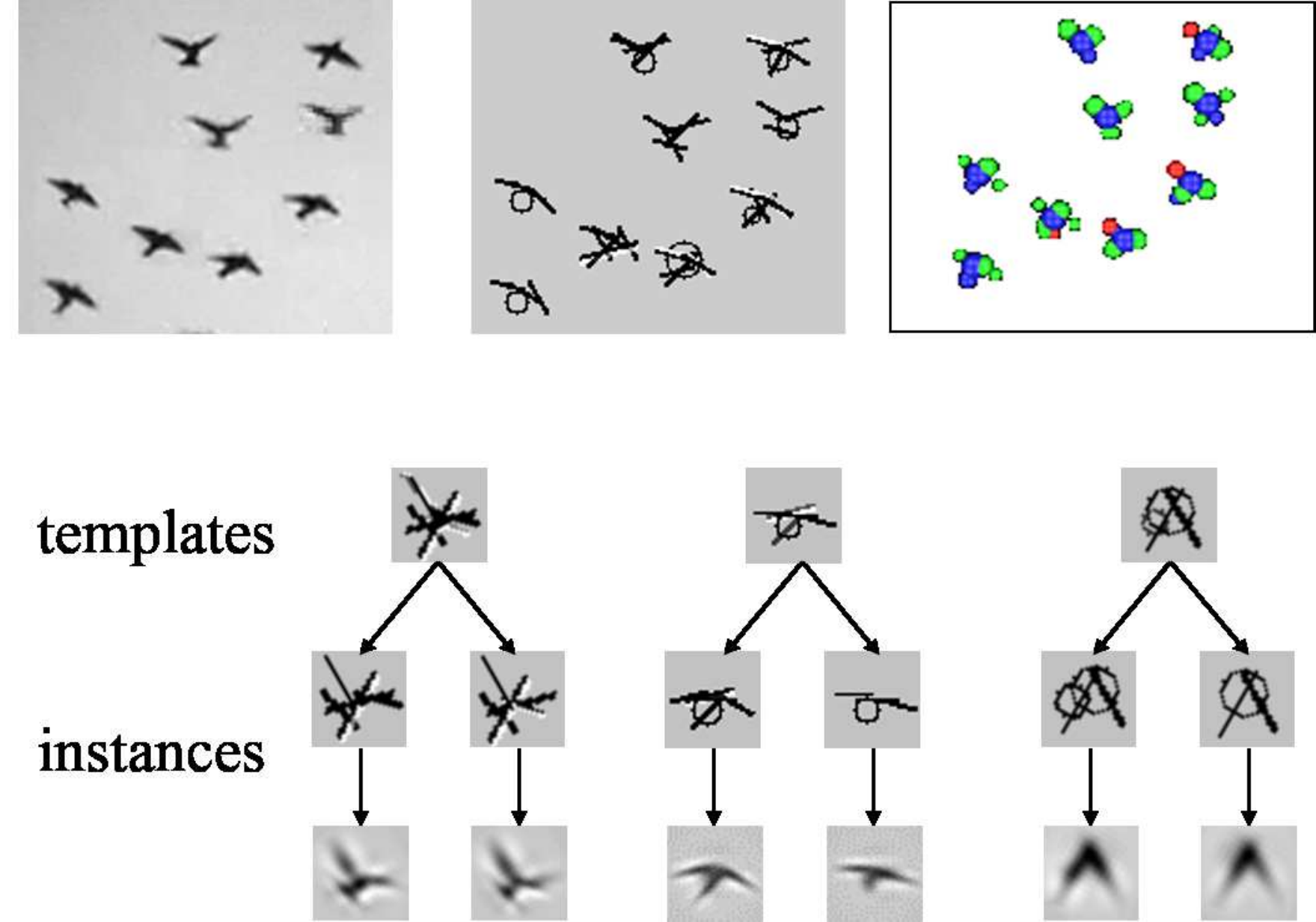}
\end{center}
		\caption{A model of textons \cite{zhu2005textons}, where each texton is a composition of a small number of wavelets. }
		\label{fig:texton}
\end{figure}
Inspired by Julesz's  work, in \cite{zhu2005textons}, the authors proposed a generative model for textons, where each texton is a composition of a small number of wavelets, as illustrated by Figure \ref{fig:texton}. The model is a generalization of the sparse coding model of    \cite{olshausen1997sparse}.

\begin{figure}[h]
\begin{center}
\includegraphics[width=0.7\textwidth]{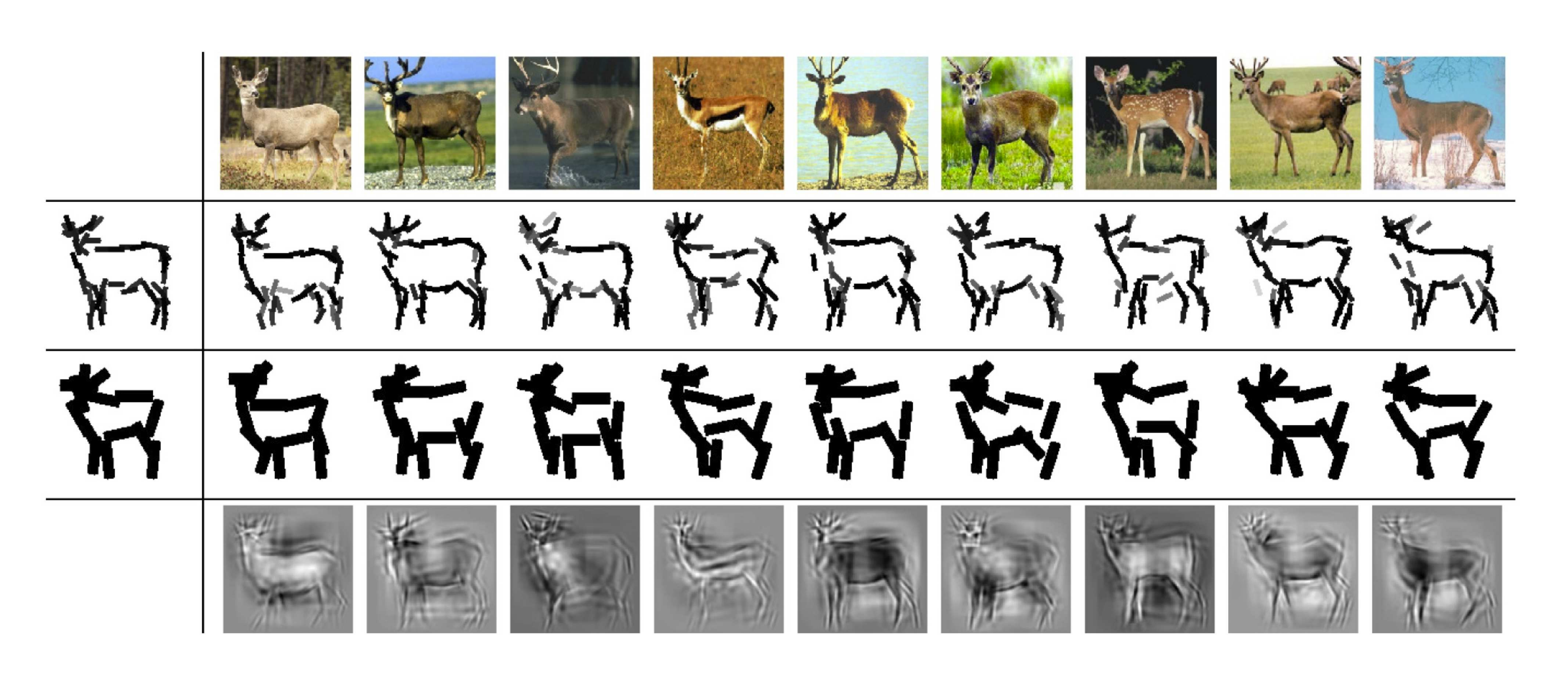}
\end{center}
		\caption{Active basis model \cite{AB, hong2013unsupervised}: each active basis template is a composition of wavelets selected from a dictionary, and the wavelets are allowed to shift their locations and orientations to account for shape deformation. Here each wavelet is illustrated by a bar. The templates are learned at two different scales. The observed images can be reconstructed by the wavelets of the deformed templates.  }
		\label{fig:AB}
\end{figure} 
Building on the texton model of \cite{zhu2005textons},  \cite{AB, hong2013unsupervised} proposed an active basis model, where each model is a composition of wavelets selected from a dictionary, and the wavelets are allowed to shift their locations and orientations to account for shape deformation. See Figure \ref{fig:AB} for an illustration. 

\begin{figure}[h]
\begin{center}
\includegraphics[width=0.6\textwidth]{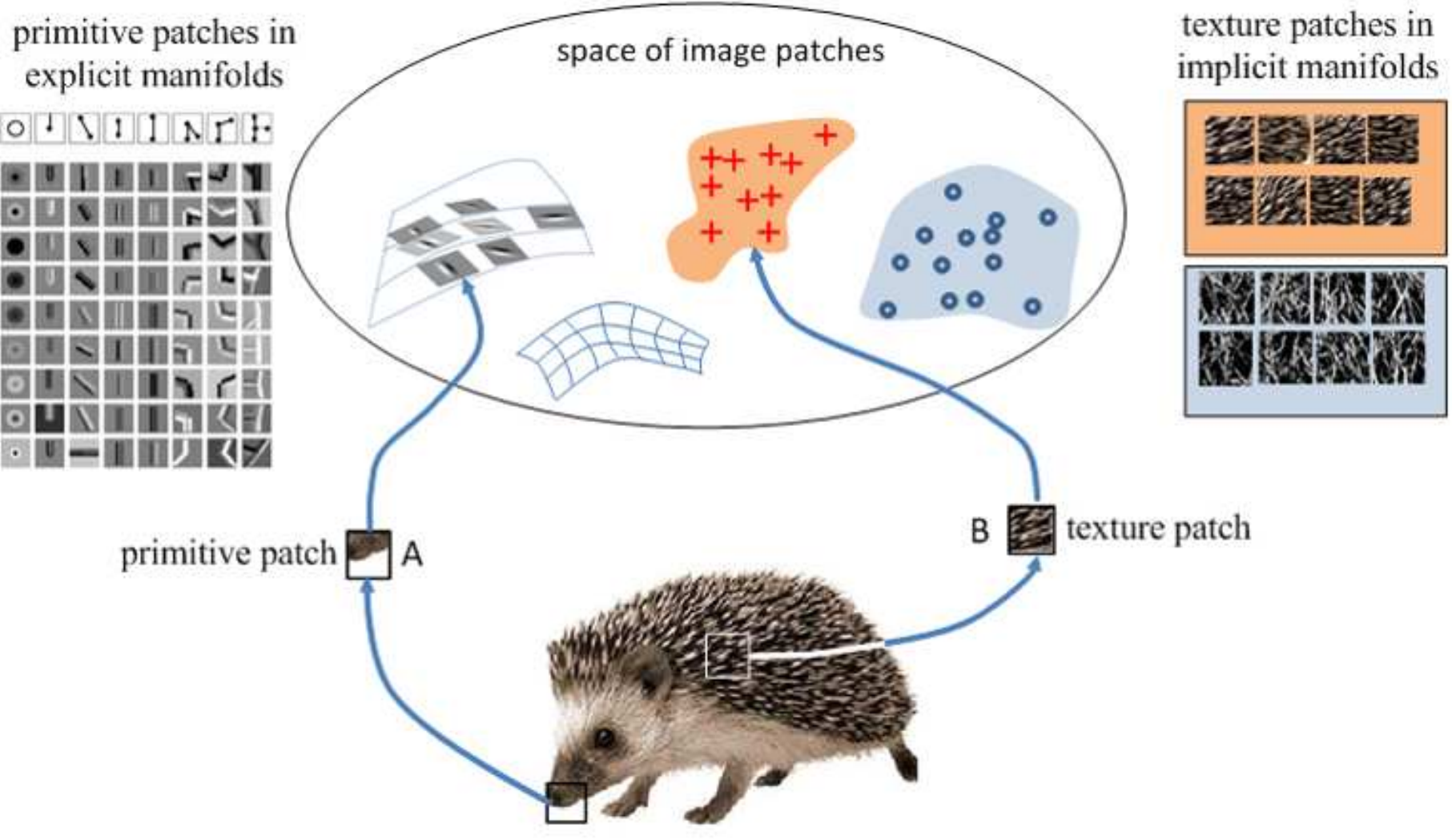}
\end{center}
		\caption{Hybrid image template \cite{hit}: integrating generative model for shape template and the descriptive model for texture. }
		\label{fig:hit}
\end{figure}

The texton model and the active basis model are generative models. However, they do not account for stochastic texture patterns. \cite{hit} proposed to integrate the generative model for shape templates and the descriptive model for stochastic textures, as illustrated by Figure \ref{fig:hit}. A similar model was developed by \cite{guo2007primal} to model both the geometric structures and stochastic textures by generative models and descriptive models respectively. 

In \cite{ guo2003modeling},  the authors provided another integration of the generative model and the descriptive model, where the lowest layer is a generative model such as the  wavelet sparse coding model \cite{olshausen1997sparse}, but the spatial distribution of the wavelets is governed by a descriptive model.

\subsection{DDMCMC: integration of discriminative and generative models} 

\begin{figure}[h]
\begin{center}
\includegraphics[width=0.4\textwidth]{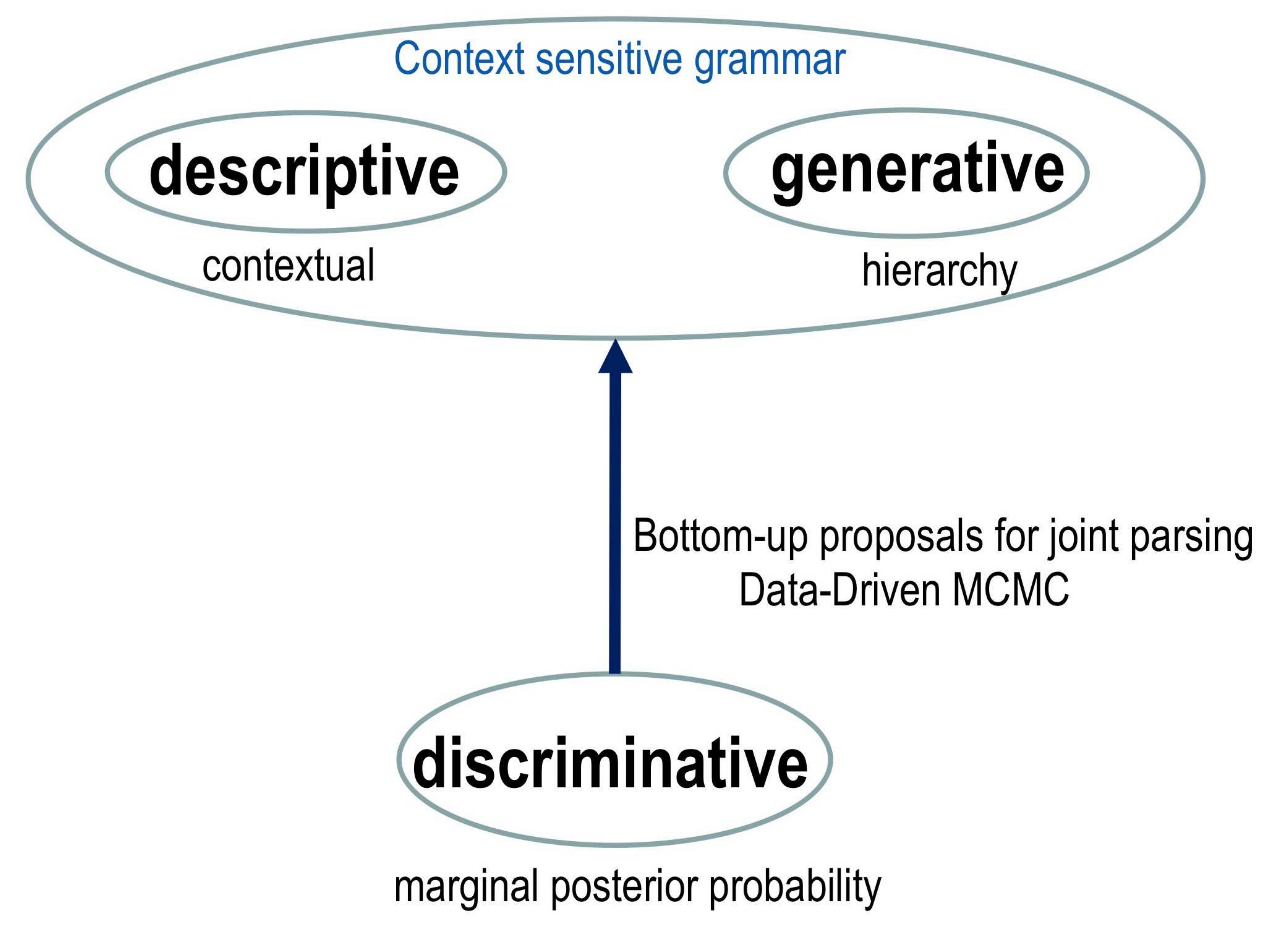}
\end{center}
		\caption{Data-driven MCMC: when fitting the generative models and descriptive models using MCMC, the discriminative models can be employed to provide proposals for MCMC transitions.  }
		\label{fig:DD}
\end{figure}

In \cite{tu2002image, tu2006parsing},  the authors proposed a data-driven MCMC method for fitting the generative models as well as the descriptive models to the data. Fitting such models usually require time-consuming MCMC. In \cite{tu2002image, tu2006parsing}, the authors proposed to speed up the MCMC by using the discriminative models to provide the proposals for the Metropolis-Hastings algorithm. See Figure \ref{fig:DD} for an illustration. 

\section{Hierarchical forms of the three families}

This section presents the hierarchical non-linear forms of the three families of models, where the non-linear mappings are parametrized by neural networks, in particular, the convolutional neural networks. 

\subsection{Recent developments}

During the past few years, deep convolutional neural networks (CNNs or ConvNets) \cite{lecun1998gradient, krizhevsky2012imagenet} and recurrent neural networks (RNNs) \cite{hochreiter1997long} have transformed the fields of computer vision, speech recognition, natural language processing, and other fields in artificial intelligence (AI). Even though these neural networks were invented decades ago, their potentials were realized only recently mainly because of the following two factors. (1) The availability of big training datasets such as Imagenet \cite{deng2009imagenet}. (2) The improvement in computing power, mainly brought by the graphical processing units (GPUs). These two factors, together with some recent clever tweaks and inventions such as rectified linear units \cite{krizhevsky2012imagenet}, batch normalization \cite{ioffe2015batch}, residual networks \cite{he2016deep}, etc., enable the training of very deep networks (e.g., 152 layers with 60 million parameters in a residual network for object recognition \cite{he2016deep}) that achieve impressive performances on many tasks in AI (a recent example being Alpha Go Zero \cite{silver2017mastering}). 

One key reason for the successes of deep neural networks is that they are universal and flexible function approximators. For instance, a feedforward neural network with rectified linear units is a piecewise linear function with recursively partitioned linear pieces that can approximate any continuous non-linear mapping \cite{montufar2014number}. However, this does not fully explain the ``unreasonable effectiveness'' of deep neural networks. The stochastic gradient descent algorithm that is commonly employed to train the neural networks is expected to approach only a local minimum of the highly non-convex objective function. However, for large and deep networks, it appears that most of the local modes are equally good \cite{choromanska2014loss} in terms of training and testing errors, and the apparent vices of local modes and stochasticity in the mini-batch on-line training algorithm actually turn out to be big virtues in that they seem to prevent overfitting and lead to good generalization \cite{choromanska2014loss}. 

The approximation capacities of the deep neural networks have been extensively exploited in supervised learning (such as  classification networks and regression networks) and reinforcement learning (such as policy networks and value networks). They have also proven to be useful for unsupervised learning and generative modeling, where the goal is to learn features or hidden variables from the observed signals  without external guidance such as class labels or rewards. The unsupervised learning is often accomplished in the context of a generative model (or an auto-encoder), which explains or characterizes the observed examples. 

 \subsection{Discriminative models by convolutional neural networks} \label{sect:ConvNets}

 The neural networks in general and the convolutional neural networks (ConvNet or CNN)  in particular were initially designed for discriminative models. Let $X$ be the $p$-dimensional input vector, and $Y$ be the output. We want to predict $Y$ by $\hat{Y}$ which is a non-linear transformation of $X$: $\hat{Y} = f_\theta(X)$, where $f$ is parametrized by parameters $\theta$. In a feedforward neural network, $f$ is a composition of $L$ layers of liner mappings followed by coordinate-wise non-linear rectifications, as illustrated by the following diagram: 
 \begin{eqnarray}
 X \rightarrow h^{(1)} \rightarrow ... h^{(l-1)} \rightarrow h^{(l)} \rightarrow ... \rightarrow h^{(L)} \rightarrow \hat{Y},
 \end{eqnarray} 
 where $h^{(l)}$ is a $d^{(l)}$ dimensional vector which is defined recursively by 
\begin{eqnarray}
h^{(l)} = f^{(l)}(W^{(l)} h^{(l-1)} + b^{(l)}),  \label{eq:recursion}
\end{eqnarray}
for $l  = 1, ..., L$. We may treat $X$ as $h^{(0)}$,  and $\hat{Y}$ as $h^{(L+1)}$ and $\theta = (W^{(l)}, b^{(l)}, l =1, ..., L+1)$. $W^{(l)}$ is the weight matrix and $b^{(l)}$ is the bias or intercept vector at layer $l$. $f^{(l)}$ is coordinate-wise transformation, i.e., for a vector $v = (v_1, ..., v_d)^{\top}$, $f^{(l)}(v) = (f^{(l)}(v_1), ..., f^{(l)}(v_d))^\top$.  

Compared to the discriminative models in the previous section, we now have multiple layers of features $(h^{(l)}, l = 1, ..., L)$. They are recursively defined via (\ref{eq:recursion}), and they are to be learned from the training data instead of being designed. 

For classification, suppose there are $K$ categories, the conditional probability of category $k$ given input $X$ is given by the following soft-max probability: 
\begin{eqnarray} 
   \Pr(Y = k \mid X) = \frac{\exp(f_{\theta_k}(X))}{\sum_{k=1}^{K} \exp(f_{\theta_k}(X))}, 
\end{eqnarray}
where $f_{\theta_k}(X)$ is the score for category $k$. We may take $f_{\theta_k}(X) = h^{(L)\T} W_k^{(L+1)} + b_k^{(L+1)}$. This final classification layer is usually called the soft-max layer.

The most commonly used non-linear rectification in modern neural nets is the Rectified Linear Unit (ReLU)   \cite{krizhevsky2012imagenet}: $f^{(l)}(a) = \max(0, a)$. The resulting function $f_\theta(X)$ can be considered a  multi-dimensional linear spline, i.e., a piecewise linear function. Recall a one-dimensional linear spline is of the form $f(x) = b + \sum_{k=1}^{d} w_k \max(0, x - a_k)$, where $a_k$ are the knots. At each knot $a_k$, the linear spline takes a turn and changes its slope by $w_k$. With enough knots, $f(x)$ can approximate any non-linear continuous function. We can view this $f(x)$ as a simplified two-layer network, with $h_k = \max(0, x-a_k)$. The basis function $\max(0, x-a_k)$ is two-piece linear function with a bending at $a_k$.  For multi-dimensional input $X$, a two-layer network with one-dimensional output is of the following form $f(X) = b^{(2)} + \sum_{k=1}^{d} W^{(2)}_k h^{(1)}_k$, where $h^{(1)}_k =  \max(0, W^{(1)}_k X + b^{(1)}_k)$, and $W^{(1)}_k$ is the $k$-th row of $W^{(1)}$.  The basis function $\max(0, W^{(1)}_k X + b^{(1)}_k)$ is again a two-piece linear function with a bending  along the line $W^{(1)}_k X + b^{(1)}_k = 0$. The dividing lines $\{W^{(1)}_k X + b^{(1)}_k = 0, k = 1, ..., d^{(1)}\}$ partition the domain of $X$ into up to $2^{d^{(1)}}$ pieces, and $f(X)$ is a continuous piecewise linear function over these pieces. 

In the multi-layer network, the hierarchical layers of $\{h^{(l)}, l = 1, ..., L\}$ partition the domain of $X$ recursively, creating a piecewise linear function with exponentially many pieces  \cite{pascanu2013number}. Such reasoning also applies to other forms of rectification functions $f^{(l)}$, as long as they are non-linear and create bending. This makes the neural network an extremely powerful machine for function approximation and interpolation. The recursive partition in neural nets is similar to CART and MARS, but is more flexible. 

{\em Back-propagation}. Both $\partial f_\theta(X)/\partial \theta$ and $\partial f_\theta(X)/\partial X$  can be computed by the chain-rule back-propagation, and they share the computation of 
$\partial h^{(l)}/\partial h^{(l-1)} = f^{(l)'}(W^{(l)} h^{(l-1)} + b^{(l)}) W^{(l)}$ in the chain rule. Because $f^{(l)}$ is coordinate-wise, $f^{(l)'}$ is a diagonal matrix.

A recent invention  \cite{he2016deep} is to reparametrize the mapping (\ref{eq:recursion}) by $h^{(l)} = h^{(l-1)} +  f^{(l)}(W^{(l)} h^{(l-1)} + b^{(l)})$, where $f^{(l)}(W^{(l)} h^{(l-1)} + b^{(l)})$ is used to model the residual term. This enables the learning of very deep networks. One may think of it as modeling an iterative algorithm where the layers $l$ can be interpreted as time steps of the iterative algorithm.

\begin{figure}[h]
	\centering
	\includegraphics[width=.25\linewidth]{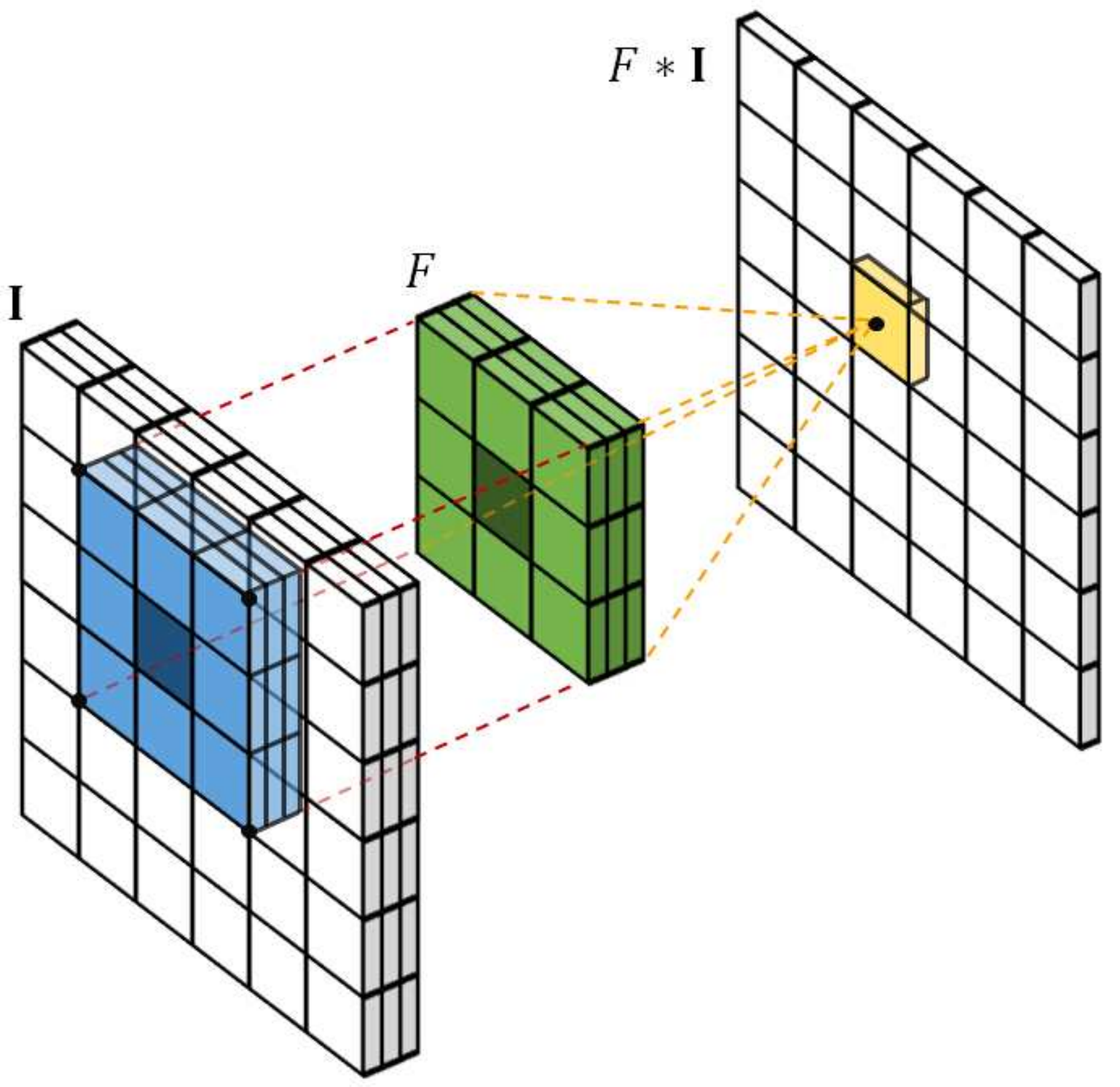}	
	\caption{ Filtering or convolution:  applying a filter of the size $3 \times 3 \times 3$ on an  image of the size $6 \times 6 \times 3$  to get a filtered image or feature map of $6 \times 6$ (with proper boundary handling). Each pixel of the filtered image is computed by the weighted sum of  the $3 \times 3 \times 3$ pixels of the input image centered at this pixel. There are 3 color channels (R, G, B),  so both the input image and the filter are three-dimensional. }
\label{fig:Conv}
\end{figure}

\begin{figure}[h]
	\centering	
	\includegraphics[width=.8\linewidth ]{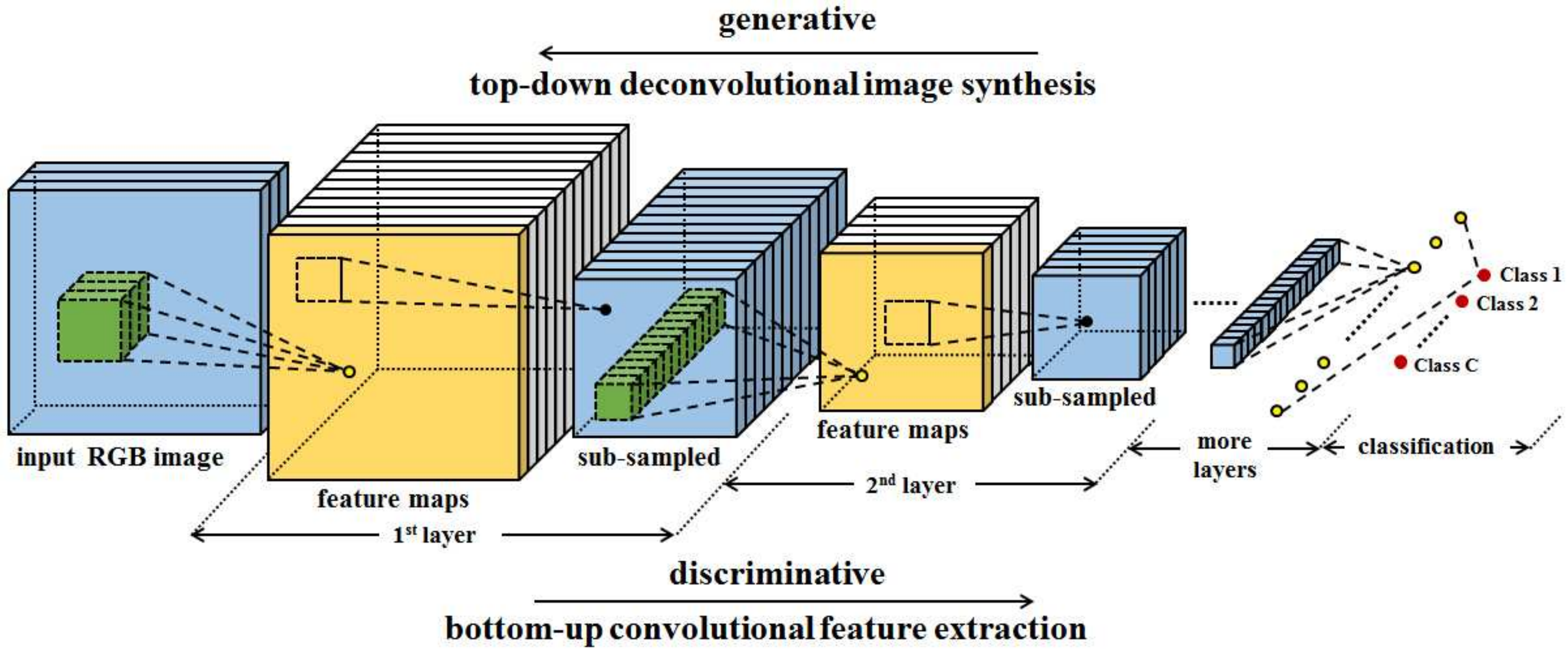}	
	\caption{Convolutional neural networks consist of multiple layers of filtering and sub-sampling operations for bottom-up feature extraction, resulting in multiple layers of feature maps and their sub-sampled versions. The top layer features are used for classification via multinomial logistic regression.  The discriminative direction is from image to category,  whereas the generative direction is from category to image. }
	\label{fig:ConvNet}
\end{figure}

{\em Convolution}. The signal $X$ can be an image,  and the linear transformations at each layer may be convolutions with localized kernel functions (i.e. filters). That is, the row vectors of $W^{(l)}$ (as well as the elements of $b^{(l)}$) form different groups, and the vectors in the same group are localized and translation invariant versions of each other, like wavelets. Each group of vectors corresponds to a filter or a kernel or a channel.  See Figures \ref{fig:Conv} and \ref{fig:ConvNet} for illustrations. Recent networks mostly use small filters of the size $3 \times 3$ \cite{simonyan2014very, szegedy2016rethinking}. The minimal size $1\times 1$ is also a popular choice \cite{lin2013network, szegedy2016rethinking}. Such a filter fuses the features of different channels at the same location, and is often used for reducing or increasing the number of channels. When computing the filtered image, we can also sub-sample it by, e.g., taking one filter response every two pixels. The filter is said to have stride 2.

\subsection{Descriptive models} \label{sect:d} 

This subsection describes the hierarchical form of the descriptive models and the maximum likelihood learning algorithm. 

We can generalize the descriptive model in the previous sections to a hierarchical form with multiple layers of features \cite{Ng2011, Dai2015ICLR,  XieLuICML, XieCVPR17}, 
\begin{eqnarray} 
 X \rightarrow h^{(1)} \rightarrow ... \rightarrow h^{(L)} \rightarrow f_\theta(X)
\end{eqnarray}
which is a bottom-up process for computing $f_\theta(X)$, and $\theta$ collects all the weight and bias parameters at all the layers. The probability distribution is 
\begin{eqnarray} 
   p_\theta(X) = \frac{1}{Z(\theta)} \exp\left[ f_\theta(X)\right] p_0(X), \label{eq:model}
\end{eqnarray}
 where  again $p_0(X)$ is the reference distribution such as Gaussian white noise model 
$p_0(X) \propto \exp \left( -{\|X\|^2}/{2 \sigma^2}\right)$. Again the normalizing constant is 
$Z(\theta) = \int  \exp( f_\theta(X)) p_0(X) dX  = \E_{p_0}[ \exp(f_\theta(X))]$.   The energy function is 
\begin{eqnarray}
U_\theta(X) = \|X\|^2/2\sigma^2 - f_\theta(X). 
\end{eqnarray}
$q_0(X)$ can also be a uniform distribution within a bounded range, then $U_\theta(X) = - f_\theta(X)$. 

The model (\ref{eq:model}) can be considered a hierarchical generalization of the FRAME model. While the energy function of the FRAME model is defined in terms of element-wise non-linear functions of filter responses, model (\ref{eq:model}) involves recursions of this structure at multiple layers according to the ConvNet. 

 Suppose we observe training examples $\{X_i, i = 1, ..., n\}$. The maximum likelihood learning seeks to maximize   
$\L(\theta) = \frac{1}{n} \sum_{i=1}^{n} \log p_\theta(X_i)$. 
The gradient of $\L(\theta)$ is 
  \begin{eqnarray} 
  \L'(\theta)& =& \frac{1}{n} \sum_{i=1}^{n} \frac{\partial}{\partial \theta} f_\theta(X_i)- \E_{\theta} \left[\frac{\partial}{\partial \theta} f_\theta(X)\right],  \label{eq:lD}
\end{eqnarray} 
where  $\E_{\theta}$ denotes the expectation with respect to $p_\theta(X)$. The key identity underlying equation (\ref{eq:lD}) is $d \log Z(\theta)/d\theta = \E_\theta[\partial f_\theta(X)/\partial \theta]$. 

The expectation in equation (\ref{eq:lD}) is analytically intractable and has to be approximated by MCMC, such as the Langevin  dynamics, which samples from $p_\theta(X)$ by  iterating the following step: 
\begin{eqnarray}
   X_{\tau+1} &=& X_\tau - \frac{s^2}{2} \frac{\partial}{\partial X} U_\theta(X_\tau) + s \mathcal{E}_{\tau} \\
   &=& X_\tau - \frac{s^2}{2} \left[ \frac{X_\tau}{\sigma^2} - \frac{\partial}{\partial X} f_\theta(X_\tau) \right] + s \mathcal{E}_{\tau},  \label{eq:LangevinD}
\end{eqnarray}
where $\tau$ indexes the time steps of the Langevin dynamics, $s$ is the step size, and $\mathcal{E}_{\tau} \sim \N(0, I_p)$ is the Gaussian white noise term. A Metropolis-Hastings step can be added to correct for the finiteness of $s$. The Langevin dynamics was used by  \cite{zhu1997GRADE} for sampling from the linear form of the descriptive model such as the FRAME model.

We can run $\tn$ parallel chains of Langevin dynamics according to (\ref{eq:LangevinD}) to obtain the synthesized examples  $\{\tX_i, i = 1, ..., \tn\}$. The Monte Carlo approximation to $\L'(\theta)$ is 
\begin{eqnarray} 
  \L'(\theta) &\approx&   \frac{\partial}{\partial \theta} \left[\frac{1}{n} \sum_{i=1}^{n} f_\theta(X_i)- \frac{1}{\tn} \sum_{i=1}^{\tn}  f_\theta(\tX_i)\right],  \label{eq:learningD}
\end{eqnarray} 
which is the difference between the observed examples and the synthesized examples. We can then update $\theta^{(t+1)} = \theta^{(t)} + \eta_t \L'(\theta^{(t)})$, with $\L'(\theta^{(t)})$ computed according to (\ref{eq:learningD}). $\eta_t$ is the learning rate. The convergence of this algorithm has been studied by \cite{robbins1951stochastic, younes1999convergence}. 

{\em Alternating back-propagation}: The learning and sampling algorithm is again an ``analysis by synthesis'' scheme. The sampling step runs the Langevin dynamics by computing $\partial f_\theta(X)/\partial X$, and the learning step updates $\theta$ by computing $\partial f_\theta(X)/\partial \theta$. Both derivatives can be computed by back-propagation, and they share the same computations of $\partial h^{(l)}/\partial h^{(l-1)}$. 

{\em Mode shifting interpretation}:  The data distribution $\P$ is likely to have many local modes. The $f_\theta(X)$ parametrized by the ConvNet can be flexible enough to creates many local modes to fit $\P$. We should learn $f_\theta(X)$ or equivalently the energy function $U_\theta(X)$ so that the energy function puts lower values on the observed examples than the unobserved examples. This is achieved by the learning and sampling algorithm, which can be interpreted as density shifting or mode shifting. In the sampling step, the Langevin dynamics settles the synthesized examples $\{\tX_i\}$ at the low energy regions or high density regions, or major modes (or basins) of $U_\theta(X)$, i.e., modes with low energies or high probabilities, so that $\frac{1}{\tn}\! \sum\nolimits_{i=1}^{\tn}\!U_\theta(\tX_i)$ tends to be low. The learning step seeks to change the energy function $U_\theta(X)$ by changing $\theta$ in order to increase  $\frac{1}{\tn}\! \sum\nolimits_{i=1}^{\tn}\!U_\theta(\tX_i)  - \frac{1}{n} \sum\nolimits_{i=1}^{n} U_\theta(X_i)$. This  has the effect of shifting the low energy or high density regions from the synthesized examples $\{\tX_i\}$  toward the observed examples $\{X_i\}$,  or  shifting the major modes of  the energy function $U_\theta(X)$ from the synthesized examples toward the observed examples, until the observed examples reside in the major modes of the model. If the major modes are too diffused around the observed examples, the learning step will sharpen them to focus on the observed examples. This mode shifting interpretation is related to Hopfield network \cite{hopfield1982neural} and attractor network \cite{Seung98learningcontinuous} with the Langevin dynamics serving as the attractor dynamics. 

The energy landscape may have numerous major modes that are not occupied by the observed examples, and these modes imagine examples that are considered similar to the observed examples. While the maximum likelihood learning matches the average statistical properties between model and data, the  ConvNet is expressive enough to create modes to encode the highly varied patterns. We still lack an in-depth understanding of the energy landscape. 

{\em Adversarial interpretation}: The learning and sampling algorithm also has an adversarial interpretation where the learning and sampling steps play a minimax game. Let the value function be defined as 
\begin{eqnarray}
V = \frac{1}{\tn} \sum\nolimits_{i=1}^{\tn}U_\theta(\tX_i) \ - \frac{1}{n} \sum\nolimits_{i=1}^{n} U_\theta(X_i).
\end{eqnarray}
The learning step updates $\theta$ to increase $V$, while the Langevin sampling step tends to relax $\{\tX_i\}$ to decrease $V$. The zero temperature limit of the Langevin sampling is gradient descent that decreases $V$, and the resulting learning and sampling algorithm is a generalized version of herding \cite{welling2009herding}. See also \cite{XieCVPR17}. This is related to Wasserstein GAN \cite{arjovsky2017wasserstein}, but the critic and the actor are the same descriptive model, i.e., the model itself is its own generator and critic.

{\em Multi-grid sampling and  learning}: 
In the high-dimensional space, e.g. image space, the model can be highly multi-modal. The MCMC in general and the Langevin dynamics in particular may have difficulty traversing different modes and it may be very time-consuming to converge. A simple and popular modification of the maximum likelihood learning is the contrastive divergence (CD) learning  \cite{Hinton2002a}, where we obtain the synthesized example by initializing a finite-step MCMC from the observed example. The CD learning is related to score matching estimator \cite{Hyvrinen05estimationof, hyvarinen2007connections} and auto-encoder \cite{vincent2011connection, Swersky2011, alain2014regularized}. Such a method has the ability to handle large training datasets via mini-batch training. However, bias may be introduced in the learned model parameters in that the synthesized images can be far from the fair examples of the current model. A further modification of CD is persistent CD  \cite{tieleman2008training}, where at the initial learning epoch the MCMC is still initialized from the observed examples, while in each subsequent learning epoch, the finite-step MCMC is initialized from the synthesized example of the previous epoch. The resulting synthesized examples can be less biased by the observed examples. However, the persistent chains may still have difficulty traversing different modes of the learned model.

\begin{figure}[h]
\begin{center}
\includegraphics[width=.07\linewidth]{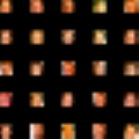} 
	\includegraphics[width=.14\linewidth]{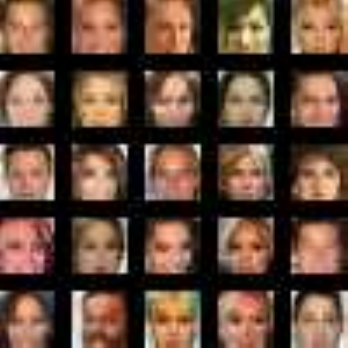} 	 
	\includegraphics[width=.28\linewidth]{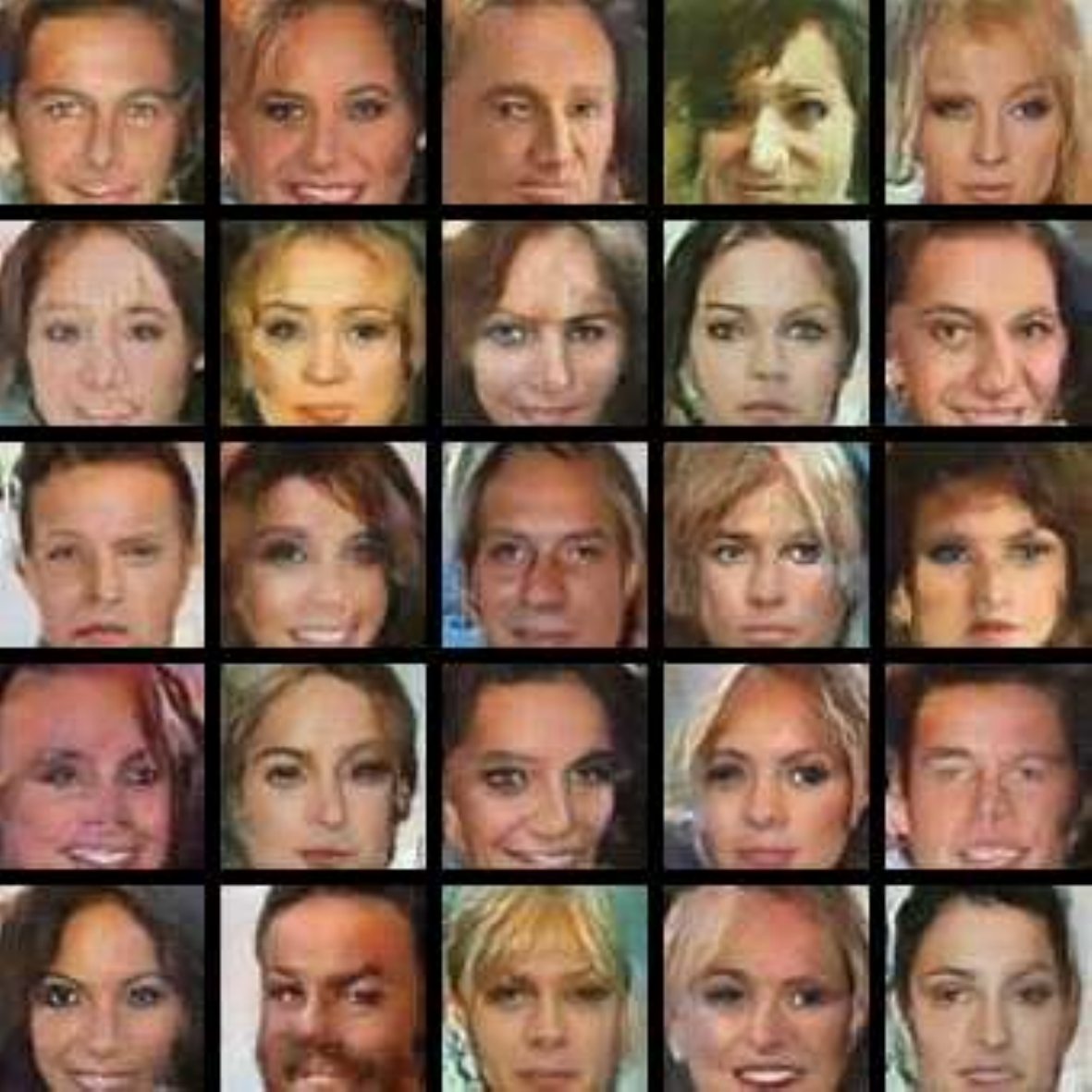}
\end{center}
   \caption{Synthesized images at multi-grids \cite{gao2017learning}. From left to right: $4 \times 4$ grid, $16 \times 16$ grid and $64 \times 64$ grid. Synthesized image at each grid is obtained by $30$ step Langevin sampling initialized from the synthesized image at the previous coarser grid, beginning with the $1 \times 1$ grid.}
\label{fig:1}
\end{figure}

\begin{figure}[h]
\begin{center}
\setlength{\tabcolsep}{1.5pt}
\begin{tabular}{cccc}
\includegraphics[width=.22\linewidth]{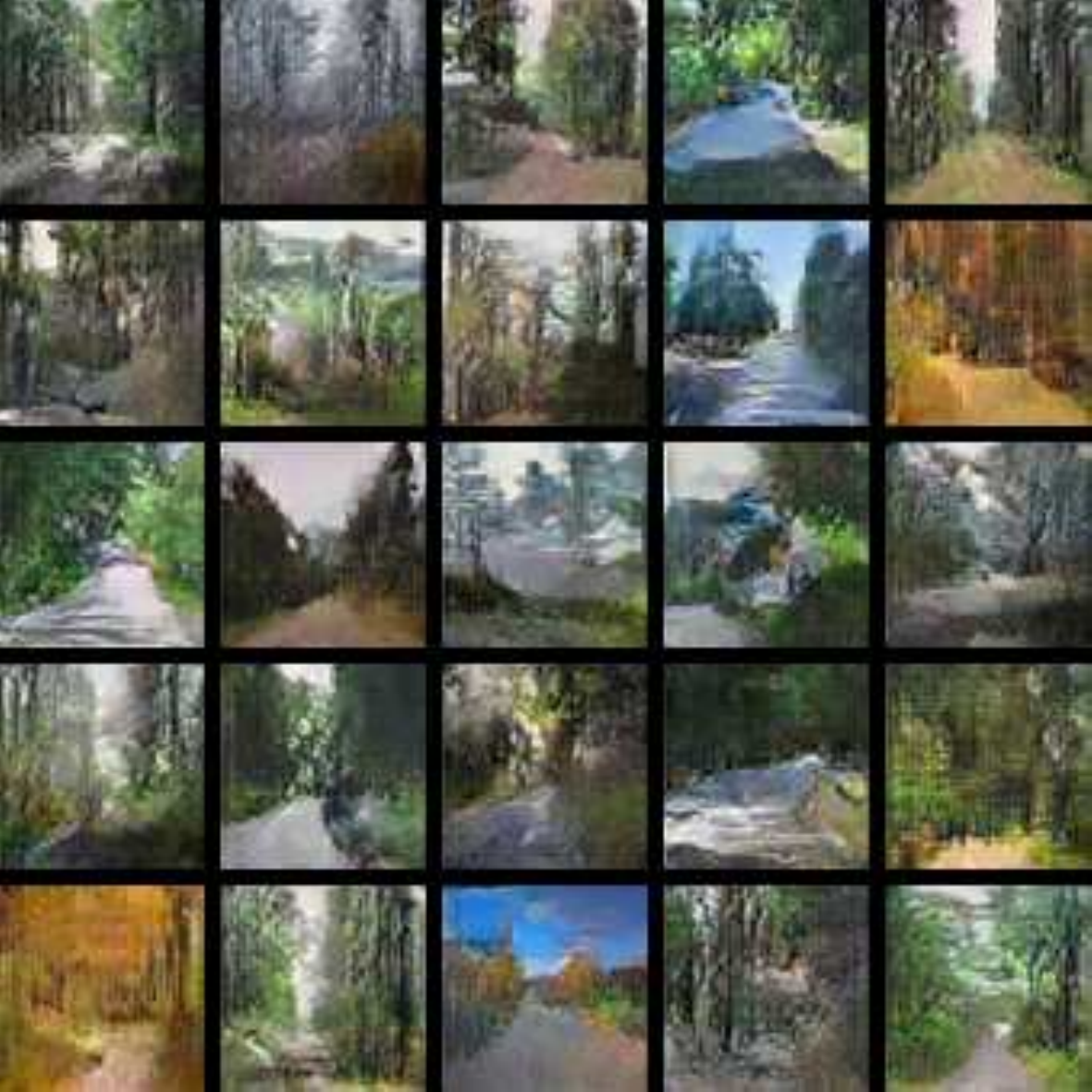} &
\includegraphics[width=.22\linewidth]{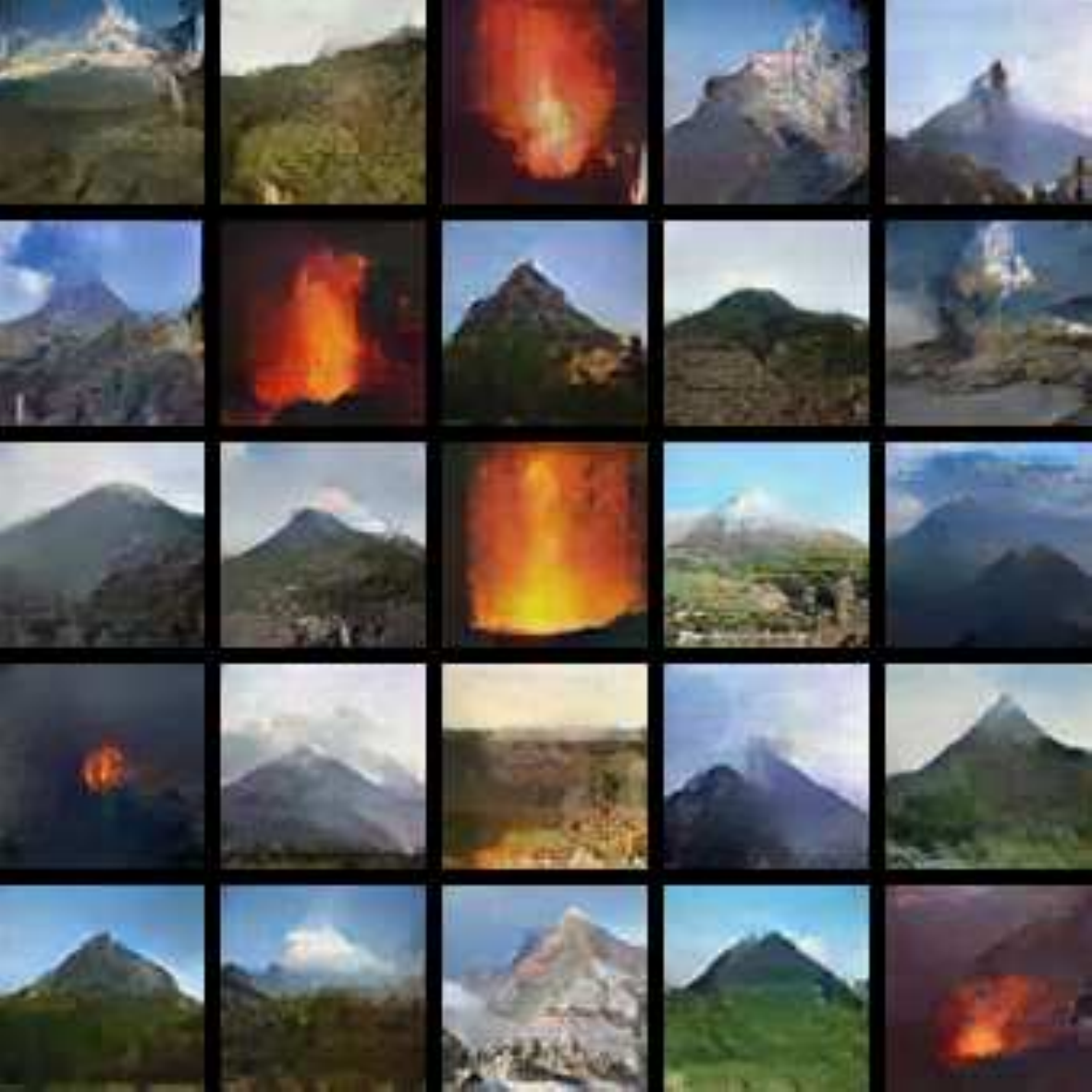} &
\includegraphics[width=.22\linewidth]{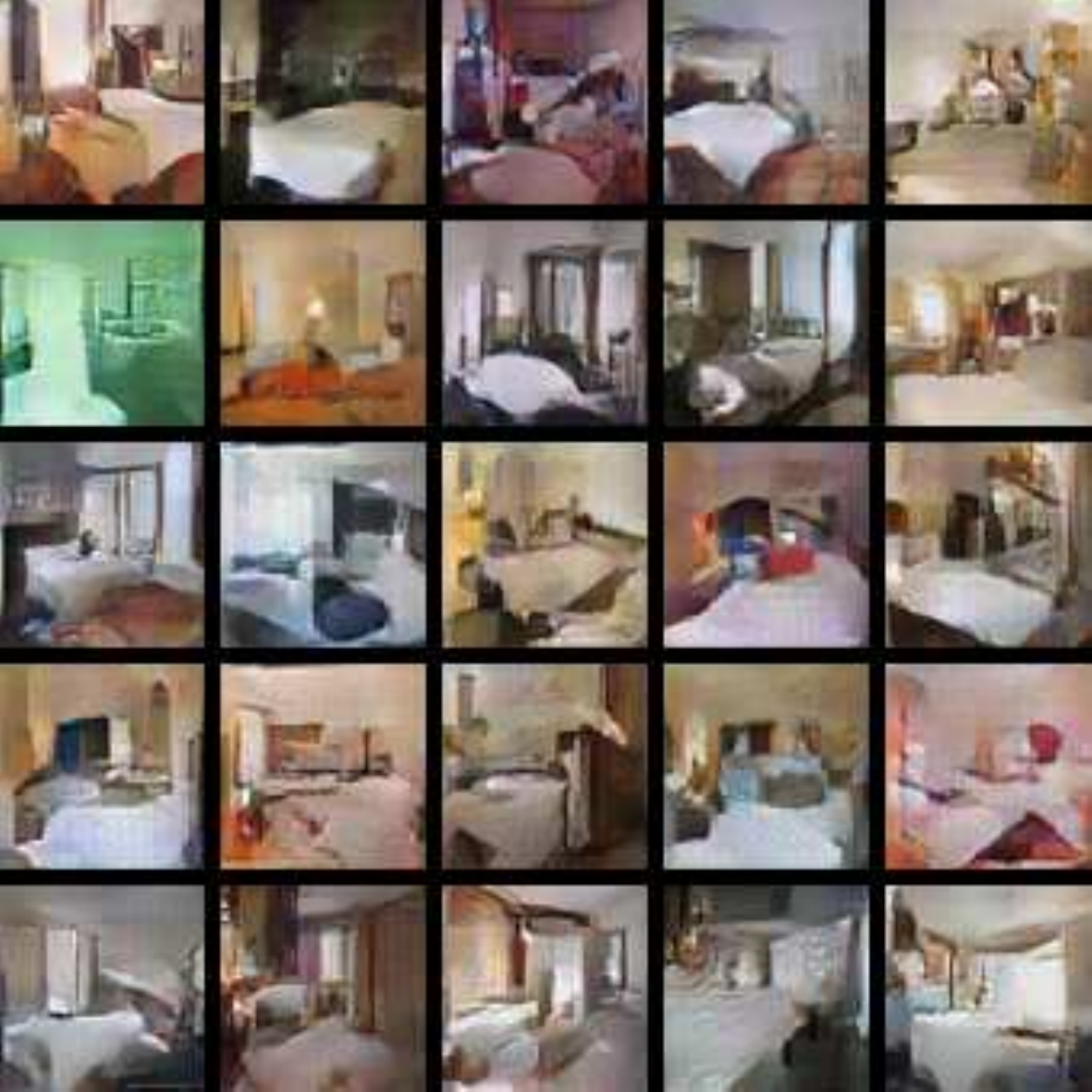} &
\includegraphics[width=.22\linewidth]{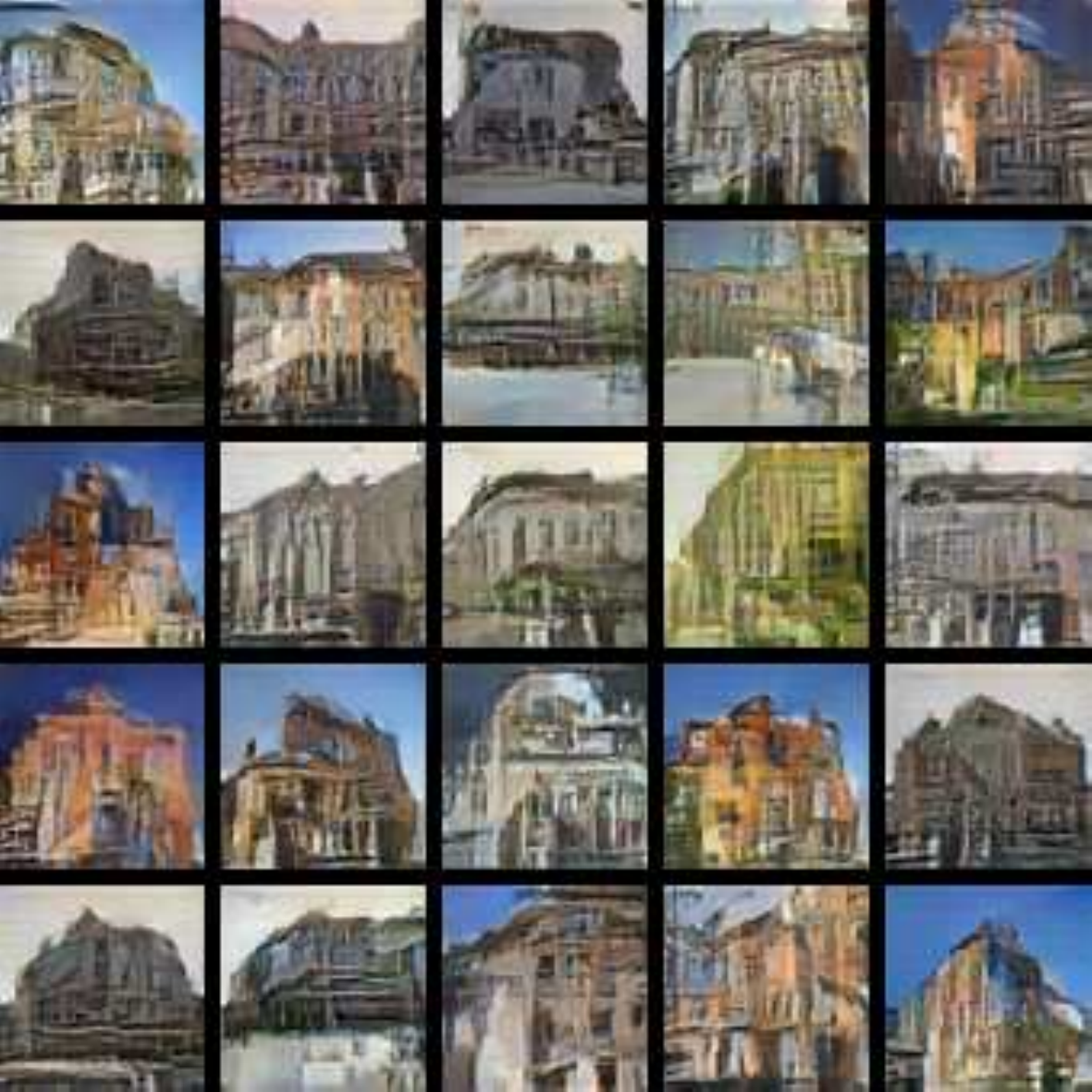}
\\
Forest road & Volcano &
Hotel room & Building facade
\end{tabular}
\caption{ Synthesized images from models learned by multi-grid method \cite{gao2017learning} from 4 categories of MIT places205 datasets.}
\label{fig:syn2}
\end{center}
\end{figure}

In  \cite{gao2017learning}, we developed a multi-grid sampling and learning method to address the above challenges under the constraint of finite budget MCMC. Specifically, we repeatedly down-scale each training image to get its multi-grid versions. Our method learns a separate descriptive model at each grid. Within each iteration of our learning algorithm, for each observed training image, we generate the corresponding synthesized images at multiple grids.  Specifically, we initialize the finite-step MCMC sampling from the minimal $1\times 1$ version of the training image, and the synthesized image at each grid serves to initialize the finite-step MCMC that samples from the model of the subsequent finer grid. See Figure \ref{fig:1} for an illustration, where we sample images sequentially at 3 grids, with 30 steps of Langevin dynamics at each grid. After obtaining the synthesized images at the multiple grids, the  models at the multiple grids  are updated separately and simultaneously based on the differences between the synthesized images and the observed training images at different grids. 

Unlike original CD or persistent CD, the learned models are capable of generating new synthesized images from scratch with a fixed budget MCMC, because we only need to initialize the MCMC by sampling from the one-dimensional histogram of the $1 \times 1$ versions of the training images.

In our experiments, the training images are resized to $64 \times 64$. Since the models of the three grids act on images of different scales, we design a specific ConvNet structure per grid: grid1 has a 3-layer network with $5 \times 5$ stride $2$ filters at the first layer and $3 \times 3$ stride $1$ filters at the next two layers; grid2 has a 4-layer network with $5 \times 5$ stride $2$ filters at the first layer and $3 \times 3$ stride $1$ filters at the next three layers; grid3 has a 3-layer network with $5 \times 5$ stride $2$ filters at the first layer, $3 \times 3$ stride $2$ filters at the second layer, and $3 \times 3$ stride $1$ filters at the third layer. Numbers of channels are $96-128-256$ at grid1 and grid3, and $96-128-256-512$ at grid2. A fully-connected layer with $1$ channel output is added on top of every grid to get the value of the function $f_\theta(X)$.   At each iteration, we run $l = 30$ steps of Langevin dynamics for each grid with step size $s = 0.3$. All networks are trained simultaneously with mini-batches of size $100$ and an initial learning rate of $0.3$. Learning rate is decayed logarithmically every $10$ iterations. 

We learn multi-grid models from several datasets including CelebA \cite{liu2015deep}, MIT places205 \cite{zhou2014learning} and CIFAR-10 \cite{krizhevsky2009learning}. In the CelebA dataset, we randomly sample 10,000 images for training.  Figure \ref{fig:1} shows the synthesized examples. 
Figure \ref{fig:syn2} shows synthesized images from models learned from $4$ categories of MIT places205 dataset by multi-grid method. We learn from each category separately. The number of training images is $15,100$ for each category.

Traditionally, the mixing time of Markov chain is defined via $d(t) = \max_x \|P^{(t)}(x, \cdot)-\pi\|_{\rm TV}$, where $P^{(t)}$ is the $t$-step transition, $\pi$ is the stationary distribution, and $\|\cdot\|_{\rm TV}$ is the total variation distance. This is the worst case scenario by choosing the least favorable point mass at $x$. In our method, however, the initial distribution at each grid can  be much more favorable, e.g., it may already agree approximately with $\pi$ on the marginal distribution of the coarser grid, so that after $t$ steps, the distribution of the sampled image can be close to $\pi$, even if this is not the case for the worst case starting point. Such non-persistent finite budget MCMC is computationally more manageable than persistent chains in learning.

\begin{figure}[h]
\begin{center}
\setlength{\tabcolsep}{1.5pt}
\begin{tabular}{cc}
\includegraphics[width=.45\linewidth]{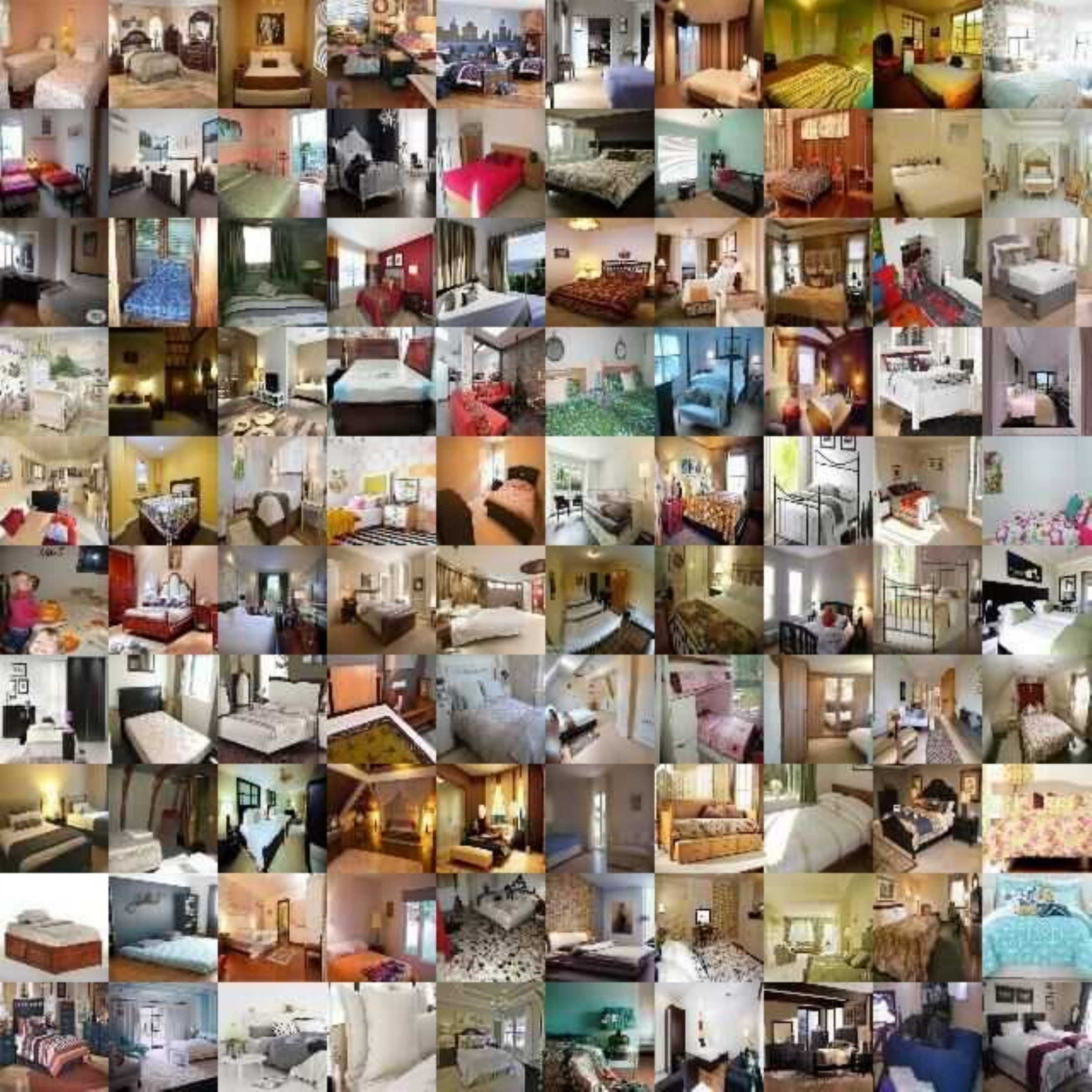} &
\includegraphics[width=.45\linewidth]{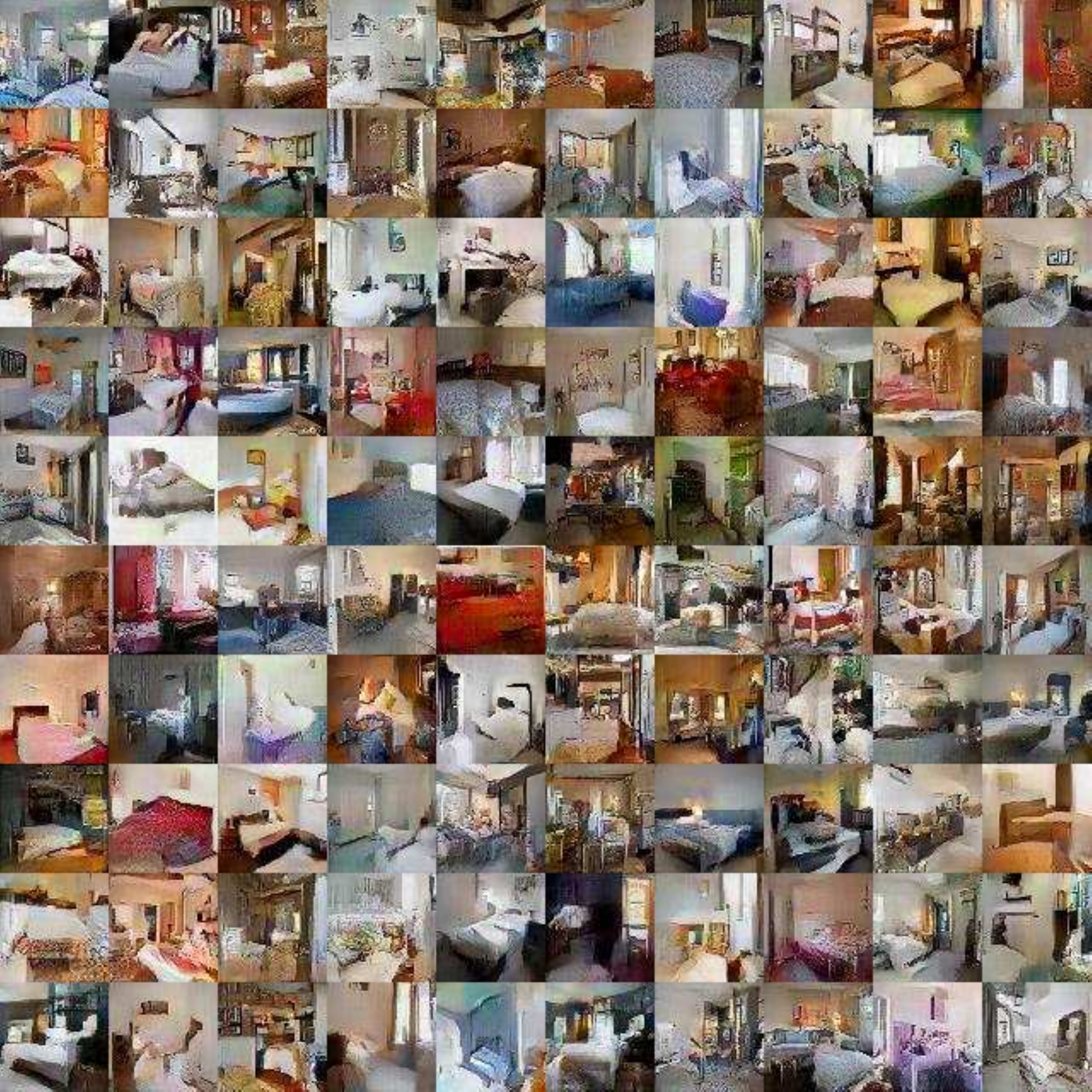} 
\end{tabular}
\caption{ Learning the multi-grid models from the LSUN bedroom dataset \cite{gao2017learning}. Left: random samples of training examples. Right: synthesized examples generated by the learned  models.}
\label{fig:LSUNm}
\end{center}
\end{figure}

To train  multi-grid models on 10,000 training images for 400 iterations with a singe Titan X GPU, it takes about $7.45$ hours. After training, it takes less than 1 second to generate a batch of 100 images. We also train the multi-grid models  on LSUN bedroom dataset \cite{DBLP:journals/corr/YuZSSX15}, which consists of roughly 3 million images. Figure \ref{fig:LSUNm} shows the learning results after 8 epochs.

\begin{figure}[h]
\begin{center}
\setlength{\tabcolsep}{2pt}
\begin{tabular}{ccc|ccc|ccc|ccc}
\includegraphics[width=.05\linewidth]{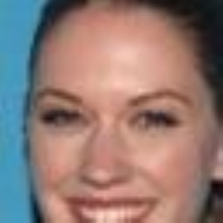}  &
\includegraphics[width=.05\linewidth]{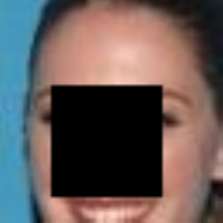}   &
\includegraphics[width=.05\linewidth]{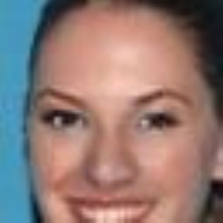}  	&
\includegraphics[width=.05\linewidth]{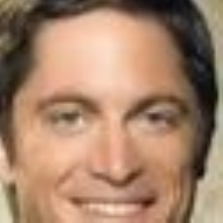}  &
\includegraphics[width=.05\linewidth]{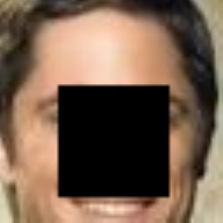}   &
\includegraphics[width=.05\linewidth]{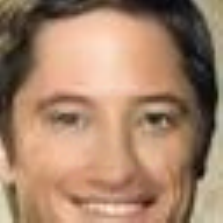}  	&
\includegraphics[width=.05\linewidth]{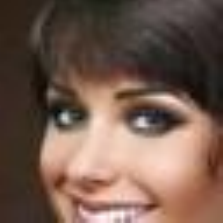}  &
\includegraphics[width=.05\linewidth]{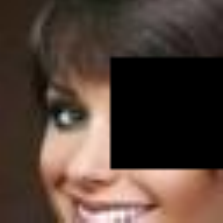}   &
\includegraphics[width=.05\linewidth]{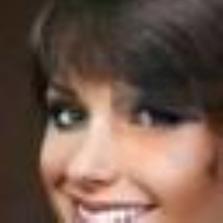}  	&
\includegraphics[width=.05\linewidth]{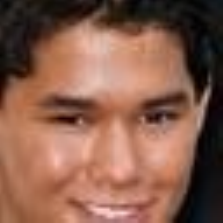}  &
\includegraphics[width=.05\linewidth]{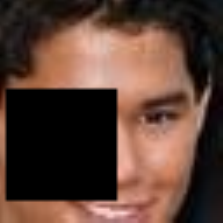}   &
\includegraphics[width=.05\linewidth]{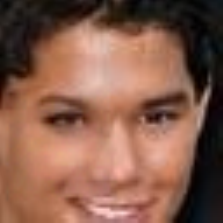}  	
\\
\includegraphics[width=.05\linewidth]{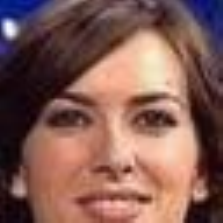}  &
\includegraphics[width=.05\linewidth]{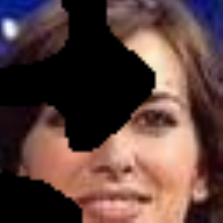}   &
\includegraphics[width=.05\linewidth]{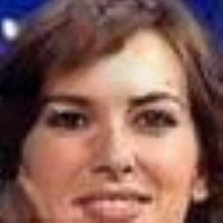}  	&
\includegraphics[width=.05\linewidth]{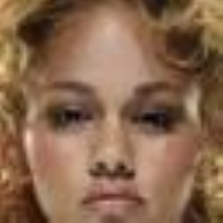}  &
\includegraphics[width=.05\linewidth]{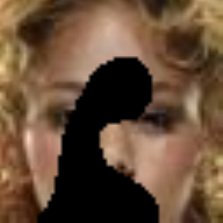}   &
\includegraphics[width=.05\linewidth]{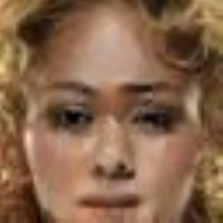}
&
\includegraphics[width=.05\linewidth]{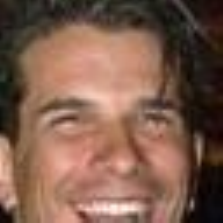}  &
\includegraphics[width=.05\linewidth]{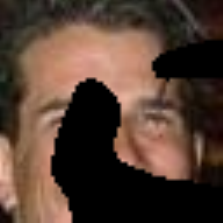}   &
\includegraphics[width=.05\linewidth]{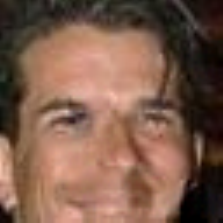}  	&
\includegraphics[width=.05\linewidth]{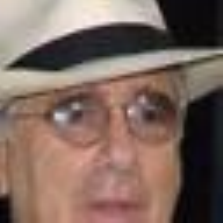}  &
\includegraphics[width=.05\linewidth]{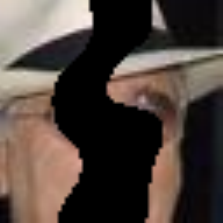}   &
\includegraphics[width=.05\linewidth]{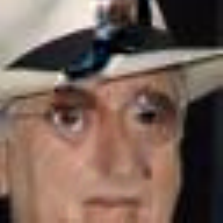}
\end{tabular}
\caption{Inpainting examples on CelebA dataset \cite{gao2017learning}. In each block from left to right: the original image; masked input; inpainted image by multi-grid method.}
\label{fig:recover}
\end{center}
\end{figure}

The learned descriptive model is a bottom-up ConvNet that consists of multiple layers of features. These features can be used for subsequent tasks such as classification. The learned models can also be used as a prior distribution for inpainting, as illustrated by Figure \ref{fig:recover}.  See \cite{gao2017learning} for experiment details and numerical evaluations. 
 
 \subsection{Introspective learning} \label{sect:logistic}
 
 This subsection describes the introspective learning method that learns the descriptive model by turning it into a discriminative model. 

Model (\ref{eq:model}) corresponds to a classifier in the following sense \cite{Dai2015ICLR,  XieLuICML, TuNIPS}. Suppose there are $K$ categories, $p_{{\theta}_k}(X)$, for $k = 1, ..., K$, in addition to the background category $p_0(X)$. The ConvNets $f_{{\theta}_k}(X)$ for $k = 1, ..., K$ may share common lower layers except the final layer for computing $f_{\theta_k}(X)$. Let $\rho_k$ be the prior probability of category $k$, $k = 0, ..., K$. Then the posterior probability for classifying an example $X$ to the category $k$ is a soft-max multi-class classifier, i.e., the multinomial logistic regression: 
\begin{eqnarray}\small
\Pr(k|X) = \frac{ \exp(f_{\theta_k}(X)+b_k)}{\sum_{k=0}^{K} \exp(f_{\theta_k}(X)+b_k)}, \label{eq:c}
\end{eqnarray}
 where $b_k = \log (\rho_k/\rho_0) - \log Z(\theta_k)$, and for $k = 0$, $f_{\theta_0}(X) = 0$, $b_0 = 0$. Conversely, if we have the soft-max classifier (\ref{eq:c}), then the distribution of each category is  $p_{\theta_k}(X)$ of the form (\ref{eq:model}). 
 Thus the descriptive model directly corresponds to the commonly used discriminative ConvNet model. 

\begin{figure}[h]
\begin{center}
\includegraphics[width=0.7\textwidth]{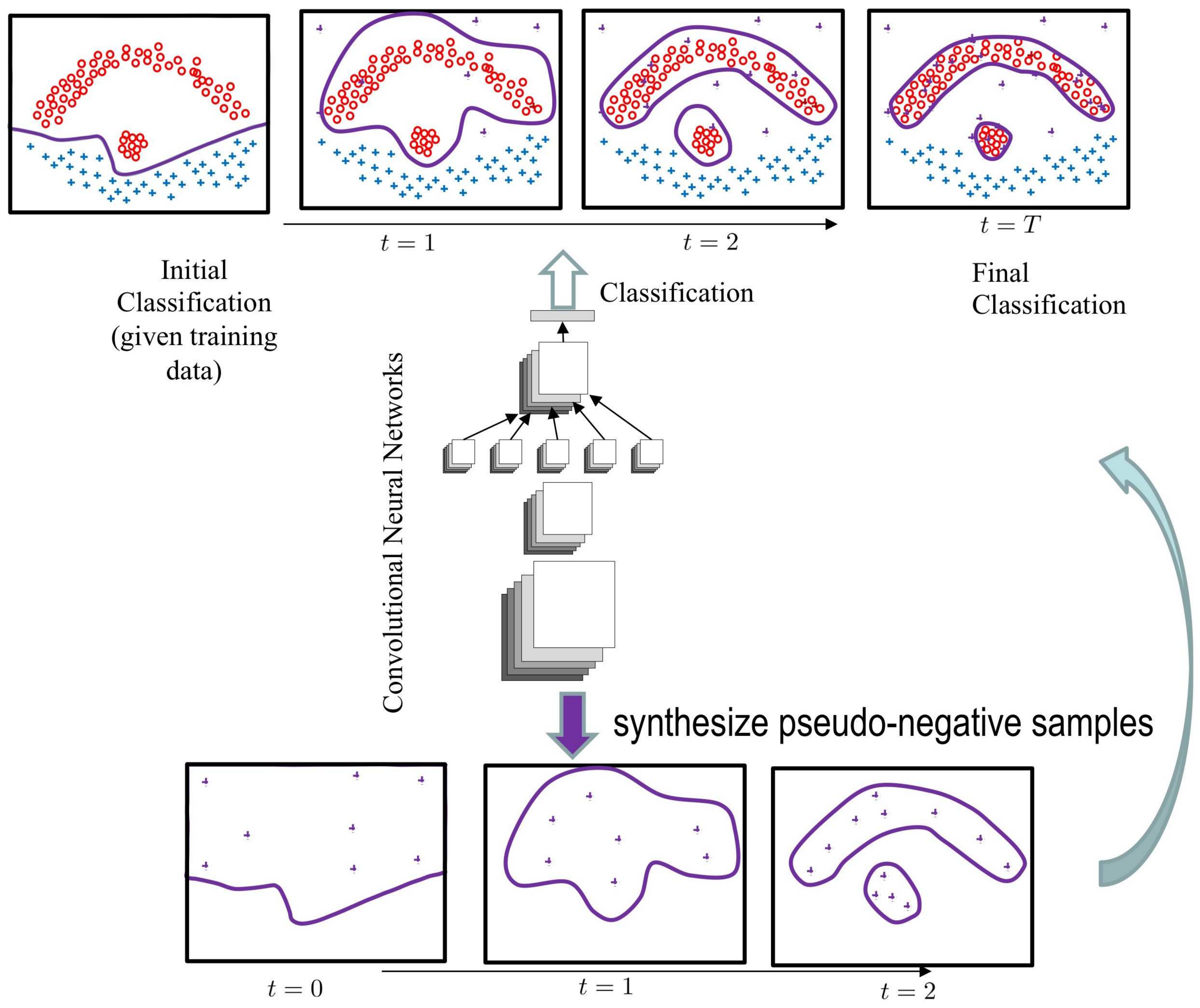}
\end{center}
		\caption{Introspective learning \cite{TuNIPS}: The discriminative ConvNet trained on the observed examples versus the synthesized examples generated by the current model can be used to update the model and to generate new examples from the updated model.  }
		\label{fig:Tu2}
\end{figure}

In the case where we only observe unlabeled examples, we may model them by a single distribution $p_1(X) = p_\theta(X)$ in (\ref{eq:model}), and treat it as the positive distribution, and treat $p_0(X)$ as the negative distribution. Let $\rho$ be the prior probability that a random example comes from $p_1$. Then the posterior probability that a random example $X$ comes from $p_1$ is 
\begin{eqnarray}\small
\Pr(1|X) = \frac{1}{1+ \exp[-(f_\theta(X)+b)]},
\end{eqnarray}
where $b = \log(\rho/(1-\rho)) - \log Z(\theta)$, i.e., a logistic regression. 

 \begin{figure}[h]
	\begin{center}
		\includegraphics[width=.5\linewidth]{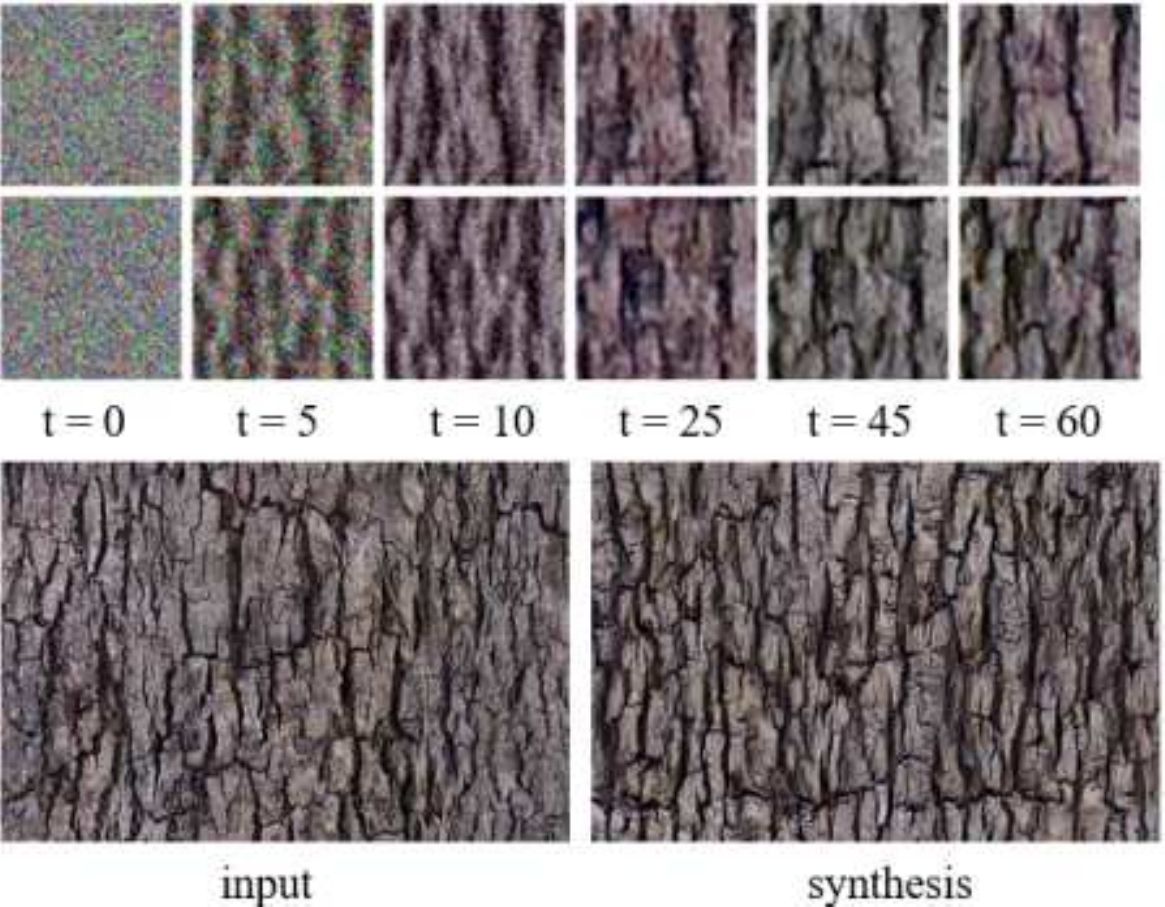}			
		\caption{Introspective learning \cite{TuNIPS}. Top row: patches of synthesized images in the introspective learning process. Bottom row: Left is the observed image. Right is the synthesized image generated by the learned model. }
		\label{fig:introspective}
	\end{center}	
\end{figure}

Generalizing \cite{tu2007learning}, \cite{TuNIPS} developed an introspective learning method for updating the model by learning a classifier or logistic regression to distinguish between the observed $\{X_i\}$ and the synthesized $\{\tX_i\}$, and tilt the current model according to the logistic regression. It is also an ``analysis by synthesis'' scheme as well as an adversarial scheme, except that the analysis is performed by a classifier. Specifically, let $p_{t}(X)$ be the current model. Each iteration of the introspective learning is as follows. The sampling step generates synthesized examples $\{\tX_i, i = 1, ..., n\}$ from $p_t(X)$. The learning step fits a logistic regression to separate the real examples $\{X_i, i = 1, ..., n\}$ from the synthesized examples $\{\tX_i, i = 1, ..., n\}$ to estimate $f_\theta(X)$ and $b$. Then we let $p_{t+1}(X) = \exp(f_\theta(X)) p_t(X)/Z$, where $\log Z = - b$. \cite{tu2007learning, TuNIPS} show that $p_t$ converges to $\P$ if the ConvNet is of infinite capacity.  See Figure \ref{fig:Tu2} for an illustration. 

 \begin{figure}[h]
\begin{center}
\includegraphics[width=0.4\textwidth]{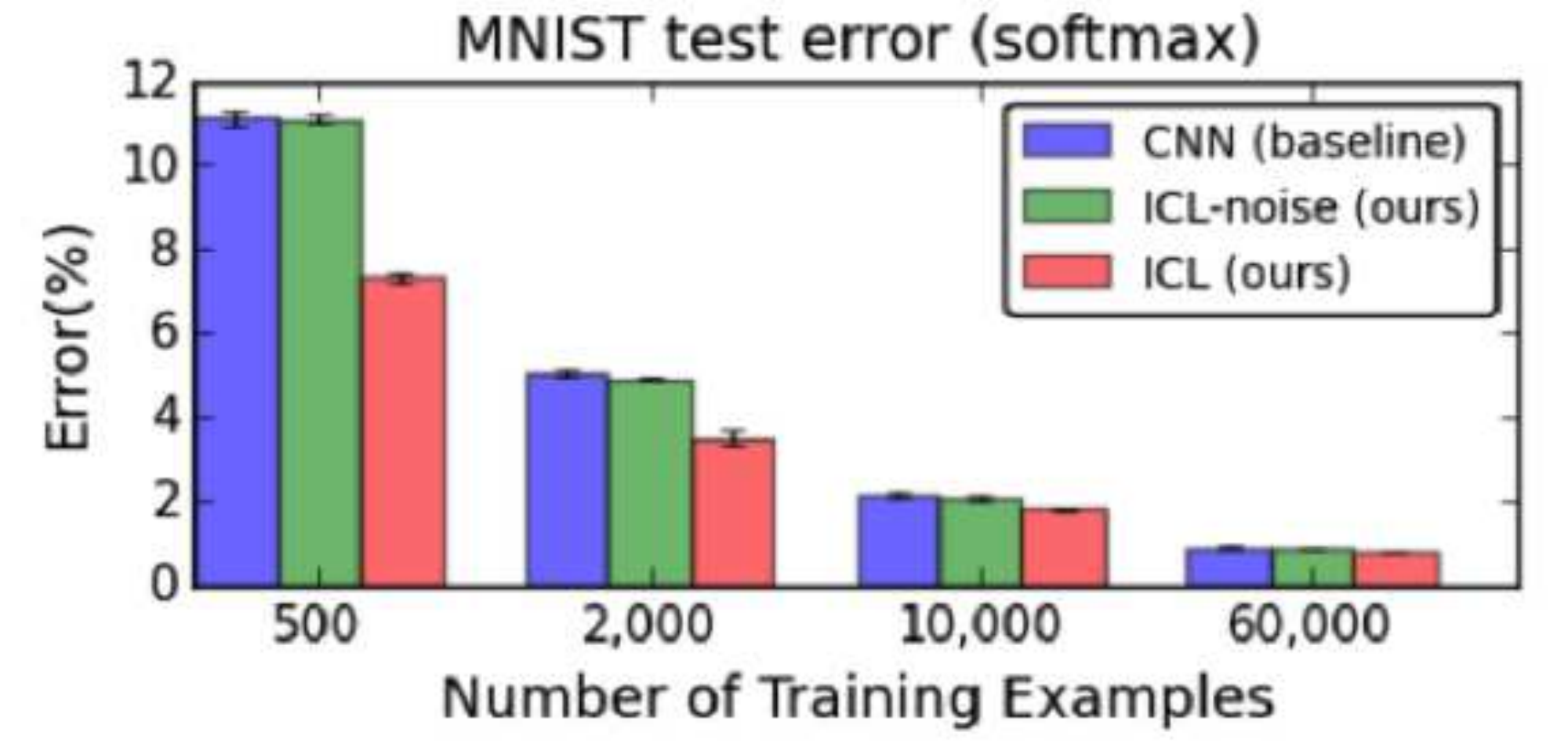}
\end{center}
		\caption{Introspective learning \cite{TuNIPS} improves the classification performances, especially if the training dataset is of small or moderate size. }
		\label{fig:Tu4}
\end{figure}

Numerical experiments in \cite{TuNIPS} show that the introspective method learns more accurate classifiers than purely discriminative methods in supervised learning, especially when the size of the training dataset is small or moderate. Figure \ref{fig:Tu4} shows the results.

The introspective learning unifies the discriminative model and the descriptive model \cite{TuNIPS, TuCVPR17, TuCVPR18}. Unlike the generative adversarial networks (GAN)  \cite{goodfellow2014generative}, the learned classifier is capable of introspection itself: it can be translated into a descriptive model to generate synthesized examples, without relying on a separate generative model.

\subsection{Generative models} \label{sect:g}

This subsection describes the hierarchical form of the generative models and the maximum likelihood learning algorithm.  

We can generalize the generative model in the previous sections to a hierarchical form with multiple layers of hidden variables
\begin{eqnarray}
h \rightarrow h^{(L)} \rightarrow ... \rightarrow h^{(1)} \rightarrow X, 
\end{eqnarray}
which is a top-down process that transforms $h$ to $X$. In the linear form of the generative model in the previous sections, the mapping from $h$ to $X$ is linear. In the hierarchical version, the mapping from $h$ to $X$ is a ConvNet defined by 
\begin{eqnarray} 
 h^{(l-1)} = g^{(l)}(W^{(l)} h^{(l)} + b^{(l)}),  \label{eq:gen}
\end{eqnarray} 
for $l = L+1, ..., 1$, where $h^{(L+1)} = h$ and $h^{(0)} = X$.  $g^{(l)}$ is the non-linear rectification function such as ReLU that is applied coordinate-wise. Let the resulting ConvNet be $X = g_\alpha(h)$, where $\alpha = (W^{(l)}, b^{(l)}, l = 1, 2, ..., L+1)$. 

The  top-down generative ConvNet was used by \cite{zeiler2011adaptive} to visualize the bottom-up ConvNet learned by the discriminative model. It was also used by \cite{Alexey2015} to learn a generative model of images of chairs, where the learning is supervised in that for each observed image of chair, a latent vector $h$ is provided to specify the type of chair (represented by a one-hot vector) as well as view point and other geometric properties. The top-down ConvNet can learn accurate mapping from $h$ to $X$, and the linear interpolation in the space of $h$ leads to very realistic non-linear interpolation in the space of $X$. 

 The generative model can  also be  learned in unsupervised setting where $h$ is unknown \cite{goodfellow2014generative, radford2015unsupervised,  KingmaCoRR13, RezendeICML2014, MnihGregor2014}. The model has the following form :
 \begin{align}
 &h \sim \N(0, I_d); \label{eq:NFA0}\\
  &X= g_\alpha(h)  + \epsilon; \epsilon \sim {\rm N}(0, \sigma^2 I_p),   \label{eq:NFA}
 \end{align}
where $h$ is the $d$-dimensional hidden vector of latent factors, $g_\alpha(h)$ is a top-down ConvNet that maps the $d$-dimensional vector $h$ to the $p$-dimensional signal $X$, where $d \leq p$. The model (\ref{eq:NFA}) is a generalization of factor analysis. While independent component analysis, sparse coding etc. generalize the prior distribution of factor analysis, the model (\ref{eq:NFA0}) and  (\ref{eq:NFA}) maintains the simple prior distribution of factor analysis, but generalizes the linear mapping in factor analysis to non-linear mapping parametrized by the top-down ConvNet (\ref{eq:gen}).  Like the word to vector representation \cite{mikolov2013distributed}, the hidden vector $h$ may capture semantically meaningful information in the signal $X$. 

The joint distribution
\begin{align} 
 \log q_\alpha(h, X) &= \log \left[q(h) q_\alpha(X|h) \right] \\
 & = - \frac{1}{2\sigma^2} \|X - g_\alpha(h)\|^2 - \frac{1}{2} \|h\|^2 + {\rm const}.
\end{align}
The marginal distribution $q_\alpha(X) = \int q_\alpha(h, X) dh$. 
The posterior distribution of the latent factors $q_\alpha(h|X)= q_\alpha(h, X)/q_\alpha(X) \propto q_\alpha(h, X)$. Here we use the notation $q_\alpha$ to denote the generative model in order to differentiate it from the descriptive model $p_\theta$. 

  In our recent work \cite{HanLu2016},  we study a maximum likelihood algorithm for learning the generative model (\ref{eq:NFA}) without resorting to an assisting network.
Specifically, if we observe a training set of examples $\{X_i, i =  1, ..., n\}$, then each $X_i$ has a corresponding latent $h_i$. We can train the generative model  by maximizing the observed-data log-likelihood
$ \L(\alpha) = \frac{1}{n} \sum_{i=1}^{n} \log q_\alpha(X_i)$. 

The gradient of $\L(\alpha)$ can be calculated according to the following identity:
\begin{align} 
\frac{\partial}{\partial \alpha} \log q_\alpha(X) \nonumber  &= \frac{1}{q_\alpha(X)} \int \left[\frac{\partial}{\partial \alpha} \log q_\alpha(h, X) \right] q_\alpha(h, X) dh \\
&= \E_{q_\alpha(h|X)} \left[\frac{\partial}{\partial \alpha} \log q_\alpha(X|h) \right]. \label{eq:cot}
\end{align}
The expectation with respect to $q_\alpha(h|X)$ can be approximated by drawing samples from $q_\alpha(h|X)$ and then computing the Monte Carlo average. 

The Langevin dynamics for sampling $h$ from $p_\alpha(h|X)$ is 
\begin{eqnarray}
&&h_{\tau + 1} = h_\tau+ \frac{s^2}{2}\left[\frac{1}{\sigma^2}( X-g_\alpha(h_\tau))  \frac{\partial}{\partial h} g_\alpha(h_\tau)  - {h_\tau}\right]  +   s \mathcal{E}_\tau ,  
\label{eq:Langevin}
\end{eqnarray} 
where $\tau$ denotes the time step, $s$ is the step size, and $\mathcal{E}_\tau \sim {\rm N}(0, I_d)$. Again we can add Metropolis-Hastings step to correct for the finiteness of $s$. 

We can use stochastic gradient algorithm of \cite{younes1999convergence}  for learning, where in each iteration, for each $X_i$, $h_i$ is sampled from $q_\alpha(h_i|X_i)$ by running a finite number of steps of Langevin dynamics starting from the current value of $h_i$. With the sampled $\{h_i\}$, we can update the parameters $\alpha$ based on the gradient ${\L}'(\alpha)$, whose Monte Carlo approximation is: 
\begin{align} 
{\L}'(\alpha) \approx \frac{1}{n}\sum_{i=1}^{n}  \frac{\partial}{\partial \alpha} \log q_\alpha(X_i|h_i) = \frac{1}{n} \sum_{i=1}^{n}   \frac{1}{\sigma^2} (X_i-g_\alpha(h_i)) \frac{\partial}{\partial \alpha} g_\alpha(h_i). \label{eq:learning}
\end{align} 
It is a non-linear regression of $X_i$ on $h_i$.  We update $\alpha^{(t+1)} = \alpha^{(t)} + \eta_t \L'(\alpha^{(t)})$, with $\L'(\alpha^{(t)})$ computed according to (\ref{eq:learning}). $\eta_t$ is the learning rate. The convergence of this algorithm follows \cite{younes1999convergence}.

{\em Alternating back-propagation}:  Like the descriptive model, the maximum likelihood learning of the generative model (\ref{eq:NFA}) also follows the alternative back-propagation scheme. The Langevin dynamics for inference needs to compute $\partial g_\alpha(h)/\partial h$. The learning step needs to compute $\partial g_\alpha(h)/\partial \alpha$. Both gradients can be computed by back-propagation and they share the computations of $\partial h^{(l-1)}/\partial h^{(l)}$. 

Our experiments show that the generative model is quite expressive.  We adopt the structure of the generator network of  \cite{radford2015unsupervised, Alexey2015}, where the top-down ConvNet consists of 5 layers. 

 \begin{figure}[h]
	\begin{center}
		\includegraphics[width=.1\linewidth]{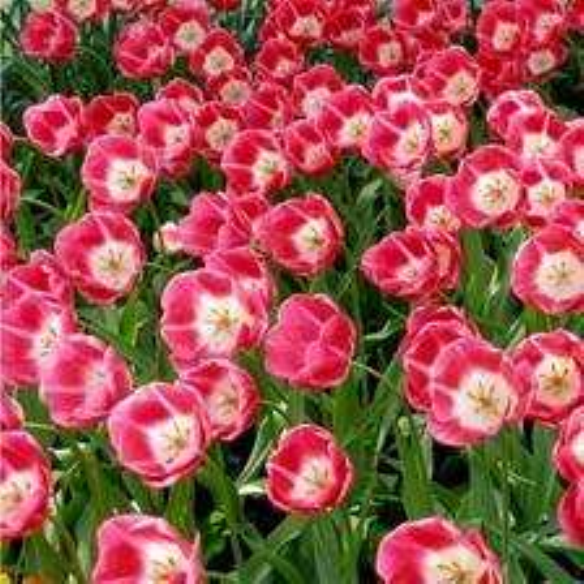}
		\includegraphics[width=.2\linewidth]{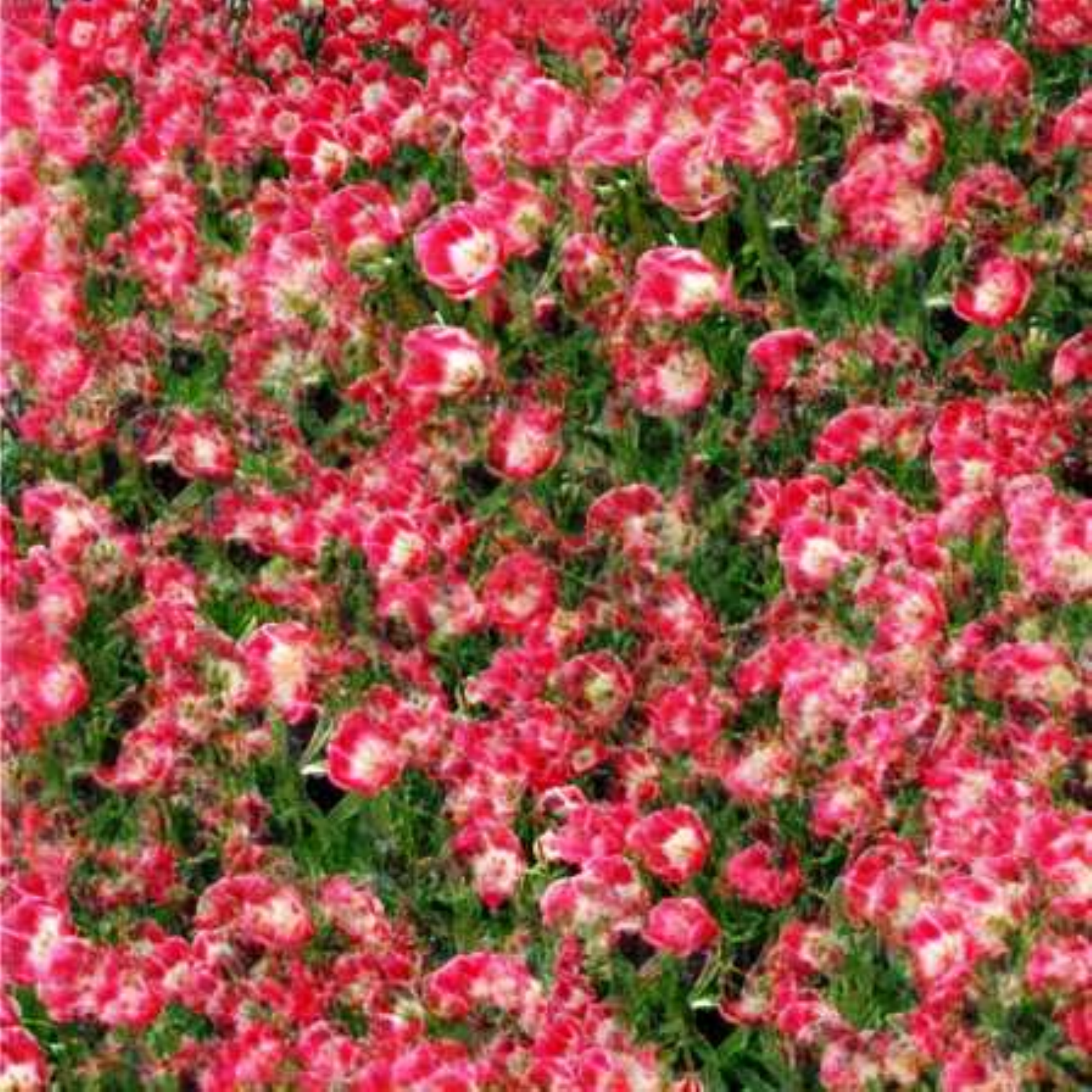}
		\hspace{1mm}
		\includegraphics[width=.1\linewidth]{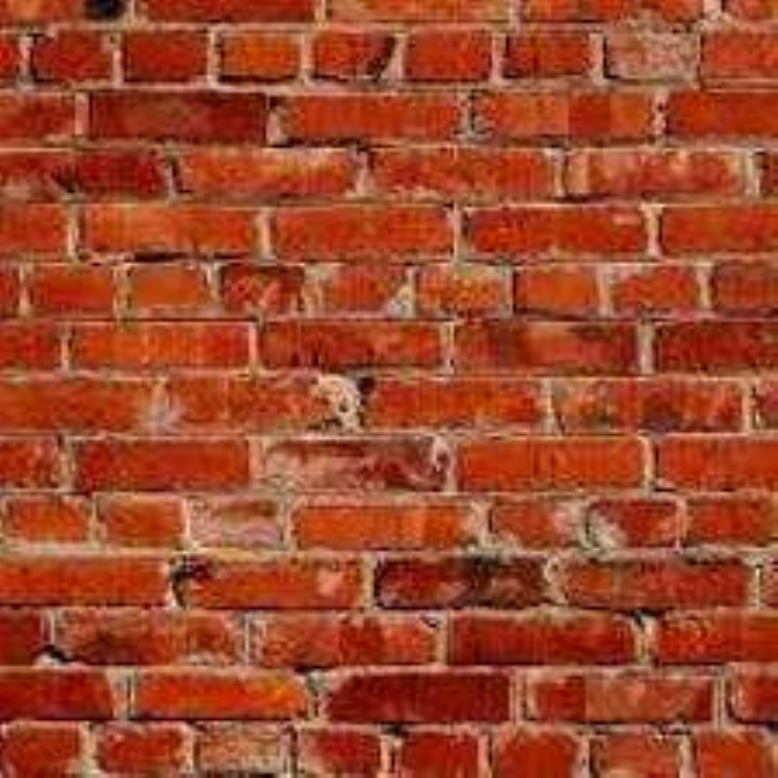}
		\includegraphics[width=.2\linewidth]{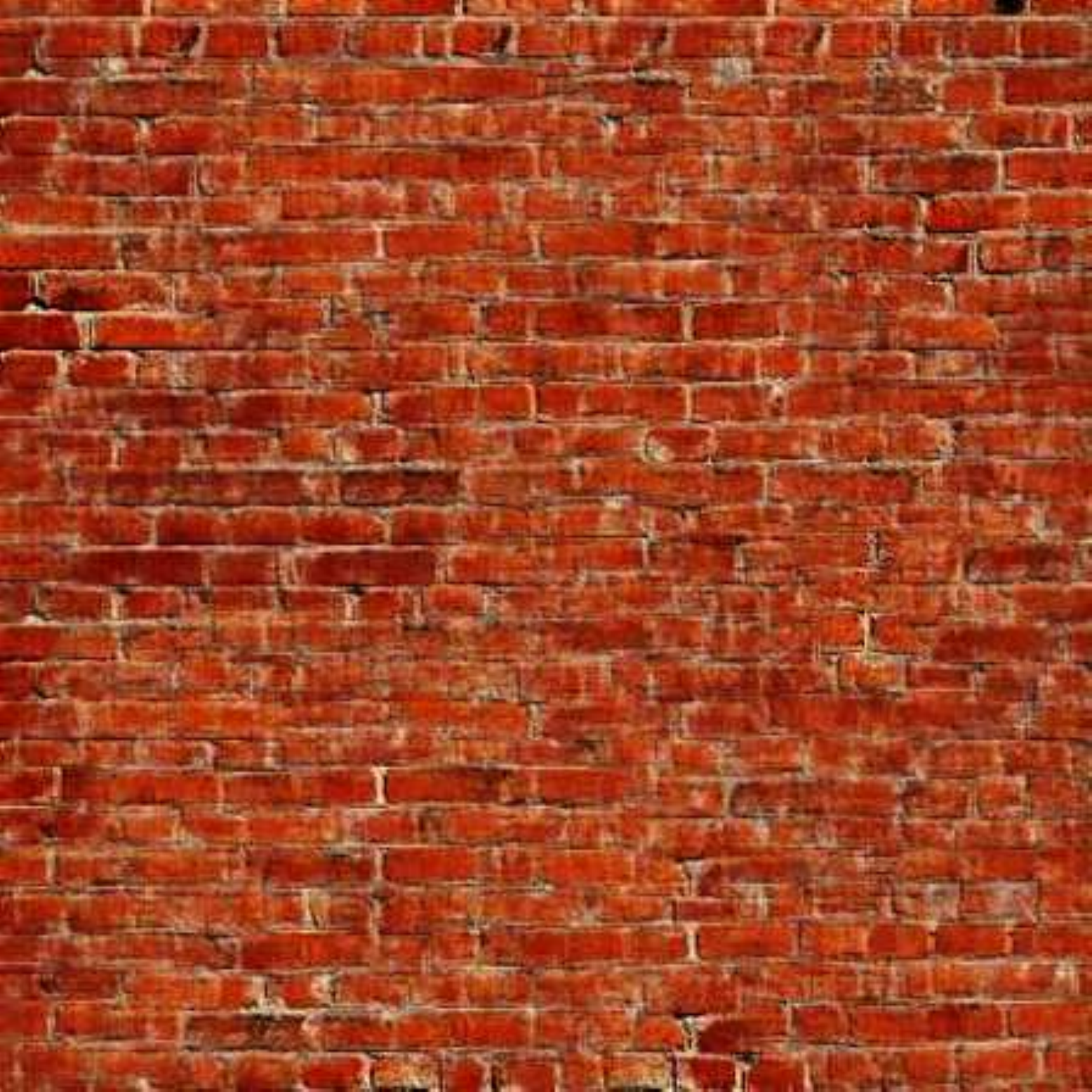}
	\end{center}
			\caption{Modeling texture patterns \cite{HanLu2016}. For each example, {Left:} the $224 \times 224$  observed image. {Right:} the $448 \times 448$ generated image. }
		\label{fig:texture1}
\end{figure}

Figure \ref{fig:texture1} shows the results of learning from texture images. We learn a separate model from each texture image. The images are collected from the Internet, and then resized to 224$\times$ 224. The synthesized images are 448 $\times$ 448.   

The factors $h$ at the top layer form a  $\sqrt{d} \times \sqrt{d}$ image, with each pixel following ${\rm N}(0, 1)$ independently. The  $\sqrt{d} \times \sqrt{d}$  image $h$ is then transformed to $X$ by the top-down ConvNet.  We use $d= 7^2$ in the learning stage for all the texture experiments. In order to obtain the synthesized image,  we randomly sample a 14 $\times$ 14 $h$ from N$(0, I)$, and then expand the learned network to generate the 448 $\times$ 448 synthesized image $g_\alpha(h)$. 

The training network is as follows. Starting from $7 \times 7$  image $h$, the network has 5 layers of convolution with $5 \times 5$ kernels, with an up-sampling factor of 2 at each layer. The number of channels in the first layer is 512, and is decreased by a factor 2 at each layer. The Langevin steps $l = 10$ with step size $s = .1$.

{\small
\begin{table}[h]
	\begin{center}{\small
		\begin{tabular}{|c|c|c|c|c|}
			\hline    experiment 	& $d=20$	& $d=60$ & $d=100$ & $d=200$	\\
			\hline    Ours		&	.0810 	& .0617 & .0549 & .0523	\\
			\hline    PCA 	&   .1038 	& .0820 & .0722 & .0621		\\ 		
			\hline 
		\end{tabular}	}	
		\caption{Reconstruction errors on testing images,  using our method and PCA. } 
		\label{tab:PCA}
	\end{center}
\end{table}
}

\begin{figure}[h]
	\centering
	\setlength{\fboxrule}{1pt}
	\setlength{\fboxsep}{0cm}
	\subfloat{
		\includegraphics[width=.045\linewidth]{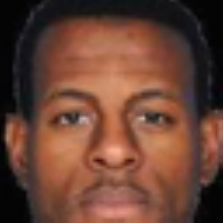}
		\includegraphics[width=.045\linewidth]{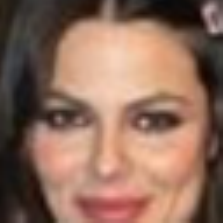}
		\includegraphics[width=.045\linewidth]{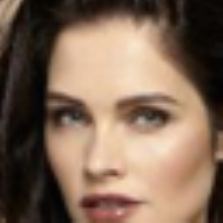}
		\includegraphics[width=.045\linewidth]{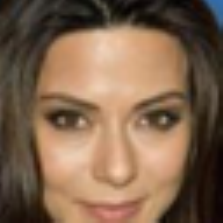}
		\includegraphics[width=.045\linewidth]{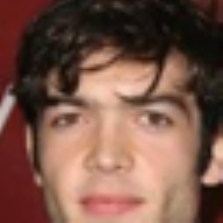}
		\includegraphics[width=.045\linewidth]{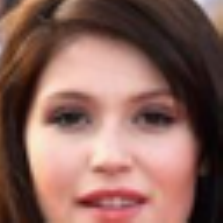}
		\includegraphics[width=.045\linewidth]{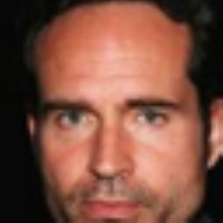}
		\includegraphics[width=.045\linewidth]{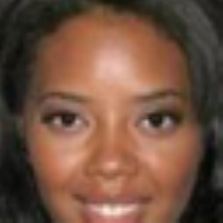}
		\includegraphics[width=.045\linewidth]{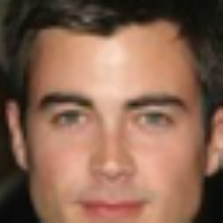}
		\includegraphics[width=.045\linewidth]{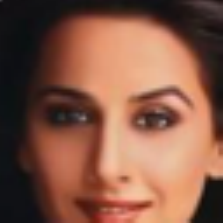}
	}\\[1px]
	
	\subfloat{
		\includegraphics[width=.045\linewidth]{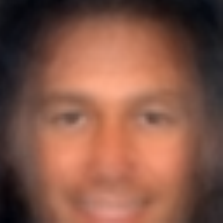}
		\includegraphics[width=.045\linewidth]{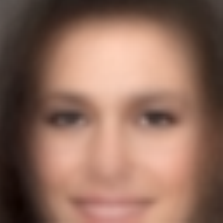}
		\includegraphics[width=.045\linewidth]{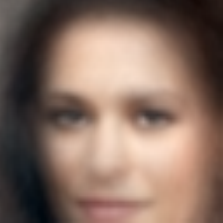}
		\includegraphics[width=.045\linewidth]{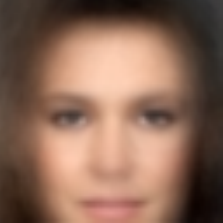}
		\includegraphics[width=.045\linewidth]{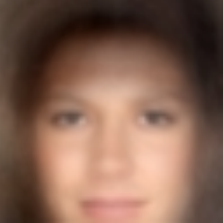}
		\includegraphics[width=.045\linewidth]{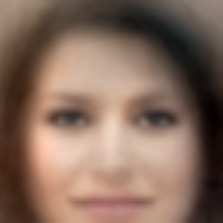}
		\includegraphics[width=.045\linewidth]{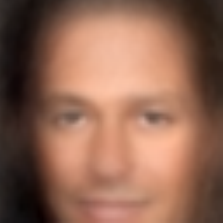}
		\includegraphics[width=.045\linewidth]{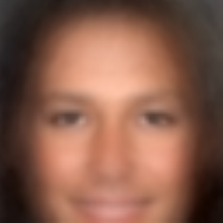}
		\includegraphics[width=.045\linewidth]{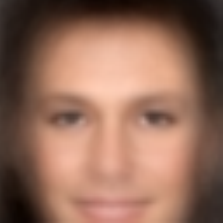}
		\includegraphics[width=.045\linewidth]{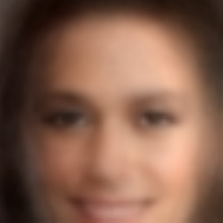}
	}\\[1px]
	
	\subfloat{
		\includegraphics[width=.045\linewidth]{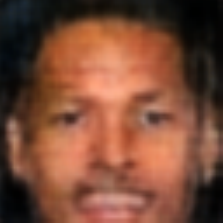}
		\includegraphics[width=.045\linewidth]{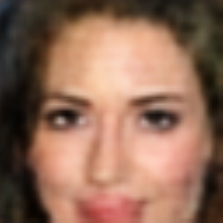}
		\includegraphics[width=.045\linewidth]{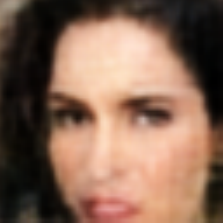}
		\includegraphics[width=.045\linewidth]{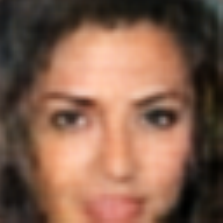}
		\includegraphics[width=.045\linewidth]{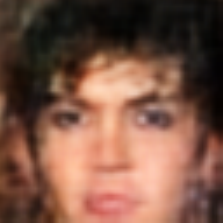}
		\includegraphics[width=.045\linewidth]{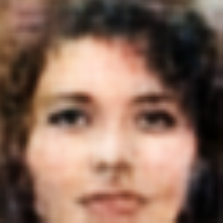}
		\includegraphics[width=.045\linewidth]{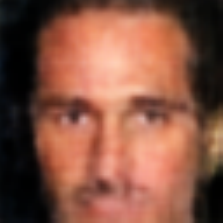}
		\includegraphics[width=.045\linewidth]{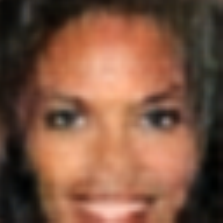}
		\includegraphics[width=.045\linewidth]{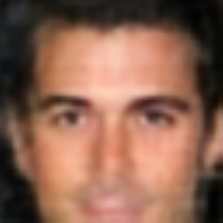}
		\includegraphics[width=.045\linewidth]{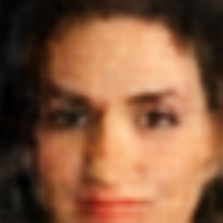}
	}	
	\caption{Comparison between \cite{HanLu2016} and PCA.  Row 1:  original testing images. Row 2: reconstructions by PCA eigenvectors learned from training images. Row 3:  Reconstructions  by the generative model learned from training images. $d = 20$ for both methods. }
	\label{fig:comp}
\end{figure}

 The generative model performs non-linear dimension reduction which can be more flexible than linear dimension reduction such as principal component analysis (PCA) or factor analysis. After learning the model from the training images,  we can evaluate  how well the learned model can generalize by computing the reconstruction errors on the testing images. We randomly select 1000 face images for training and 300 images for testing from CelebA dataset.  After learning, we infer the latent factors $h$ for each testing image using Langevin dynamics, and then reconstruct the testing image by $g_\alpha(h)$ using the inferred $h$ and the learned $\alpha$. Table~\ref{tab:PCA} shows the reconstruction error (measured by average per pixel difference relative to the range of the pixel intensities) of our method  as compared to PCA learning for different latent dimensions $d$. Figure~\ref{fig:comp} shows some reconstructed testing images. For PCA, we learn the $d$ eigenvectors from the training images, and then project the testing images on the learned eigenvectors for reconstruction.

\begin{figure}[h]
	\begin{center}
	
	  \includegraphics[width=.05\linewidth]{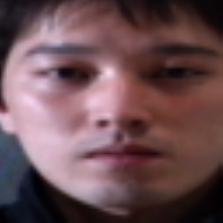}
	  \hspace{1mm}
	  \includegraphics[width=.05\linewidth]{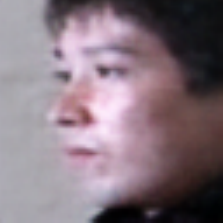}
		\includegraphics[width=.05\linewidth]{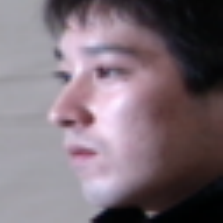}
		\hspace{1mm}
		\includegraphics[width=.05\linewidth]{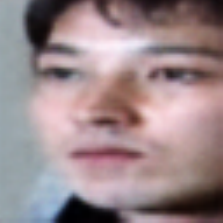}
		\includegraphics[width=.05\linewidth]{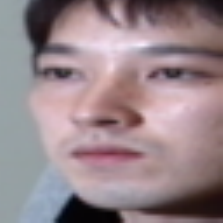}
		\hspace{1mm}
		\includegraphics[width=.05\linewidth]{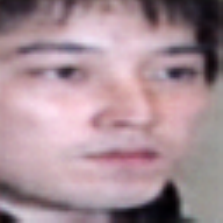}
		\includegraphics[width=.05\linewidth]{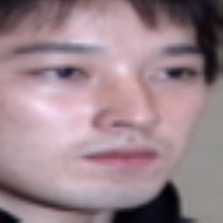}
		\hspace{1mm}
		\includegraphics[width=.05\linewidth]{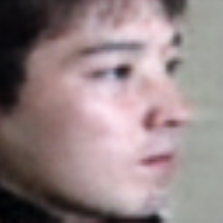}
		\includegraphics[width=.05\linewidth]{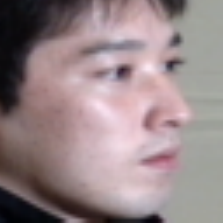}
		\\ \vspace{1mm}
		
		\includegraphics[width=.05\linewidth]{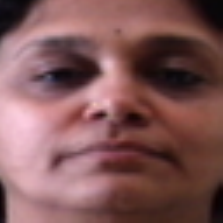}
	  \hspace{1mm}
	  \includegraphics[width=.05\linewidth]{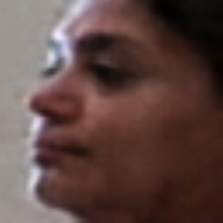}
		\includegraphics[width=.05\linewidth]{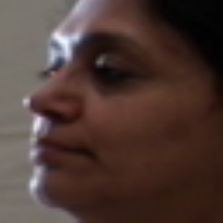}
		\hspace{1mm}
		\includegraphics[width=.05\linewidth]{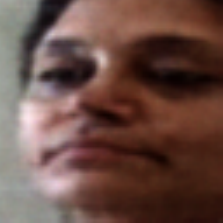}
		\includegraphics[width=.05\linewidth]{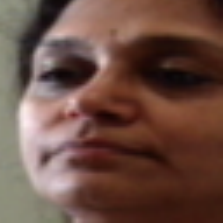}
		\hspace{1mm}
		\includegraphics[width=.05\linewidth]{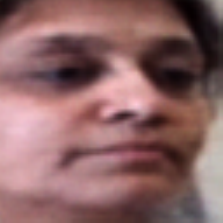}
		\includegraphics[width=.05\linewidth]{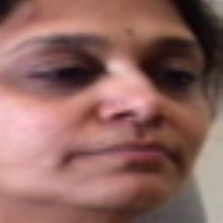}
		\hspace{1mm}
		\includegraphics[width=.05\linewidth]{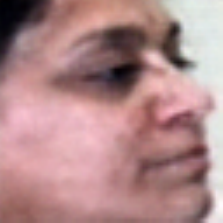}
		\includegraphics[width=.05\linewidth]{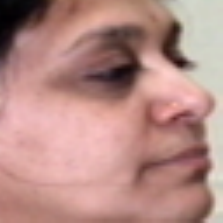}	
		\end{center}
\caption{Face rotation results on testing images \cite{tianshare}. First column: face image under standard pose ($0^\circ$). Second to fifth column: each pair shows the rotated face by our method (left) and the ground truth target (right).}
\label{fig:rotate}
\end{figure}

In our recent work \cite{tianshare}, we generalize the generative model for faces from multiple poses or views and learn the model from the Multi-PIE database \cite{Gross2010}. Let $X_i^{(j)}$ be the $j$-th view of the $i$-th subject, for $j = 1, ..., m$. We can model $X_i^{(j)} = g_{\alpha_j}(h_i)$, where different views share the same latent vector $h_i$, but they are generated by different $g_{\alpha_j}$. We can learn $(\alpha_j, j = 1, ..., m)$ using our learning algorithm. This enables us to change from one view to other views as illustrated by Figure \ref{fig:rotate}.

\section{Variational, adversarial and cooperative learning}

Both the descriptive model and the generative model involve intractable integrals. In the descriptive model, the normalizing constant is intractable. In the generative model, the marginal distribution of the observed signal is intractable. Consequently, the maximum likelihood learning algorithms of both models require MCMC sampling such as Langevin dynamics. To learn the descriptive model, we need to sample the synthesized examples. To learn the generative model, we need to sample the latent variables. It is possible to avoid MCMC sampling by variational and adversarial learning. It is also possible to speed up MCMC sampling by cooperative learning. 

\subsection{Variational auto-encoder} 

This subsection describes the variational learning of the generative model, where an inference model is learned to replace the MCMC sampling of the latent variables. 

The maximum likelihood learning of the generative model seeks to minimize the divergence $\KL(\P(X)\|q_\alpha(X))$, where $q_\alpha(X) = \int q(h)q_\alpha(X|h)dh$ is the marginal distribution that is intractable. The variational auto-encoder (VAE) \cite{KingmaCoRR13, RezendeICML2014, MnihGregor2014} changes the objective  to  
 \begin{eqnarray}
\min_\alpha \min_\phi  \KL(\P(X) \rho_\phi(h|X)\|q(h) q_\alpha(X|h)),  \label{eq:V11}
 \end{eqnarray}
where $\rho_\phi(h|X)$ is an analytically tractable approximation to $q_\alpha(h|X)$, and is called the inference model with parameter $\phi$. Compared to the maximum likelihood objective $\KL(\P(X)\|q_\alpha(X))$, which is the KL-divergence between the marginal distributions of $X$, the VAE objective is the KL-divergence between the joint distributions of $(h, X)$, i.e., $\P(X) \rho_\phi(h|X)$ and $q_\alpha(h, X) = q(h) q_\alpha(X|h)$, which is tractable because it does not involve the marginal $q_\alpha(X)$. The VAE objective is an upper bound of the maximum likelihood objective 
\begin{eqnarray}
\KL(\P(X) \rho_\phi(h|X)\|q_\alpha(h, X))  =  \KL(\P(X)\|q_\alpha(X)) + \KL(\rho_\phi(h|X)\|q_\alpha(h|X)).  \label{eq:V1}
\end{eqnarray}
The accuracy of the VAE objective as an approximation to the maximum likelihood objective depends on the accuracy of the inference model $\rho_\phi(h|X)$ as an approximation to the true posterior distribution $q_\alpha(h|X)$. 

\begin{figure}[h]
	\begin{center}
		\includegraphics[width=.3\linewidth]{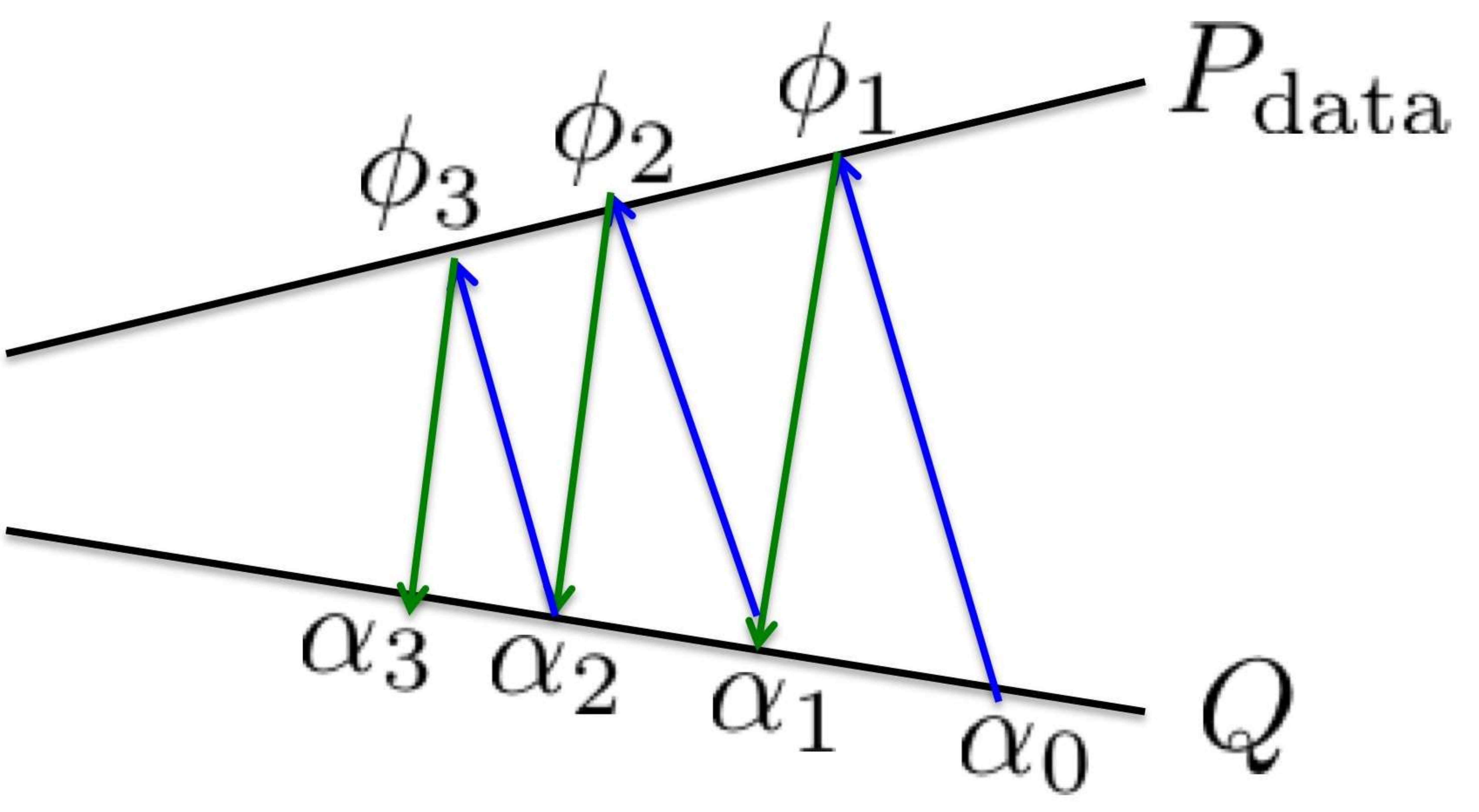}		
		\caption{VAE as alternating projection, where the straight lines illustrate the families of $\P$ and $Q$ respectively, and each point is a distribution parametrized by $\phi$ or $\alpha$. }
		\label{fig:VAE}
	\end{center}	
\end{figure}

For simplicity and slightly abusing the notation, write $\P(h, X) = \P(X) \rho_\phi(h|X)$, where $\P$ here is understood as the distribution of the complete data $(h, X)$, with $h$ imputed by $\rho_\phi(h|X)$, and $Q(h, X) = q(h) q_\alpha(X|h)$. The VAE is 
\begin{eqnarray}
\min_\alpha \min_\phi \KL(\P\|Q).
\end{eqnarray} 
We can think of VAE from the perspective of alternating projection. (1) Fix $\alpha$, find $\phi$ by minimizing $\KL(\P\|Q)$. This is to project the current $Q$ onto the family of $\P$. (2) Fix $\phi$, find $\alpha$ by minimizing $\KL(\P\|Q)$. This is to project the current $\P$ onto the family of $Q$. Compared to the EM algorithm, projection (1) corresponds to the E-step to impute the missing data in the form of $\rho_\phi(h|X)$, and projection (2) corresponds to the M-step to fit the complete model $q(h) q_\alpha(X|h)$. The basic idea is illustrated by Figure \ref{fig:VAE}. 

 The problem (\ref{eq:V11})  is equivalent to maximizing 
\begin{eqnarray}
   && \E_{\P} \big[ \E_\phi[\log q_\alpha(h, X)]  +{\rm entropy}(\rho_\phi(h|X)) \big] \label{eq:V2} \\
   &=& \E_{\P} \big[ \E_\phi[\log q_\alpha(X|h)]  - \KL(\rho_\phi(h|X)\|q(h))\big] \label{eq:V3}
\end{eqnarray}
where $\E_\phi$ denotes the expectation with respect to $\rho_\phi(h|X)$, and $\E_{\P}$ can be computed by averaging over the training examples. In (\ref{eq:V2}) and (\ref{eq:V3}), we have $q_\alpha(h, X)$ and $q_\alpha(X|h)$, as a result of merging $q_\alpha(X)$ and $q_\alpha(h|X)$ in (\ref{eq:V1}), and both $q_\alpha(h, X)$ and $q_\alpha(X|h)$ are computationally tractable. If $\rho_\phi(h|X) = q_{\alpha}(h|X)$, then maximizing (\ref{eq:V2}) with respect to $\alpha$ becomes the EM algorithm. 

 One popular choice of $\rho_\phi(h|X)$ is ${\rm N}(\mu_\phi(X), \sigma^2_\phi(X))$, where both $\mu_\phi(X)$ and $\sigma^2_\phi(X)$ can be represented by bottom-up neural networks with parameter $\phi$. 
 
 \begin{figure}[h]
	\begin{center}
		\includegraphics[width=.5\linewidth]{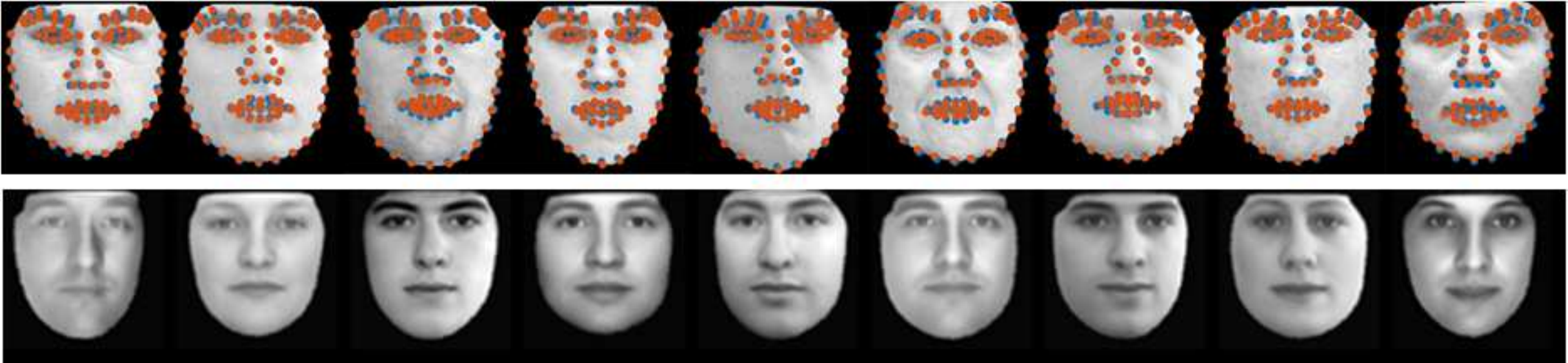}\hspace{0.5mm}			
		\caption{Top Row: training images with landmarks. Bottom Row: synthesized  images generated by the learned AAM model \cite{tianaam}.}
		\label{fig:AAM_train}
	\end{center}	
\end{figure}

In our recent work \cite{tianaam}, we show that VAE can replicate the active appearance model (AAM) \cite{cootes2001active}. Our experiments were inspired by a recent paper \cite{chang2017code} that studied neurons in the middle lateral (ML) / middle fundus (MF) and anterior medial (AM) areas of the primate brain that are responsible for face recognition. Specifically, \cite{chang2017code} recorded how these neurons respond to face stimuli generated by a pre-trained AAM model. We show that the observed properties of neurons' responses can be qualitatively replicated by VAE. The AAM model has an explicit shape representation in the form of landmarks, where the landmarks follow a shape model learned by principal component analysis. The faces can be aligned based on the landmarks, and the aligned faces follow an appearance model learned by another principal component analysis. The learning of the shape and appearance models require the landmarks in the training data. Figure \ref{fig:AAM_train} shows examples of  face images to train AAM, and the synthesized face images from the trained AAM. 

\begin{figure}[h]
	\begin{center}
		\includegraphics[width=.3\linewidth]{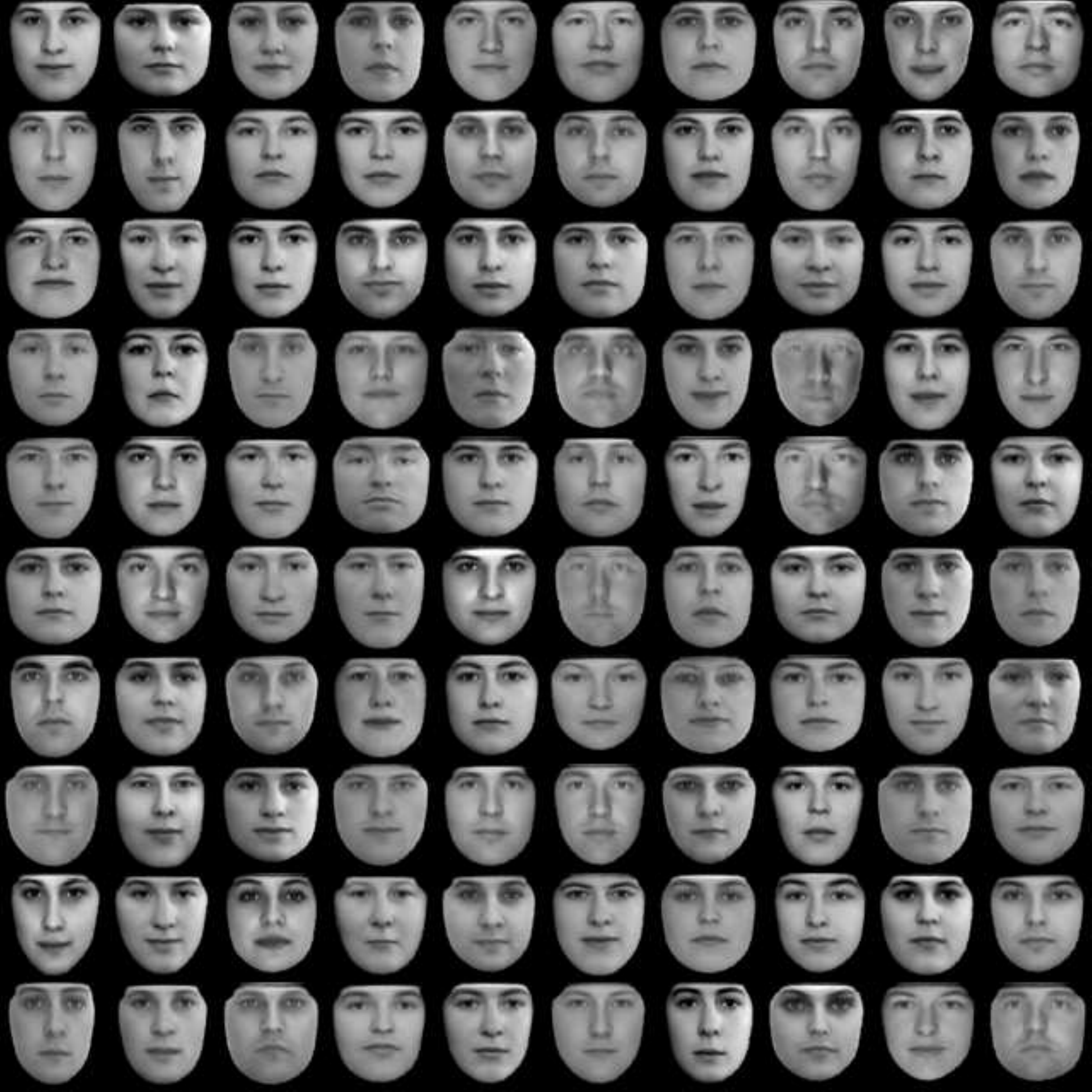}\hspace{0.5mm}
		\includegraphics[width=.3\linewidth]{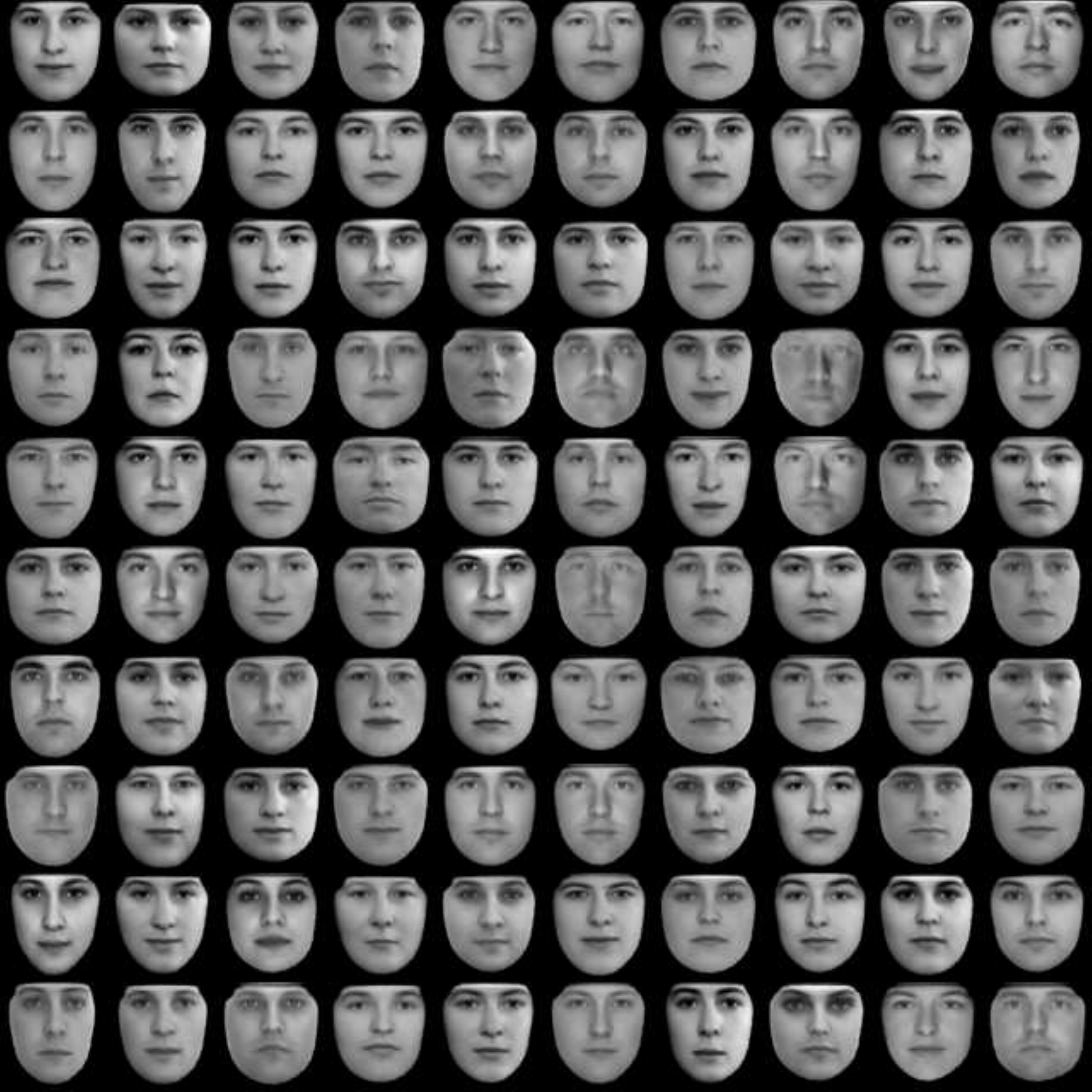}
				\includegraphics[width=.3\linewidth]{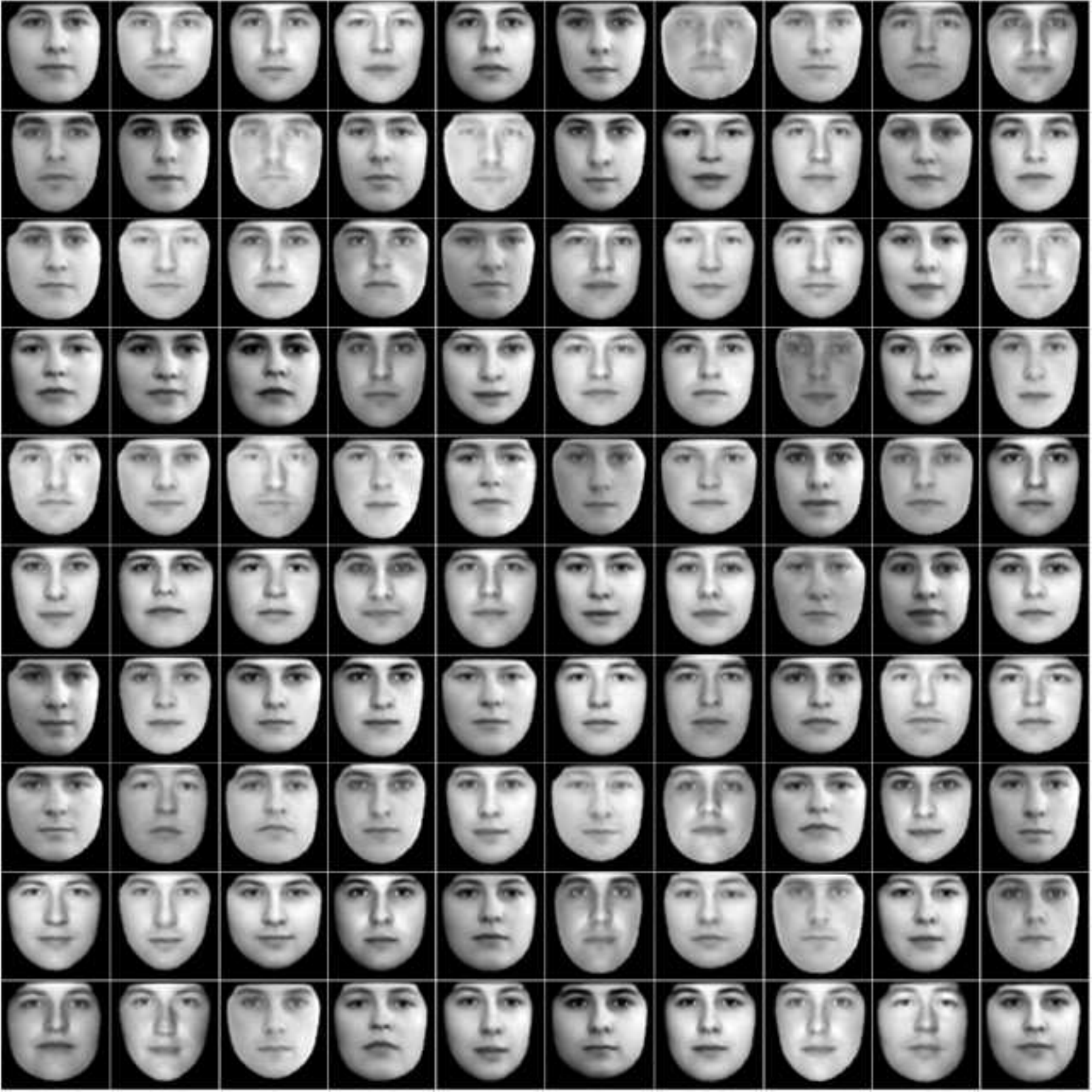}				
		\caption{Replicating AAM by VAE \cite{tianaam}. Left: test faces generated by AAM. Middle: reconstructed faces by the learned generative model. Right: synthesized images generated by the learned generative model. }
		\label{fig:AAM_rep}
	\end{center}
\end{figure}

After learning the AAM model, we generate $20,000$ face images from the learned model. We then learn a VAE model from these images without the landmarks. Figure \ref{fig:AAM_rep} displays test images generated by the AAM model,  their corresponding reconstructions by the learned VAE model, and the synthesized images generated by the learned VAE model. 

In \cite{chang2017code} the authors show that the responses from face patches ML/MF and AM have strong linear relationship with the shape and appearance variables in the original AAM model, where the responses of some neurons are highly correlated with the shape variables while the responses of other neurons are highly correlated with the appearance variables. In fact, one can further recover the original face images from the responses of these neurons, by linearly transforming the neurons' responses to the shape and appearance variables of the AAM, and then generating the image by the AAM variables. Apparently the neurons' responses form a code of the input face image that captures both the shape and appearance information of the input image. We show that the code learned by VAE, i.e., $\mu_\phi(X)$, has very strong linear relationship with the shape and appearance variables in AAM that generates $X$. The $R^2$ measure is over $96\%$. The biological observations found by \cite{chang2017code} can be qualitatively reproduced by VAE. Even though the AAM model is highly non-linear due to shape deformation, the generative model has no difficulty replicating the AAM model without the supervision in the form of landmarks on the faces.

\subsection{Adversarial contrastive divergence} 

This subsection describes the adversarial learning of the descriptive model, where a generative model is learned to replace the MCMC sampling of the descriptive model. 

\begin{figure}[h]
	\begin{center}
		\includegraphics[width=.25\linewidth]{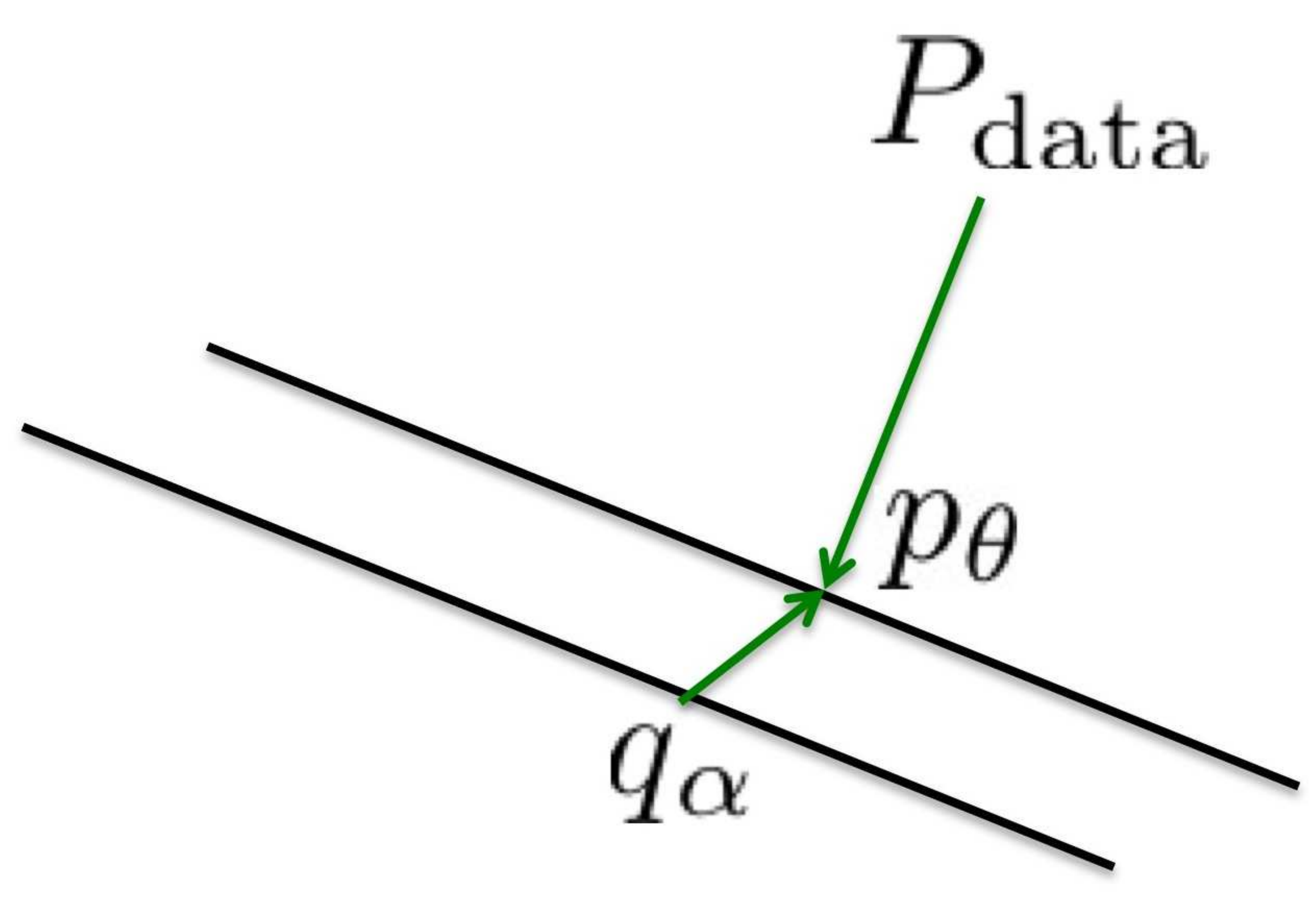}		
		\caption{Adversarial contrastive divergence. The straight lines illustrate the families of the descriptive and generative models, and each point is a probability distribution. The generative model seeks to approximate the descriptive model, while the descriptive model seeks to get close to the data distribution in contrast to the generative model. }
		\label{fig:ACD}
	\end{center}	
\end{figure}

The maximum likelihood learning of the descriptive model seeks to minimize the divergence $\KL(\P(X)\|p_\theta(X))$, where the normalizing constant $Z(\theta)$ in $p_\theta$ is intractable. Recently \cite{Bengio2016} and \cite{dai2017calibrating} proposed to train the descriptive model  $p_\theta$ and the generative model $q_\alpha$ jointly, which amounts to modifying the objective  to 
\begin{eqnarray} 
   \min_\theta \max_\alpha  [  \KL(\P(X)\|p_\theta(X)) - \KL(q_\alpha(X)\|p_\theta(X))]. \label{eq:A1}
\end{eqnarray}
See Figure \ref{fig:ACD} for an illustration.  By maximizing over $\alpha$, we minimize $\KL(q_\alpha(X)\|p_\theta(X))$, so that the objective function in (\ref{eq:A1}) is a good approximation to $\KL(\P\|p_\theta)$. Because of the minimax nature of the objective, the learning is adversarial, where $\theta$ and $\alpha$ play a minimax game. While the generative model seeks to get close to the descriptive model, the descriptive model seeks to get close to the data distribution and to get away from the generative model. That is, the descriptive model can be considered a critic of the generative model by comparing it to the data distribution. 

The objective (\ref{eq:A1}) contrasts  interestingly with the objective for variational learning in (\ref{eq:V1}). In the variational objective, we upper bound $\KL(\P\|q_\alpha)$ by adding another KL-divergence,  so that we minimize over both $\alpha$ and $\phi$. However, in the adversarial objective (\ref{eq:A1}), we lower bound $\KL(\P\|p_\theta)$ by subtracting from it another KL-divergence, hence we need to find its saddle point. Thus the sign in front of the second KL-divergence determines whether it is variational learning or adversarial learning. 

The adversarial objective (\ref{eq:A1}) is also a form of contrastive divergence, except that the synthesized examples are provided by the generative model $q_\alpha$ directly, instead of being obtained by running a finite-step MCMC from the observed examples. We may call (\ref{eq:A1}) the adversarial contrastive divergence. It is equivalent to 
\begin{eqnarray} 
   \min_\theta \max_\alpha \left[ \E_{\P}[U_\theta(X)] - \E_{\theta}[U_\theta(X)] + {\rm entropy}(q_\alpha) \right], \label{eq:A2}
\end{eqnarray}
which is the form proposed by  \cite{dai2017calibrating}. In this form, the $\log Z(\theta)$ term is canceled out, so that we do not have to deal with this intractable term. 

However, the entropy term $ {\rm entropy}(q_\alpha)$ or the second KL-divergence in (\ref{eq:A1}) is not in closed form, and still needs approximation. We can again use the variational approach to approximate  $\KL(q_\alpha(X)\|p_\theta(X))$ by 
\begin{eqnarray}
\KL(q_\alpha(X)\|p_\theta(X)) + \KL(q_\alpha(h|X)\|\rho_\phi(h|X)) = \KL(q_\alpha(h, X)\|p_\theta(X) \rho_\phi(h|X)),  \label{eq:A3}
\end{eqnarray}
where $\rho_\phi(h|X)$ is again a learned inference model. This leads to the method used by \cite{dai2017calibrating}. Again we only need to deal with the tractable joint model $q_\alpha(h, X)$. Thus the learning problem becomes 
\begin{eqnarray} 
   \min_\theta \max_\alpha \max_\phi  [\KL(\P(X)\|p_\theta(X)) -  \KL(q_\alpha(h, X)\|p_\theta(X) \rho_\phi(h|X))]. \label{eq:A4}
\end{eqnarray}
There are three networks that need to be learned, including the descriptive model $p_\theta$, the generative model $q_\alpha$, and the inference model $\rho_\phi$. Write $\P(h, X) = \P(X) \rho_\phi(h|X)$, $Q(h, X) = q(h) q_\alpha(X|h)$, and $P(h, X) = p_\theta(X) \rho_\phi(h|X)$. The above objective is 
\begin{eqnarray}
 \min_\theta \max_\alpha \max_\phi [\KL(\P\|P) - \KL(Q\|P)].
 \end{eqnarray}

 Compared to the variational learning in (\ref{eq:V1}), $\rho_\phi(h|X)$ appears on the left side of KL-divergence in (\ref{eq:V1}), but it appears on the right side of KL-divergence in (\ref{eq:A4}). The learning of $\rho_\phi(h|X)$ is from the synthesized data generated by $q_\alpha(h, X)$ instead of real data. This is similar to the sleep phase of the wake-sleep algorithm \cite{hinton1995wake}.

\begin{figure}[h]
	\centering
	\includegraphics[width=.35\linewidth]{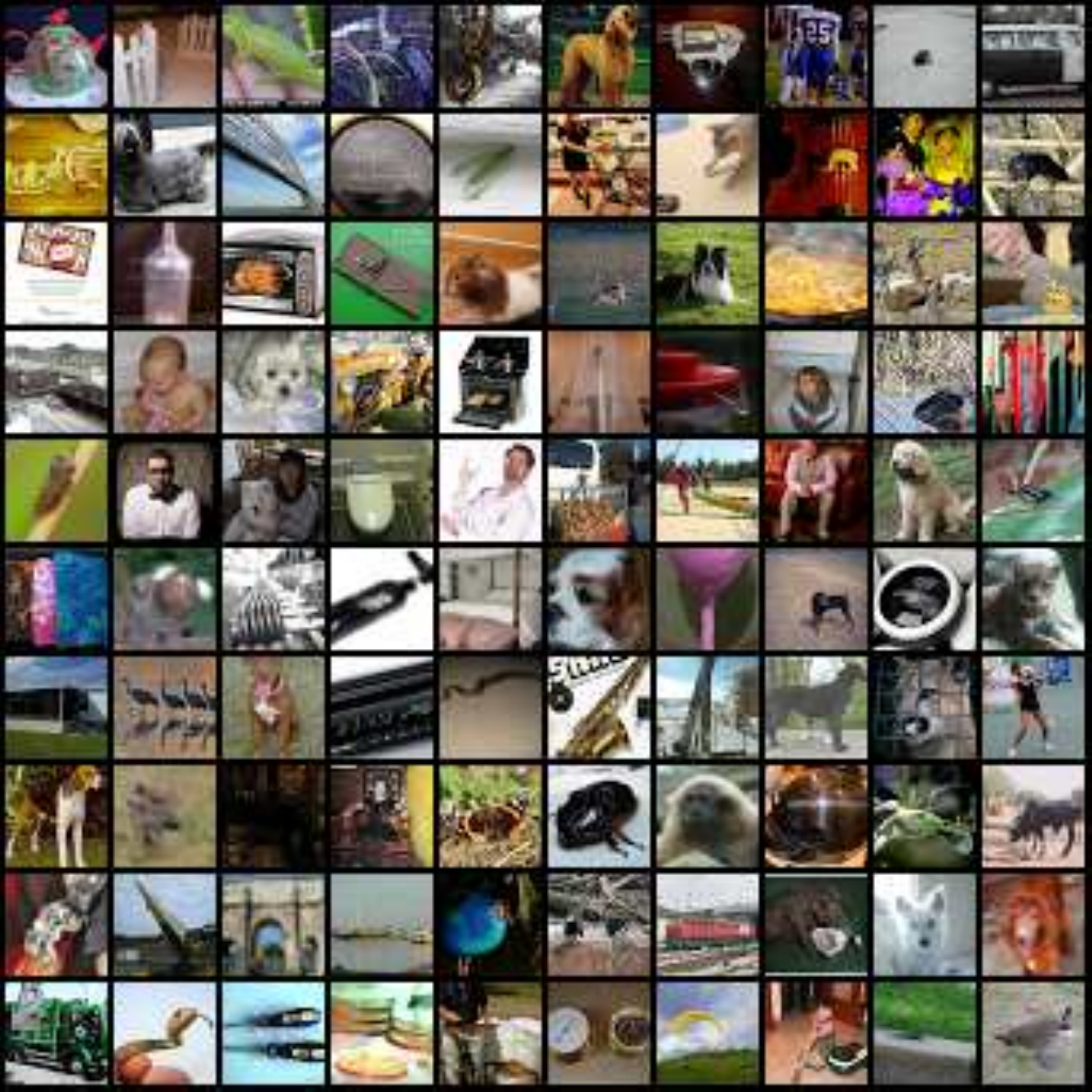} \hspace{2mm}
	\includegraphics[width=.35\linewidth]{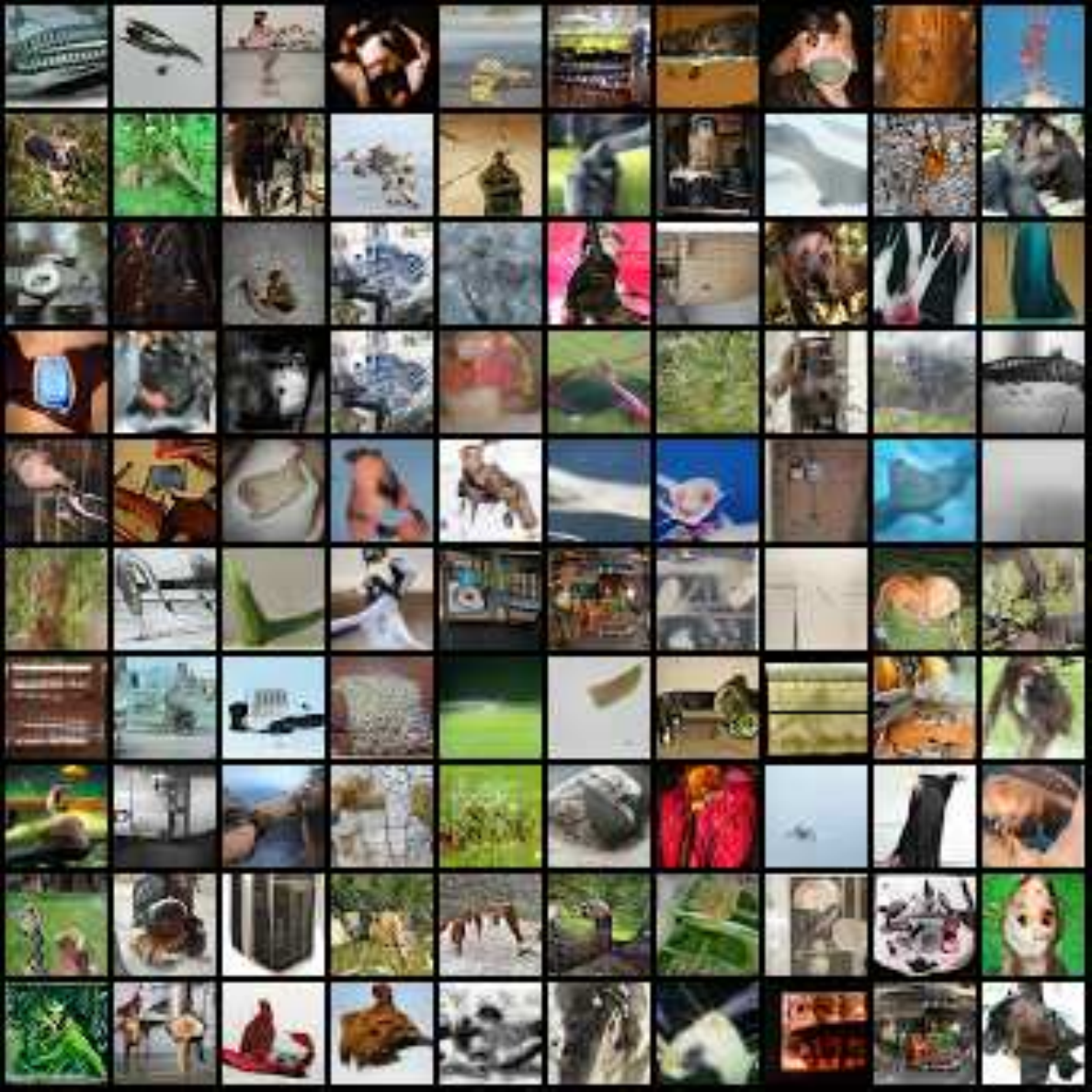}
	\caption{Learning the models from the ImageNet dataset. Left: random samples of training examples. Right: synthesized examples generated by the learned generative model.  }	
	\label{fig:syn_imagenet}
\end{figure}

We train the three nets  on the down-sampled 32x32 imageNet dataset \cite{deng2009imagenet} (roughly 1 million images). For the generative model, starting from the latent vector $h$ of 100 dimensions,  we use 5 layers of  kernels of stride 2,  where the sizes of kernels of the first 4 layers are $4 \times 4$, and the size of the kernels of the bottom layer is $3\times 3$. The numbers of channels at these layers are 512, 512, 256, 128, 3 respectively. Each  layer is followed by batch normalization and ReLU non-linearity, except the last layer where tanh is used. For the inference model, we use the mirror structure as the generative model. We build the last layer separately to model the posterior mean and variance. For the descriptive model,  we use the same structure as the inference net.  

Figure  \ref{fig:syn_imagenet} displays the learning results, where the left panel shows randomly selected training examples and the right panel shows the random examples generated by the learned generative model. 

Another possibility of adversarial contrastive divergence learning is to learn a joint energy-based model $p_\theta(h, X)$ by 
\begin{eqnarray}
  \min_\theta \min_\phi \max_\alpha[\KL(\P(X) \rho_\phi(h|X)\|p_\theta(h, X)) - \KL(q(h)q_\alpha(X|h)\|p_\theta(h, X))]. 
\end{eqnarray}

\subsection{Integrating variational and adversarial learning}

We can integrate or unify the variational and adversarial learning methods.

\begin{figure}[h]
	\begin{center}
		\includegraphics[width=.4\linewidth]{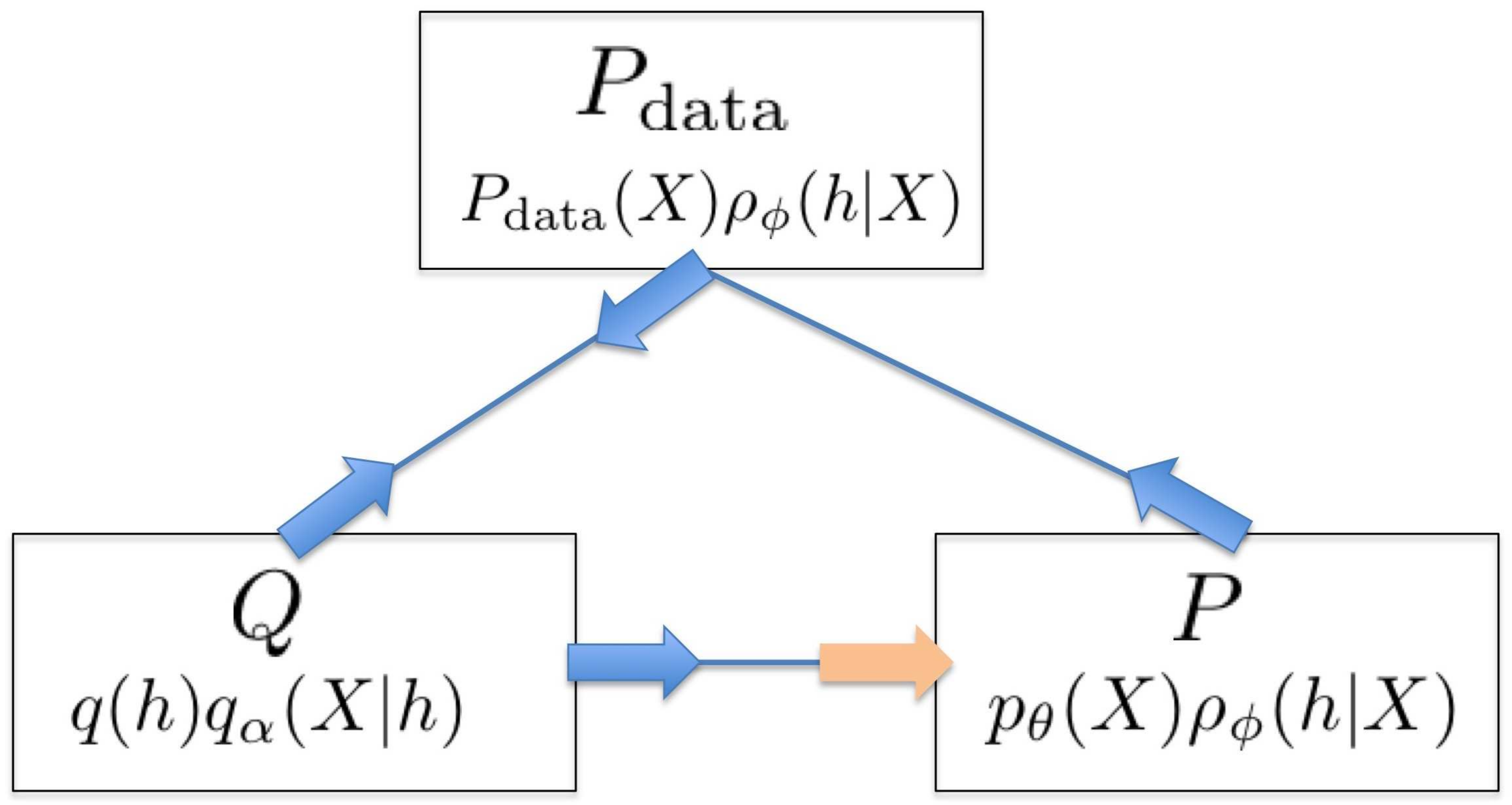}		
		\caption{Divergence triangle \cite{Han2018}. The generative model seeks to get close to the data distribution as well as the descriptive model. The descriptive model seeks to get close to the data distribution and get away from the generative model. }
		\label{fig:DT}
	\end{center}	
\end{figure}

Following the notation of previous subsections, write $\P(h, X) = \P(X) \rho_\phi(h|X)$, $P(h, X) = p_\theta(X) \rho_\phi(h|X)$, and $Q(h, X) = q(h) q_\alpha(X|h)$. It has been noticed by the recent work \cite{Han2018} that the variational objective $\KL(\P\|Q)$ and the adversarial objective $\KL(\P\|P) - \KL(Q\|P)$ can be combined into
\begin{eqnarray}
\max_\theta \min_\alpha \min_\phi [\KL(\P\|Q) + \KL(Q\|P) - \KL(\P\|P) ], 
\end{eqnarray} 
which is in the form of a triangle formed by $\P$, $P$, and $Q$. See Figure \ref{fig:DT} for an illustration. As shown by \cite{Han2018}, one can learn the descriptive model, the generative model, and the inference model jointly using the above objective.

\begin{figure}[h]
	\begin{center}
		\includegraphics[width=.32\linewidth]{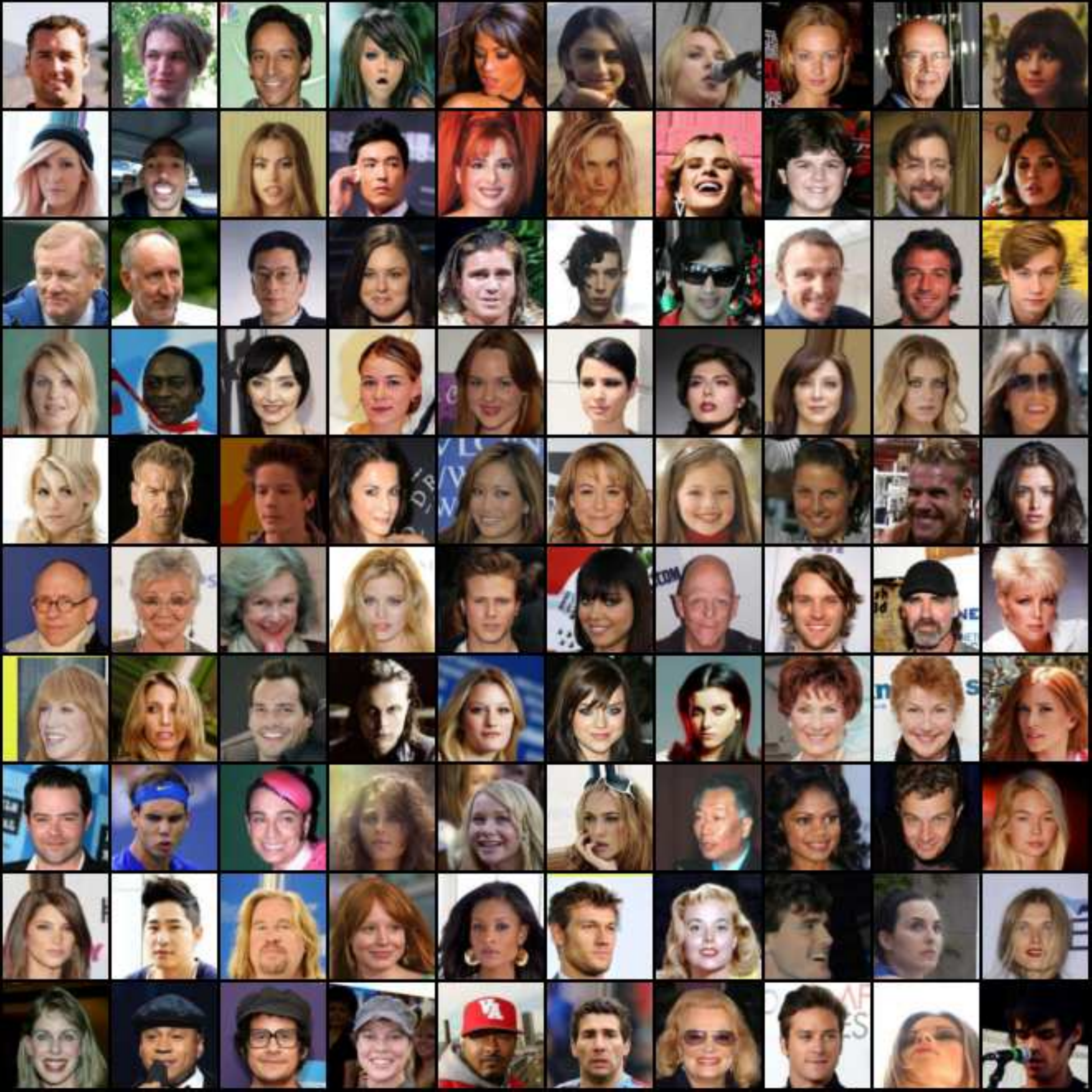}
		\includegraphics[width=.32\linewidth]{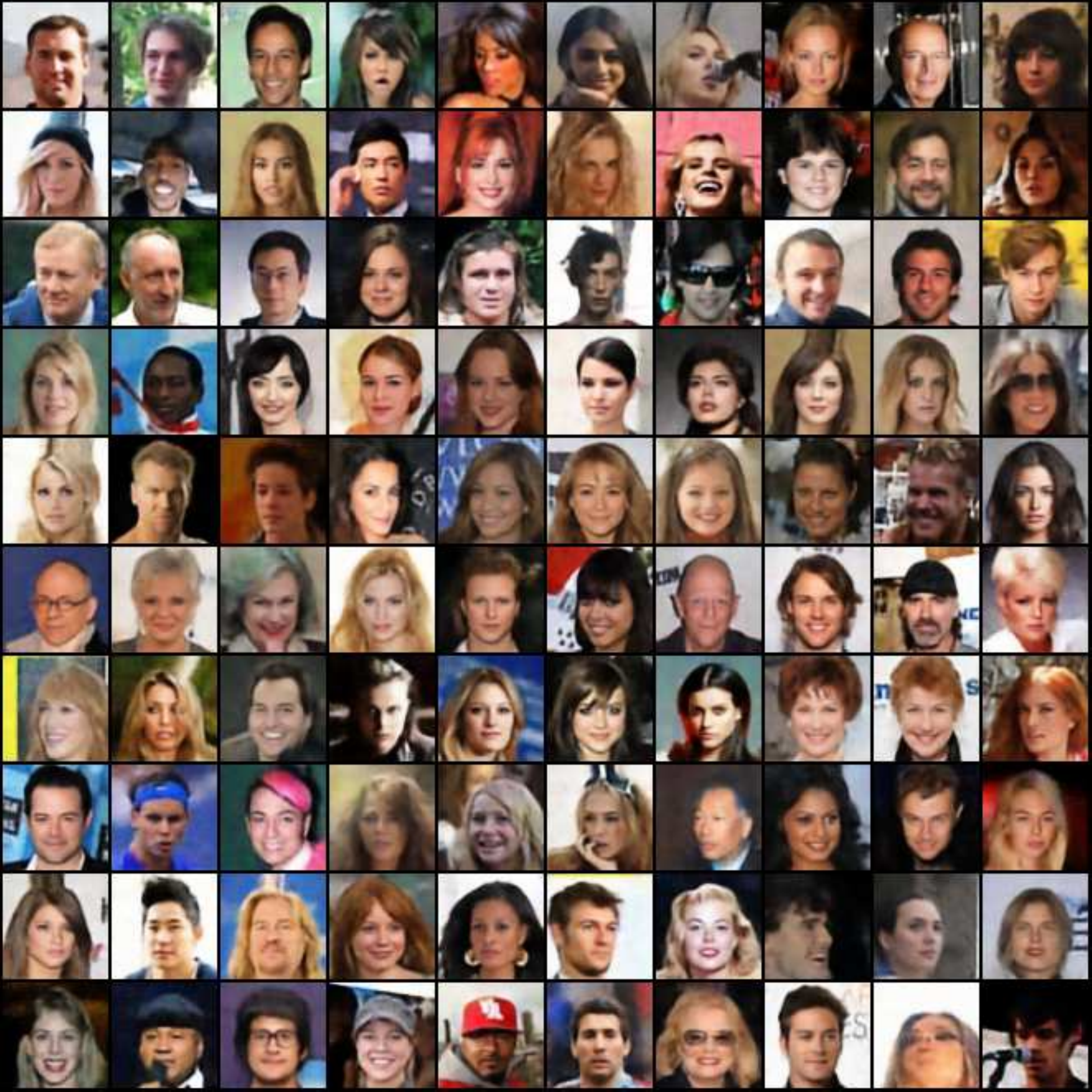}	
		\includegraphics[width=.32\linewidth]{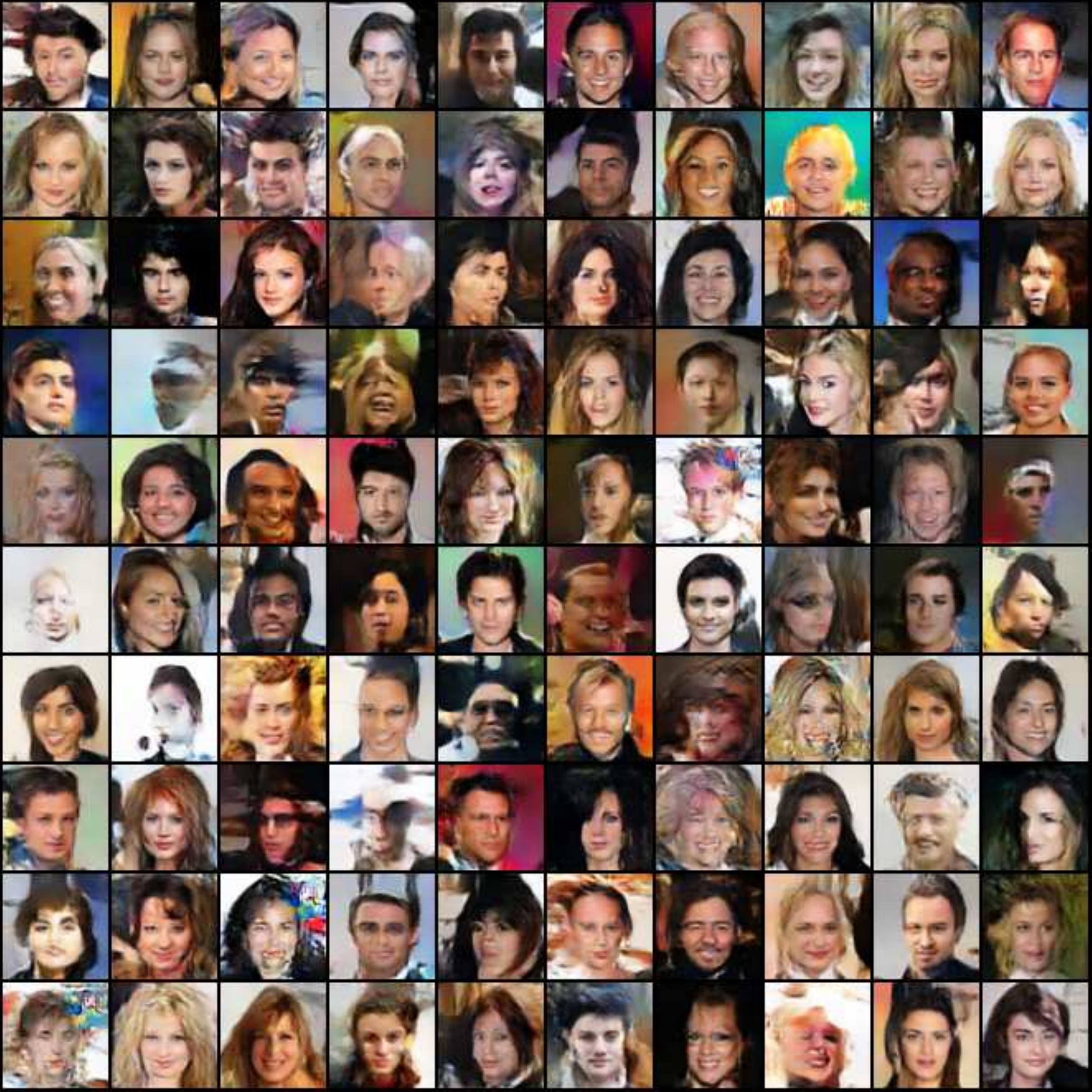}\\ 
		\caption{Learning the models from CelebA dataset \cite{Han2018}. From left to right: original images, reconstructed images, and generated images. }
		\label{fig:triangle}
	\end{center}
\end{figure}

Figure \ref{fig:triangle} displays an example in \cite{Han2018} where the models are learned from the CelebA dataset. The left panel shows some random training examples, the middle panel shows the corresponding reconstructed examples by the learned inference model, and the right panel shows some examples generated by the learned generative model. 

\subsection{Cooperative learning} \label{sect:c}

This subsection describes the cooperative training of the descriptive and generative models which jumpstart each other's MCMC sampling. 

 We can  learn the descriptive model and the generative model separately, and we have been able to scale up the learning to big datasets. However, the separate learning algorithms can still be slow due to MCMC sampling. Recently we discover that we can speed up the learning of the two models by coupling the  two maximum likelihood learning algorithms  into a cooperative  algorithm that we call the CoopNets algorithm \cite{CoopNets2016, coopnets2018}. It is based on the following two key observations. (1) The generative model can generate examples directly, so we can use it as an approximate sampler of the descriptive model. (2) The generative model can be learned more easily if the latent factors are known, which is the case with the synthesized examples. 

{\em Generative model as a sampler}. The generative model can serve as an approximate sampler of the descriptive model. To sample from the descriptive model, we can initialize the synthesized examples by generating examples from the generative model. We first generate $\hh_i \sim \N(0, I_d)$, and then generate $\hX_i = g(\hh_i; \alpha) + \epsilon_i$, for $i = 1, ..., \tn$. If the current generative model $q_\alpha$ is close to the current descriptive model $p_\theta$,  then the generated $\{\hX_i\}$ should be a good initialization for sampling from the descriptive model, i.e., starting from the $\{\hX_i, i = 1, ..., \tn\}$, we run Langevin dynamics  for $l$ steps to get  $\{\tX_i, i = 1, ..., \tn\}$, which are revised versions of $\{\hX_i\}$.  These $\{\tX_i\}$ can be used as the synthesized examples from the descriptive model.  We can then update $\theta$ in the same way as we learn the descriptive model. 

{\em MCMC teaching}. The descriptive model can teach the generative model via MCMC.  In order to update $\alpha$ of the generative model, we treat the $\{\tX_i, i = 1, ..., \tn\}$ produced by the above procedure as the training data for the generative model. Since these $\{\tX_i\}$ are obtained by the Langevin  dynamics initialized from  $\{\hX_i, i = 1, ..., \tn\}$, which are generated by the generative model with known latent factors $\{\hh_i, i = 1, ..., \tn\}$, we can update $\alpha$ by learning from $\{(\hh_i, \tX_i), i = 1, ..., \tn\}$, which is a supervised learning problem, or more specifically, a non-linear regression of $\tX_i$ on $\hh_i$. At $\alpha^{(t)}$,  the latent vector  $\hh_i$ generates and thus reconstructs the initial example  $\hX_i$. After updating $\alpha$, we want $\hh_i$ to reconstruct the revised example $\tX_i$. That is, we revise $\alpha$ to absorb the MCMC transition from $\hX_i$ to $\tX_i$ for sampling the descriptive model, so that the generative model shifts its density from $\{\hX_i\}$ to $\{\tX_i\}$. The left diagram in (\ref{eq:m}) illustrates the basic idea. \begin{eqnarray}
\begin{tikzpicture}
  \matrix (m) [matrix of math nodes,row sep=3em,column sep=4em,minimum width=2em]
  {
     \hh_i & \\
      \hX_i & \tX_i \\};
  \path[-stealth]
    (m-1-1) edge [double] node [left] {$\alpha^{(t)}$} (m-2-1)
        (m-1-1) edge [double] node [right] {$\;\;\alpha^{(t+1)}$} (m-2-2)      
            (m-2-1) edge [dashed]  node [below] {$\theta^{(t)}$} (m-2-2);
                     \end{tikzpicture} 
            \begin{tikzpicture}
  \matrix (m) [matrix of math nodes,row sep=3em,column sep=4em,minimum width=2em]
  {
     \hh_i & \th_i \\
      \hX_i & \tX_i \\};
  \path[-stealth]
      (m-1-1) edge  [dashed]  node [above] {$\alpha^{(t)}$} (m-1-2)
    (m-1-1) edge [double] node [left] {$\alpha^{(t)}$} (m-2-1)
        (m-1-2) edge [double]  node [right] {$\alpha^{(t+1)}$} (m-2-2)
          (m-2-1) edge  [dashed]   node [below] {$\theta^{(t)}$} (m-2-2);
            \end{tikzpicture}      
            \label{eq:m} 
            \end{eqnarray}
In the two diagrams in (\ref{eq:m}),  the double-line arrows indicate generation and reconstruction by the generative model, while the dashed-line arrows indicate Langevin dynamics for MCMC sampling and inference in the two models. 
The diagram on the right in (\ref{eq:m}) illustrates a more rigorous method, where we initialize the Langevin inference  of $\{h_i, i = 1, ..., \tn\}$  from $\{\hh_i\}$, and  then update $\alpha$  based on $\{(\th_i, \tX_i), i = 1, ..., \tn\}$. The diagram on the right shows how the two models jumpstart each other's MCMC. 

 The learning of the descriptive model is based on the modified contrastive divergence, 
\begin{eqnarray} 
{\rm KL}(\P\|p_\theta) - {\rm KL}(M_\theta q_\alpha\|p_\theta), \label{eq:MCD}
\end{eqnarray}
where  $q_\alpha$ provides the initialization of the finite-step MCMC, whose transition kernel is denoted $M_\theta$, and $M_\theta q_\alpha$ denotes the marginal distribution obtained after running $M_\theta$ from $q_\alpha$. The learning of the generative model is based on how $M_\theta q_\alpha$ modifies $q_\alpha$,  and is accomplished by  $\min_{q_\alpha} {\rm KL}(M_\theta q_{\alpha^{(t)}}\| q_\alpha)$.  In the idealized case of infinite capacity of $q_\alpha$ so that the KL-divergence can be minimized to zero, the learned $q_\alpha$ will satisfy $q_\alpha = M_\theta q_\alpha$, i.e., $q_\alpha$ is the stationary distribution of $M_\theta$. But the stationary distribution of $M_\theta$ is nothing but $p_\theta$. Thus the learned $q_\alpha$ will be the same as $q_\theta$. Then the second KL-divergence in (\ref{eq:MCD}) will become zero, and the learning of the descriptive model is to minimize ${\rm KL}(\P\|p_\theta)$, which is maximum likelihood.

\begin{figure}[h]
	\centering
	\includegraphics[height=.4\linewidth, width=.4\linewidth]{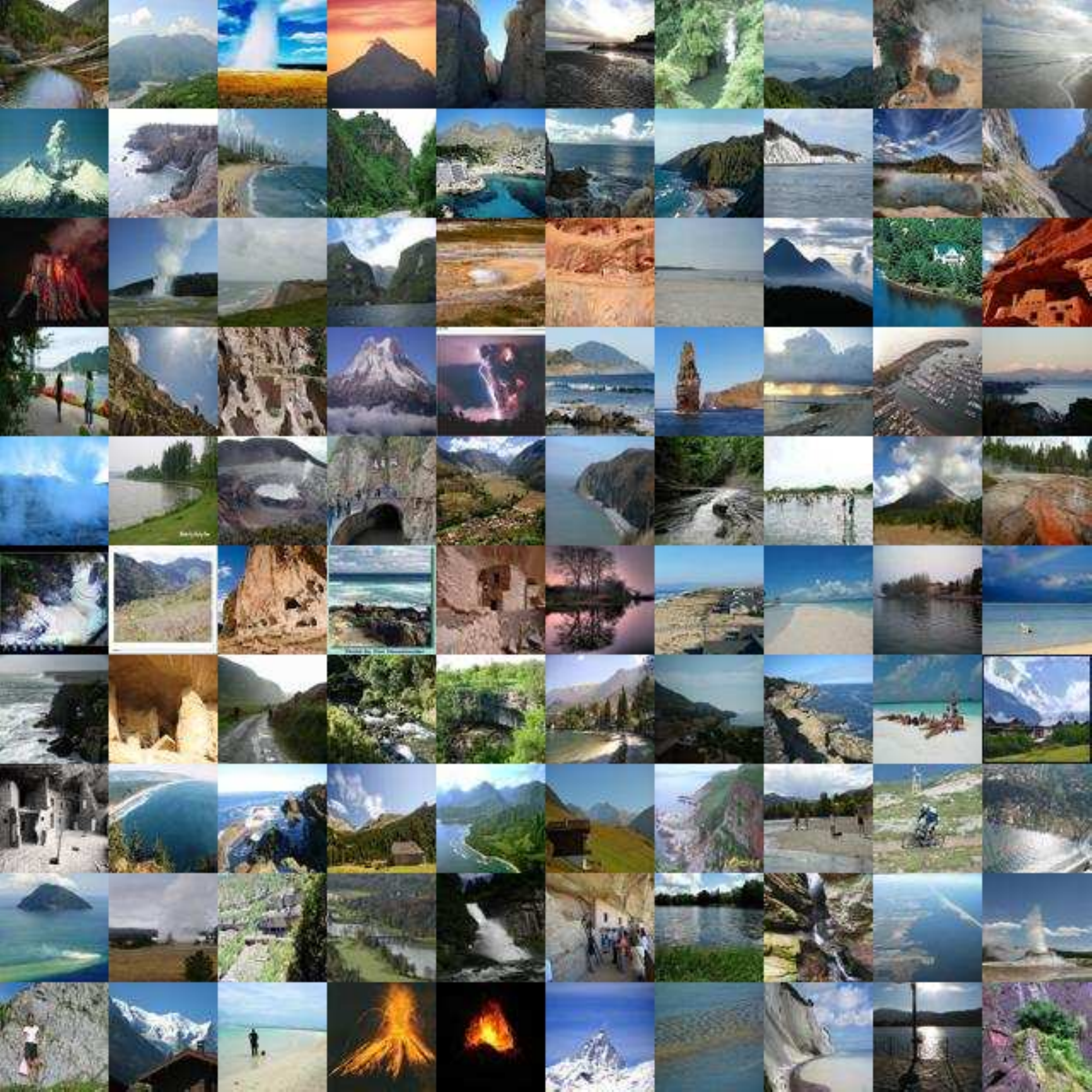} \hspace{2mm}
	\includegraphics[height=.4\linewidth, width=.4\linewidth]{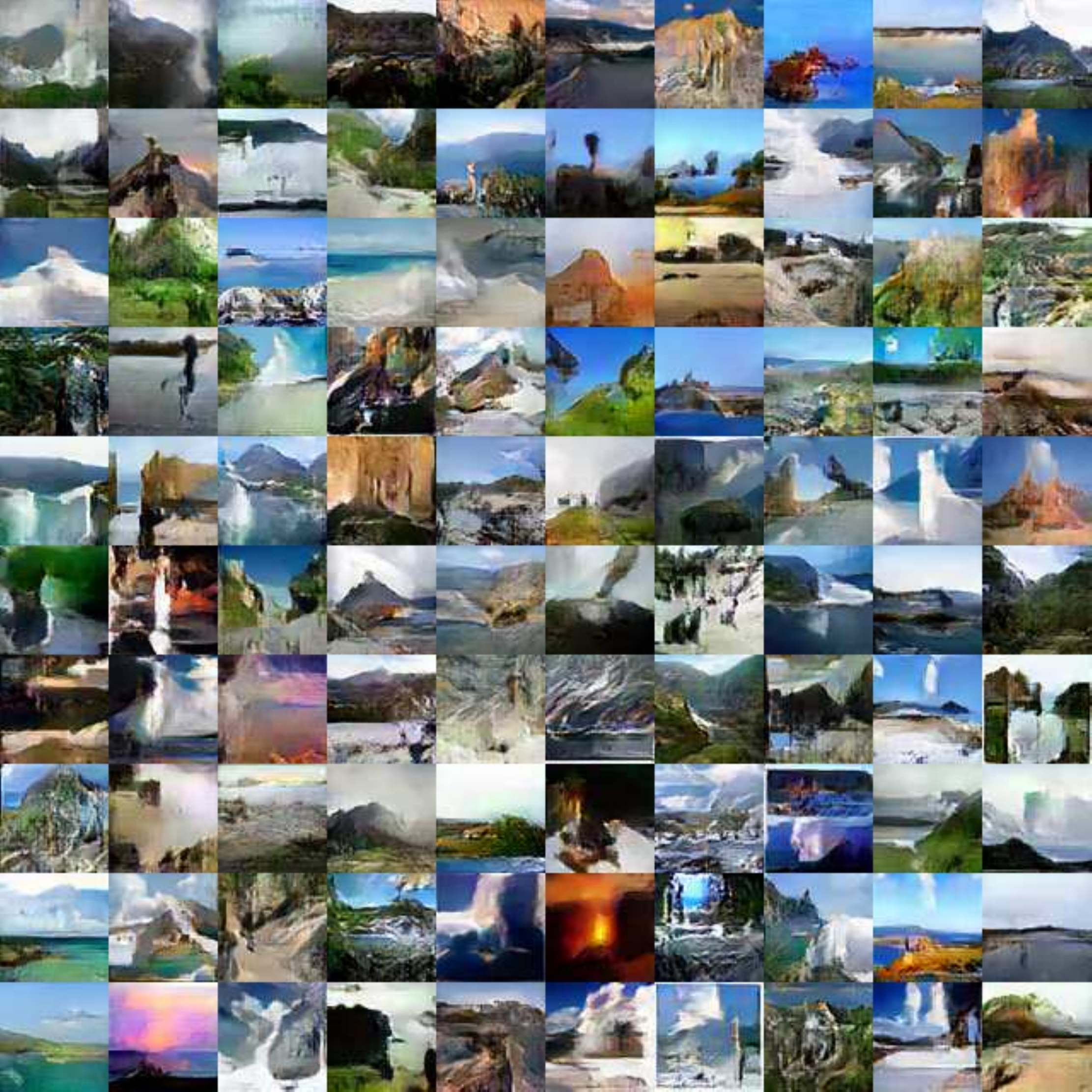}
	\caption{Cooperative learning \cite{CoopNets2016, coopnets2018}. The training set consists of 11,000 images (64 $\times$ 64) randomly sampled from 10 Imagenet scene categories. Left panel: random samples of training  images.  Right panel: random samples generated by the learned models. }	
	\label{fig:syn_10category1}
\end{figure}

We conduct experiments on learning from Imagenet dataset \cite{deng2009imagenet}. We adopt a 4-layer descriptive model and a 5-layer generative model. We set the number of Langevin dynamics steps in each learning iteration to $l=10$. The number of learning iterations is $1,000$. After learning the models, we synthesize images using the learned models. 

In our first experiment, we learn from images that are randomly sampled from 10 Imagenet scene categories. The number of images sampled from each category is 1100. We mix the images from all the categories as a single training set.  Figure \ref{fig:syn_10category1} displays the observed examples randomly sampled from the training set, and the synthesized examples generated by the CoopNets algorithm.   

\begin{figure}[h]
	\centering	
	\includegraphics[width=.45\linewidth]{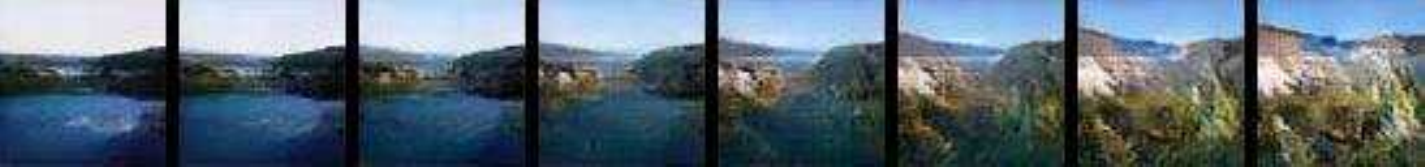} \hspace{1mm}
	\includegraphics[width=.45\linewidth]{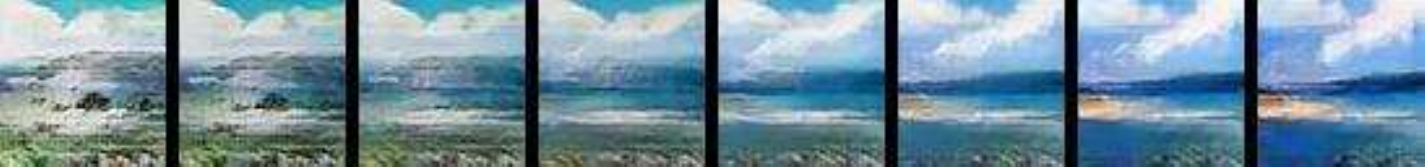} \\ \vspace{.5mm}	 
	\includegraphics[width=.45\linewidth]{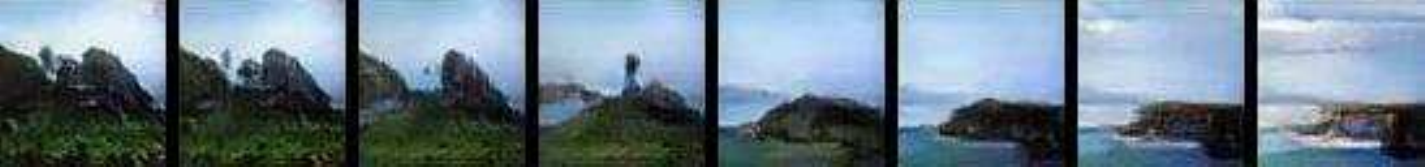} \hspace{1mm}
	\includegraphics[width=.45\linewidth]{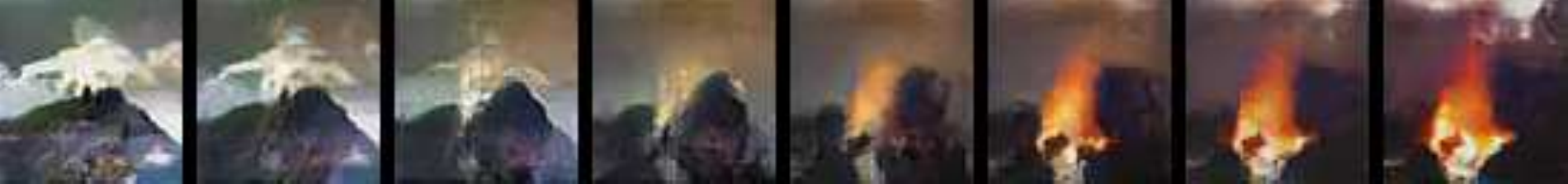}	
	\caption{Cooperative learning \cite{CoopNets2016, coopnets2018}. Interpolation between latent vectors of the images on the two ends.}	
	\label{fig:iterpolation}
\end{figure}

 Figure \ref{fig:iterpolation} shows 4 examples of interpolating between  latent vectors $h$. For each row, the images at the two ends are generated by $h$ vectors randomly sampled from ${\rm N}(0,I_d)$. Each image in the middle is obtained by first interpolating the $h$ vectors of the two end images, and then generating the image using the learned models.  This experiment shows that we learn smooth generative model that traces the manifold of the data distribution. 

We evaluate the synthesis quality by the Inception score \cite{salimans2016improved}.  Our method is competitive to DCGAN \cite{radford2015unsupervised}, EBGAN \cite{zhao2016energy}, Wasserstein GAN \cite{arjovsky2017wasserstein}, InfoGAN \cite{chen2016infogan}, VAE \cite{KingmaCoRR13}, the method of \cite{Bengio2016}. 

Compared to  the three nets in \cite{dai2017calibrating}, the cooperative learning method only needs two nets. Moreover, the finite-step MCMC serves to bridge the generative model and the descriptive model, so that the synthesized examples are closer to fair samples from the descriptive model. 

\section{Discussion}
To summarize the relationships between the non-hierarchical linear forms and the hierarchical non-linear forms of the three families of models, the non-hierarchical form has one layer of features or hidden variables, and they are designed. The hierarchical form has multiple layers of features or hidden variables, and all the layers are learned from the data. 

\begin{figure}[h]
\begin{center}
\includegraphics[width=0.4\textwidth]{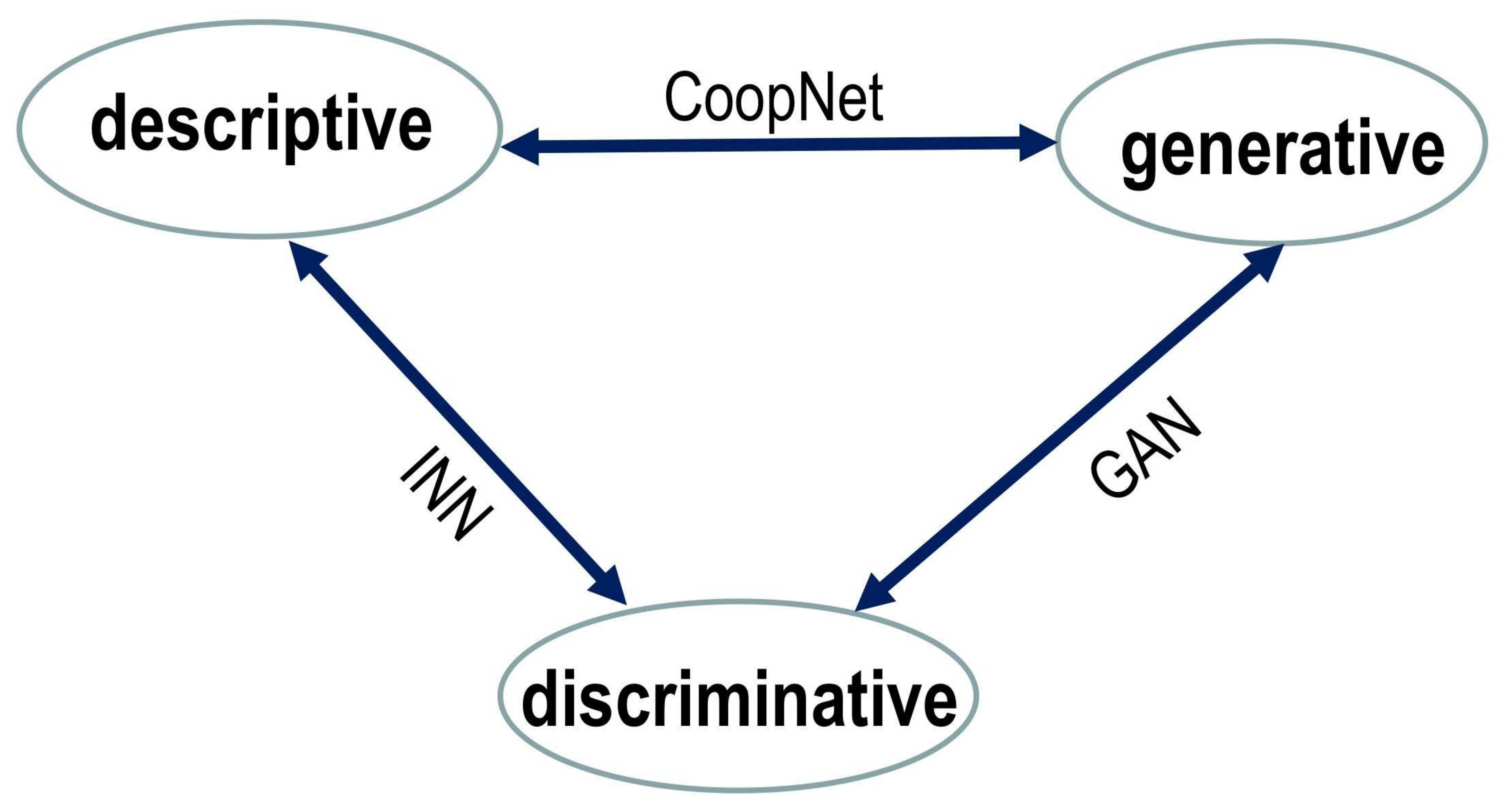}
\end{center}
		\caption{The connections between the three families of models. The discriminative and the generative models are connected by the generative adversarial networks (GAN). The discriminative and the descriptive models are connected by the introspective neural networks (INN). The descriptive and the generative models are connected by cooperative learning. }
		\label{fig:three}
\end{figure}

To summarize the relationships between the three families of models, we have the following connections: 
\begin{enumerate}
\item The discriminative model and the descriptive model can be translated into each other by the Bayes rule. The introspective learning method unifies the two models. 

\item The descriptive model and the generative model can be learned together by adversarial contrastive divergence or the cooperative learning method via MCMC teaching. 

\item The discriminative model and the generative model can be learned together by adversarial training. 
\end{enumerate}
See Figure \ref{fig:three} for an illustration. 

Besides the  models reviewed in this paper, there are other probabilistic models, such as the deep Boltzmann machine \cite{Hinton06, salakhutdinov2009deep, lee2009convolutional},  which is an energy-based model with multiple layers of latent variables,  auto-regressive models \cite{oord2016pixel}, the deep generalizations of the independent component analysis model \cite{dinh2014nice, dinh2016density}.   

In the cooperative learning, the descriptive model and the generative model are parametrized by separate networks. It is more desirable to integrate the two classes of models within a common network. 

\begin{figure}[h]
\begin{center}
\includegraphics[width=0.8\textwidth]{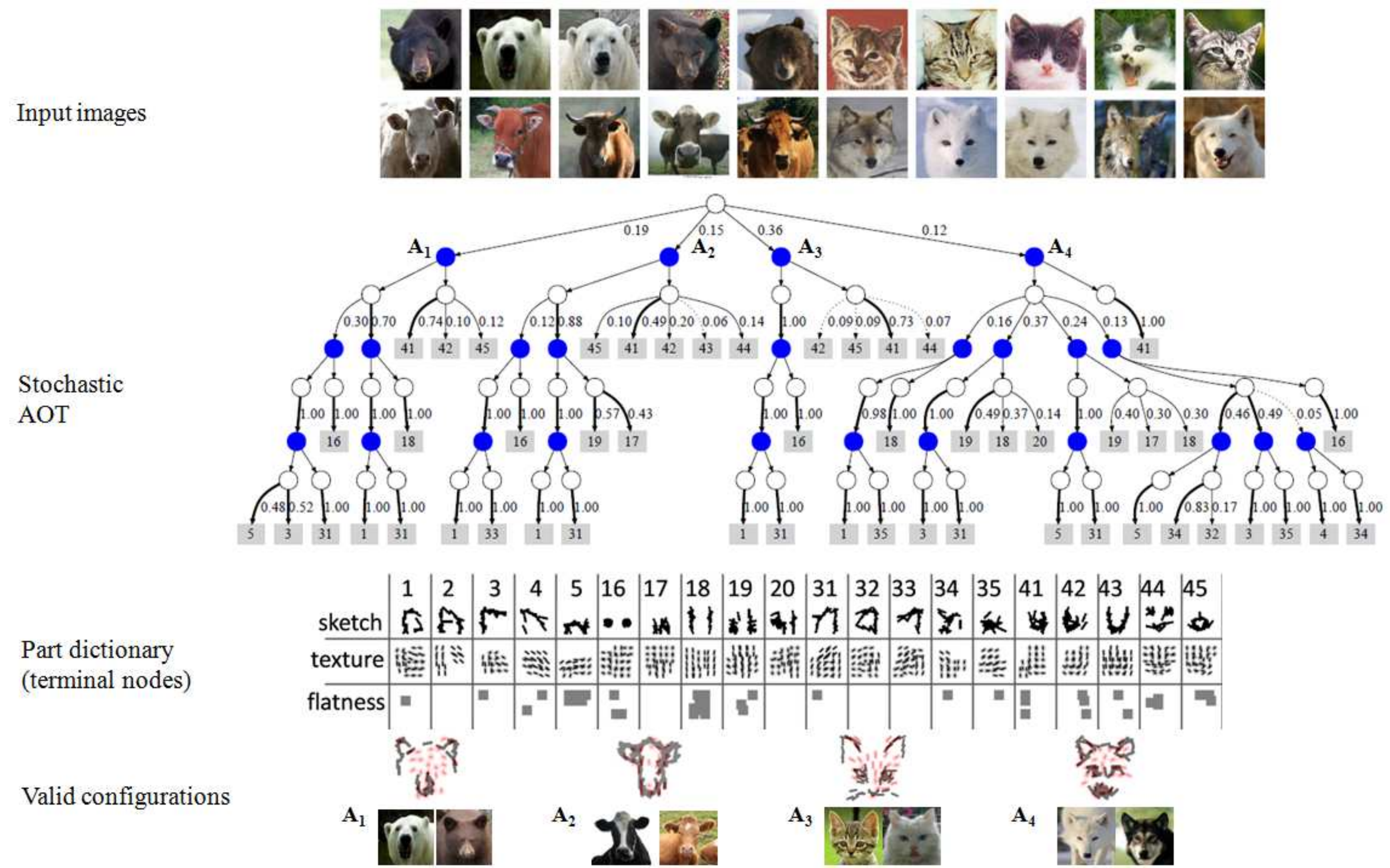}
\end{center}
		\caption{And-Or template \cite{si2013learning} for modeling  recursive compositions of alternative parts: Each And node (blue circle) is a composition of some Or nodes. Each Or node (blank circle) is a probability distribution over some And nodes. An And node models the composition of parts. An Or node models the alternative choices of each part. }
		\label{fig:AOT}
\end{figure}

The existing models are still quite far from what Grenander might have searched for, in that they are still more or less black box models with ConvNet parametrizations. A more interpretable model is the And-Or graph  \cite{zhu2007stochastic}, which alternates between layers of And nodes and Or nodes. An And node models the composition of parts, while an Or node models the alternative choices of parts according to a certain probability distribution. Such an And-Or grammar can generalize to unseen patterns by reconfiguration of parts. In fact the neural network can be interpreted as a dense version of And-Or graph in that the linear weighted sum can be interpreted as And nodes and the rectification and max pooling can be interpreted as Or nodes. Figure \ref{fig:AOT} shows an example of And-Or template of animal faces \cite{si2013learning}.   

\begin{figure}[h]
\begin{center}
\includegraphics[width=0.5\textwidth]{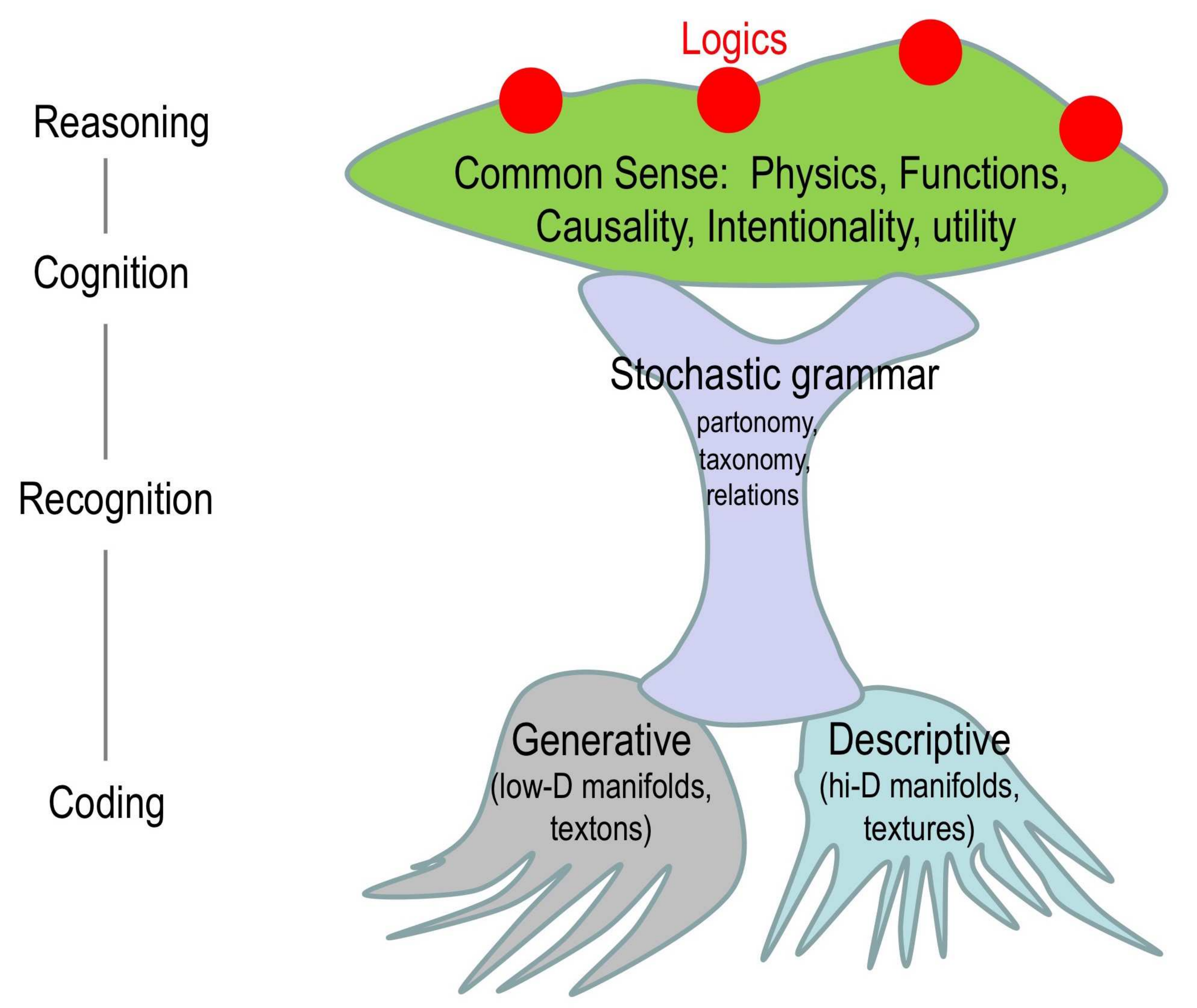}
\end{center}
		\caption{Hierarchical representation of patterns, with simple generative and descriptive models for textons and textures at the lower layers, the stochastic grammar in the middle layers, and logic reasoning with common sense at higher layers. }
		\label{fig:H}
\end{figure}

Ideally, as illustrated by Figure \ref{fig:H},  we should have simple descriptive and generative models at the lowest layers, with the descriptive models accounting for high dimensional or high entropy patterns such as stochastic textures, and the generative models accounting for low-dimensional or low entropy patterns such as textons. In the middle layers we should have stochastic grammars to define the explicit compositional patterns of objects and their parts, as well as their relations  \cite{geman2002composition, zhu2007stochastic}. At the top layer, we  should have logical reasoning based on the learned  common sense about physics, funtionality and causality. It is our hope that a unified model of this form can be developed in the future.

\section*{Acknolwedgment}

The work is supported by NSF DMS 1310391, DARPA SIMPLEX N66001-15-C-4035,  ONR MURI N00014-16-1-2007, and  DARPA ARO W911NF-16-1-0579.

\bibliography{mybibfile}
\bibliographystyle{amsplain}

\end{document}